\documentclass[runningheads]{llncs}
\usepackage[T1]{fontenc}
\usepackage{graphicx}

\usepackage{color}
\usepackage{amsmath}
\usepackage{amsfonts}
\usepackage{amssymb}
\usepackage{epstopdf}
\usepackage{multirow}
\usepackage{tikz}
\usepackage{tikzit}

\tikzstyle{blue dot}=[fill={rgb,255: red,31; green,119; blue,180}, draw=black, shape=circle]
\tikzstyle{orange dot}=[fill={rgb,255: red,255; green,127; blue,14}, draw=black, shape=circle]
\tikzstyle{green dot}=[fill={rgb,255: red,44; green,160; blue,44}, draw=black, shape=circle]
\tikzstyle{red dot}=[fill={rgb,255: red,214; green,39; blue,40}, draw=black, shape=circle]
\tikzstyle{purple dot}=[fill={rgb,255: red,148; green,103; blue,189}, draw=black, shape=circle]
\tikzstyle{brown dot}=[fill={rgb,255: red,140; green,86; blue,75}, draw=black, shape=circle]

\tikzstyle{dashed arrow}=[->, dashed]
\tikzstyle{arrow}=[->]

\usepackage[colorinlistoftodos]{todonotes}
\usepackage{subfig}
\usepackage{algorithm}
\usepackage{algpseudocode}
\usepackage[numbers]{natbib}
\usepackage{bbm}

\usepackage{dsfont}
\usepackage{framed}
\usepackage{verbatim}

\usepackage{enumitem}
\usepackage{float}
\usepackage{mathtools}
\usepackage{hyperref}
\usepackage{tikz}
\usetikzlibrary{arrows}
\usetikzlibrary{positioning}
\usetikzlibrary{fit}
\usetikzlibrary{calc}
\tikzstyle{A}=[circle,
thick,
minimum size=.5cm,
draw=blue!80,
fill=blue!20
]
\tikzstyle{B}=[circle,
thick,
minimum size=.5cm,
draw=orange!80,
fill=orange!25]

\makeatletter
\def\lastdist{}
\def\diststyle#1{#1}
\def\@printdistname[#1]{\ifx\hfuzz#1\hfuzz\diststyle{\mathrm{\lastdist}}\else\diststyle{\mathrm{\lastdist}}_{#1}\fi}%
\def\@printdist[#1]#2{\@printdistname[#1]\mathopen{}\left(#2\right)\mathclose{}}%
\def\makedist#1#2{%
  \expandafter\def\csname @#1\endcsname{\def\lastdist{{#2}}\@printdist}%
  \expandafter\def\csname #1\endcsname{\@testopt{\csname @#1\endcsname}{}}%
  \expandafter\def\csname @@#1\endcsname{\def\lastdist{{#2}}\@printdistname}%
  \expandafter\def\csname txt#1\endcsname{\@testopt{\csname @@#1\endcsname}{}}%
  \expandafter\def\csname s#1\endcsname{\mathrm{#1}}%
}%
\makeatother
\makedist{GammaP}{\ensuremath{\Gamma\mathrm{P}}}
\makedist{UniformD}{\ensuremath{\mathcal{U}}}
\makedist{Poisson}{Poisson}
\makedist{Gaussian}{\ensuremath{\mathcal{N}}}
\makedist{tPoisson}{tPoisson}
\makedist{zPoisson}{zPoisson}
\makedist{GammaD}{Gamma}
\makedist{BetaD}{Beta}
\makedist{exp}{exp}
\makedist{PP}{PP}
\makedist{CRP}{CRP}
\makedist{Binomial}{Binomial}
\makedist{tBinomial}{tBinomial}
\makedist{PK}{PK}
\makedist{logn}{lognormal}

\newcommand{\1}[1]{\mathbbm{1}_{#1}}
\newcommand{\GG}{\text{GG}}

\begin{document}

\title{Bayesian Nonparametrics for Sparse Dynamic Networks}

\titlerunning{Bayesian Nonparametrics for Sparse Dynamic Networks}

\author{Cian Naik\inst{1} \and
Fran\c cois Caron\inst{1} \and
Judith Rousseau\inst{1} \and
Yee Whye Teh\inst{1,3}\and
Konstantina Palla\inst{2}}
\authorrunning{C. Naik et al.}

\institute{Department of Statistics, University of Oxford, Oxford, United Kingdom \and
Microsoft Research, Cambridge, United Kingdom \and Google Deepmind, London, United Kingdom}

\maketitle             
\begin{abstract}
In this paper we propose a Bayesian nonparametric approach to modelling sparse time-varying networks. A positive parameter is associated to each node of a network, which models the sociability of that node. Sociabilities are assumed to evolve over time, and are modelled via a dynamic point process model. The model is able to capture long term evolution of the sociabilities. Moreover, it yields sparse graphs, where the number of edges grows subquadratically with the number of nodes. The evolution of the sociabilities is described by a tractable time-varying generalised gamma process. We provide some theoretical insights into the model and apply it to three datasets: a simulated network, a network of hyperlinks between communities on Reddit, and a network of co-occurences of words in Reuters news articles after the September $11^{th}$ attacks.

\end{abstract}
\section{Introduction}

This article is concerned with the analysis of dynamic networks, where one observes the evolution of links among a set of objects over time. As an example, links may represent social interactions between individuals over time or the co-occurrence of words in a newspaper over time. Probabilistic approaches treat the dynamic networks of interest as random graphs, where the vertices (nodes) and edges correspond to objects and links respectively. In the graph setting, sparsity is defined in terms of the rate in which the numbers of edges grows as the number of nodes increases. In a \textit{sparse} graph the number of edges grows sub-quadratically in the number of nodes. Hence, in a large graphs, two nodes chosen at random are very unlikely to be linked. 

While sparsity is a property found in many real-world network datasets \citep{Newman2009}, most of the popular Bayesian models used in network analysis account for dense graphs, i.e. where the number of edges grows quadratically in the number of nodes, see \citep{Orbanz2015} for a review. A recent Bayesian nonparametric approach, proposed by \citep{Caron2017} and later developed in a number of articles~\citep{Veitch2015,Herlau2015,Borgs2018,Todeschini2016,naik2021sparse} represents the graph as an infinite point process on $\mathbb{R}^2_+$, giving rise to a class of sparse random graphs. This class of sparse models is projective and admits a representation theorem due to \citep{Kallenberg1990}. 

In this paper, we are interested in the dynamic domain and aim to probabilistically model the evolution of sparse graphs over time, where edges may appear and disappear, and the node popularity may change over time. We build on the sparse graph model of \citep{Caron2017} and extend it to deal with time series of network data.  We describe a fully generative and projective approach for the construction of sparse dynamic graphs. It is challenging to perform exact inference using the framework we introduce, and thus we consider an approximate inference method, using a finite-dimensional approximation introduced by \citep{lee2016finite}.

The rest of the article is structured as follows. In Section \ref{sec:background} we give some background on the sparse network model of \citep{Caron2017}. Section \ref{sec:model} describes the novel statistical dynamic network model we introduce in detail. In Section \ref{sec:properties}, we describe the sparsity properties of the proposed model. The approximate inference method, based on a truncation of the infinite-dimensional model is described in Section~\ref{sec:inference}. In Section \ref{sec:experiments} we present illustrations of our approach to three different dynamic networks with thousands of nodes and edges.

\section{Background: model of Caron and Fox for sparse static networks}
\label{sec:background}
We recall in this section the model of \citep{Caron2017} for sparse multigraphs. Let $\alpha>0$ be a positive real tuning the size of the network. A finite multigraph of size $\alpha>0$ is represented by a point process on $[0,\alpha]^2$
\[
N=\sum_{i, j} n_{ij} \delta_{(\theta_i,\theta_j)}
\]
where $n_{ij}=n_{ji}\in\{0,1,2,\ldots\}$, $i\leq j$, represents the number of interactions between individuals $i$ and $j$, and the $\theta_i\in[0,\alpha]$ can be interpreted as node labels. Those node labels are introduced for the model's construction, but are not observed nor inferred. Each node $i$ is assigned a sociability parameter $w_i>0$. Let $W=\sum_i w_i\delta_{\theta_i}$ be the corresponding random measure on $[0,\alpha]$. We assume that $W$ is a generalised gamma completely random measure~\citep{Kingman1967,Hougaard1986,Brix1999,Lijoi2007}. That is, $\{(w_i,\theta_i)_{i\geq 1}\}$ are the points of a Poisson point process with mean measure $\nu(w)dw \1{\theta\leq \alpha}d\theta$ where $\mathbbm{1}_{A}=1$ if the statement $A$ is true and $0$ otherwise, and $\nu$ is a L\'evy intensity on $(0,\infty)$ defined as
\begin{equation}
\nu(w)=\frac{1}{\Gamma(1-\sigma)}w^{-1-\sigma}e^{-\tau w}
\end{equation}
with hyperparameters $\sigma<1$ and $\tau>0$. We write simply $W\sim \GG(\alpha,\sigma,\tau)$.

To each pair of nodes $i,j$, we assign a number of latent interactions $n_{ij}$, where
\begin{equation}n_{ij}\mid w_{i},w_{j} \sim\left \{
\begin{array}{ll}
  \Poisson{2w_{i}w_{j}} & i< j , \quad n_{ji}= n_{ij}\\
  \Poisson{w_{i}w_{j}} & i=j
\end{array}\right .
\end{equation}
Finally, two nodes are said to be connected if they have at least one interaction; let $z_{ij}=\1{n_{ij}>0}$ be the binary variable indicating if two nodes are connected. When $\sigma>0$, such model yields sparse graphs with power-law degree distributions~\citep{Caron2017,Caron2017a}.

\begin{figure}[th]
\scalebox{0.8}{\begin{tikzpicture}
	\begin{pgfonlayer}{nodelayer}
		\node [style=none] (0) at (-7, 0) {};
		\node [style=orange dot] (1) at (-3, 2) {$W_{t-1}$};
		\node [style=blue dot] (2) at (1, 0) {$C_{t-1}$};
		\node [style=orange dot] (3) at (5, 2) {$W_{t}$};
		\node [style=blue dot] (4) at (9, 0) {$C_t$};
		\node [style=orange dot] (5) at (13, 2) {$W_{t+1}$};
		\node [style=blue dot] (6) at (17, 0) {$C_{t+1}$};
		\node [style=none] (7) at (21, 2) {};
		\node [style=red dot] (20) at (1, 6) {$\alpha,\sigma,\tau$};
		\node [style=red dot] (21) at (9, 8) {$\phi$};
		\node [style=green dot] (22) at (-3, -4) {$n_{t-1}$};
		\node [style=green dot] (23) at (5, -4) {$n_{t}$};
		\node [style=green dot] (24) at (13, -4) {$n_{t+1}$};
	\end{pgfonlayer}
	\begin{pgfonlayer}{edgelayer}
		\draw [style=arrow] (1) to (2);
		\draw [style=arrow] (2) to (3);
		\draw [style=arrow] (3) to (4);
		\draw [style=arrow] (4) to (5);
		\draw [style=arrow] (5) to (6);
		\draw [style=arrow] (20) to (1);
		\draw [style=arrow] (20) to (3);
		\draw [style=arrow] (20) to (5);
		\draw [style=arrow] (21) to (2);
		\draw [style=arrow] (21) to (4);
		\draw [style=arrow] (21) to (6);
		\draw [style=dashed arrow] (0.center) to (1);
		\draw [style=dashed arrow] (6) to (7.center);
		\draw [style=arrow] (1) to (22);
		\draw [style=arrow] (3) to (23);
		\draw [style=arrow] (5) to (24);
	\end{pgfonlayer}
\end{tikzpicture}}
\caption{Graphical representation of the model. The counts $n_{t}$ are derived from the sociabilities $W_t$, whose time-evolution depends on the counts $C_t$ and hyperparameters $\alpha, \sigma, \tau$ and $\phi$.}
\label{fig:graphical_ii}
\end{figure}
\section{Dynamic statistical network model}
\label{sec:model}

In order to study dynamically evolving networks, we assume that at each time $t=1,2,\ldots,T$, we observe a set of interactions between a number of nodes. This set of interaction is represented by a point process $N_t$ over $[0,\alpha]^2$ as in Equation~\eqref{eq:Zt}, where $\alpha$ tunes the size of the graphs. 
\begin{equation}
N_t=\sum_{i, j} n_{tij} \delta_{(\theta_i,\theta_j)}.
\label{eq:Zt}
\end{equation}
Here, $n_{tij}$ is the number of interactions between $i$ and $j$ at time $t$, and the $\theta_i$ are unique node labels.

The dynamic point process $N_t$ is obtained as follows. We assume that each node $i$ at time $t$ has a \textit{sociability} parameter $w_{ti}\in \mathbb{R_+}$, that can be thought of as a measure of the node's willingness to interact with other nodes at time $t$. We consider the associated collection of random measures on $\mathbb R_+$, for $t=1,\ldots,T$
$$
W_t=\sum_i w_{ti} \delta_{\theta_i},~~~t=1,\ldots,T.
$$
We first describe in Section \ref{sec:latentcount} the model for the latent interactions. Then we describe in Section \ref{sec:sociability} the model for the time-varying sociability parameters $(W_t)_{t\geq 1}$. The overall probabilistic model is summarised in Figure \ref{fig:graphical_ii}.

\subsection{Dynamic network model based on observed interactions}
\label{sec:latentcount}
In the dynamic setting, what we observe in practice is often counts of interactions between nodes, e.g. hyperlinks, emails or co-occurrences, rather than a binary indicator of whether there is a connection between them. So for each pair of nodes $i\leq j$, we let $(n_{tij})_{t=1,2,\ldots T, j \geq i}$ be the interaction count between them at time t. We assume that $n_{tij}$ can be modelled as
\begin{equation}\label{eq:n_poiss}
n_{tij}\mid w_{ti},w_{tj} \sim\left \{
\begin{array}{ll}
  \Poisson{2w_{ti}w_{tj}} & i< j, \quad  n_{tji} = n_{tij}\\
  \Poisson{w_{ti}w_{tj}} & i=j
\end{array}\right.
\end{equation}
This model can be easily adapted to graphs with directed edges, by modifying the Equation (\ref{eq:n_poiss}) to 
$$
n_{tij}\sim {Poisson}(w_{ti}w_{tj})
$$
for all $i\neq j$, where $n_{tij}$ now represents the number of interactions from $i$ to $j$. The resulting inference algorithm essentially remains the same, and from now on we assume we are in the undirected edge setting. 
 As in the static case, we can reconstruct the binary graph by letting $z_{tij}= \1{(n_{tij}>0)}$ be the binary variable indicating if nodes $i$ and $j$ are connected at time $t$, i.e. two nodes are connected at time $t$ if and only if $n_{tij}>0$. To avoid ambiguity, we say that the number of edges in the graph at time $t$ is $\sum_{i>j}z_{tij}$, rather than counting the number of interactions between pairs of nodes. Marginalizing out the interaction counts $n_{tij}$, we have for $i\neq j$:
\begin{equation}
\Pr(z_{tij}=1\mid (w_{t-k,i},w_{t-k,j})_{k=0,\ldots,t-1})=1-e^{-2w_{ti}w_{tj}}.\label{eq:marginalz}
\end{equation}

\subsection{A dependent generalised gamma process for the sociability parameters}
\label{sec:sociability}
\begin{figure*}[t]
\centering
\subfloat[$t=1$\label{fig:sigma_degree_t_0}]{\includegraphics[width=0.25\textwidth]{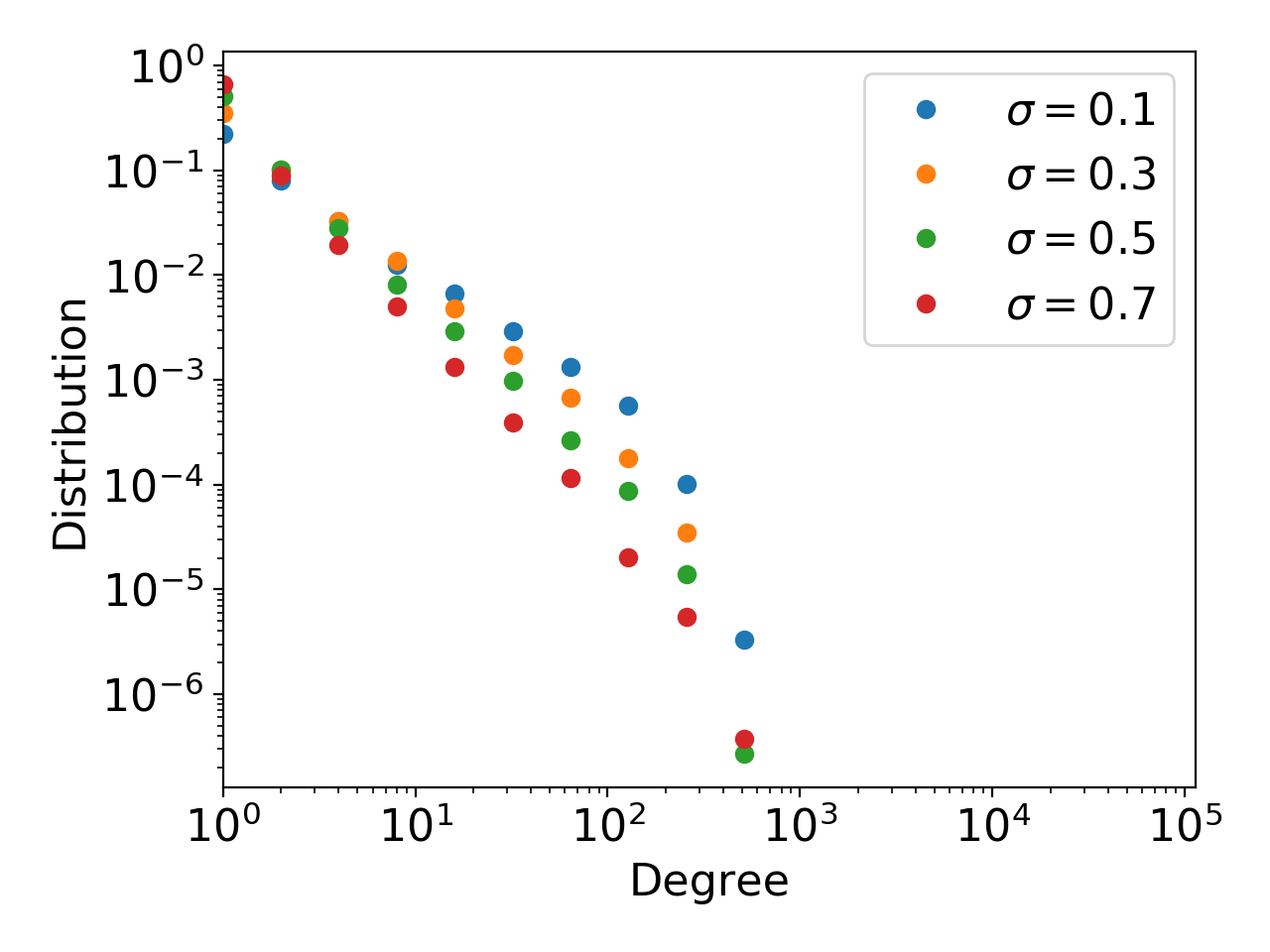}}
\subfloat[$t=2$\label{fig:sigma_degree_t_1}]{\includegraphics[width=0.25\textwidth]{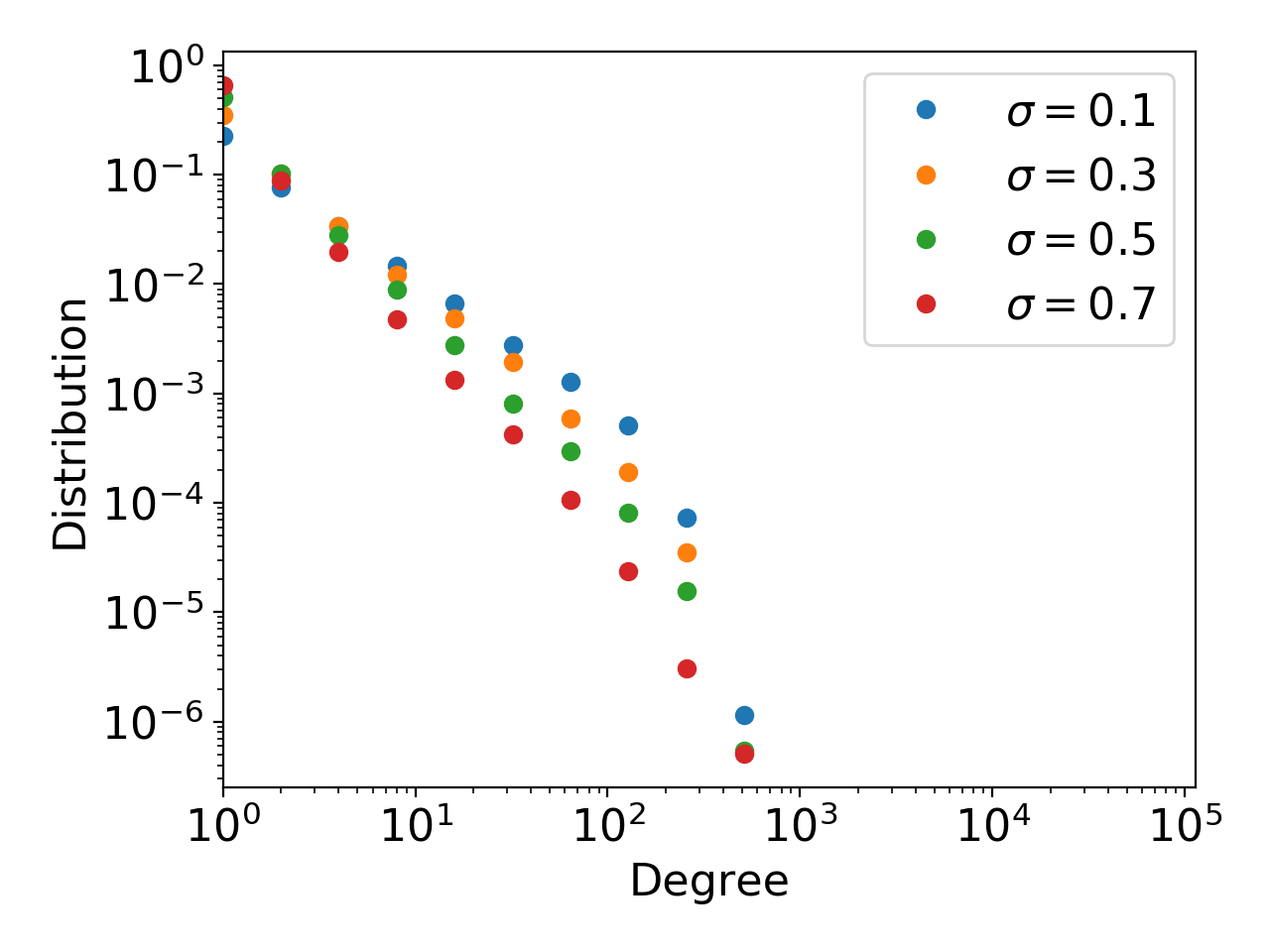}}
\subfloat[$t=3$\label{fig:sigma_degree_t_2}]{\includegraphics[width=0.25\textwidth]{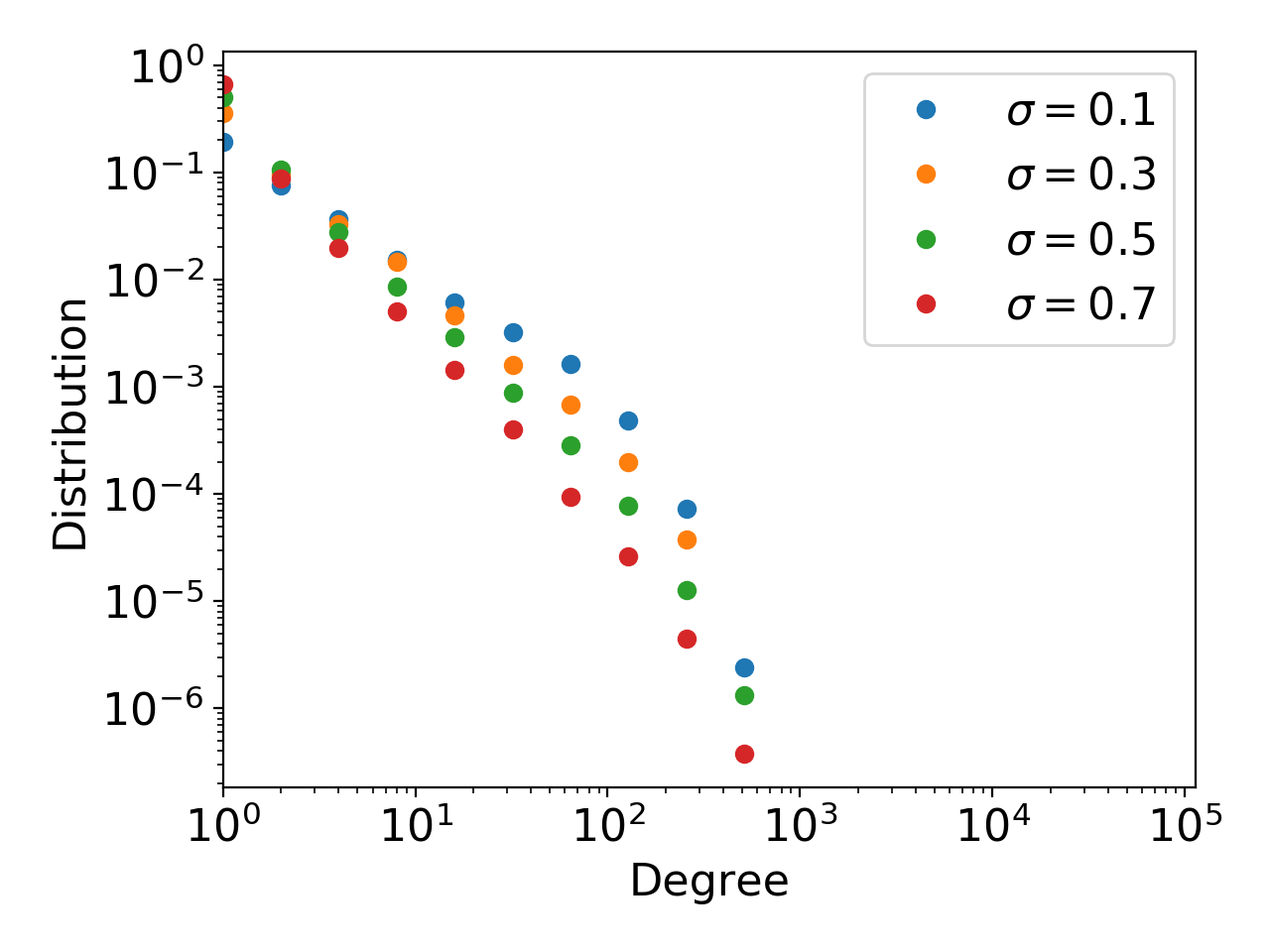}}
\subfloat[$t=4$\label{fig:sigma_degree_t_3}]{\includegraphics[width=0.25\textwidth]{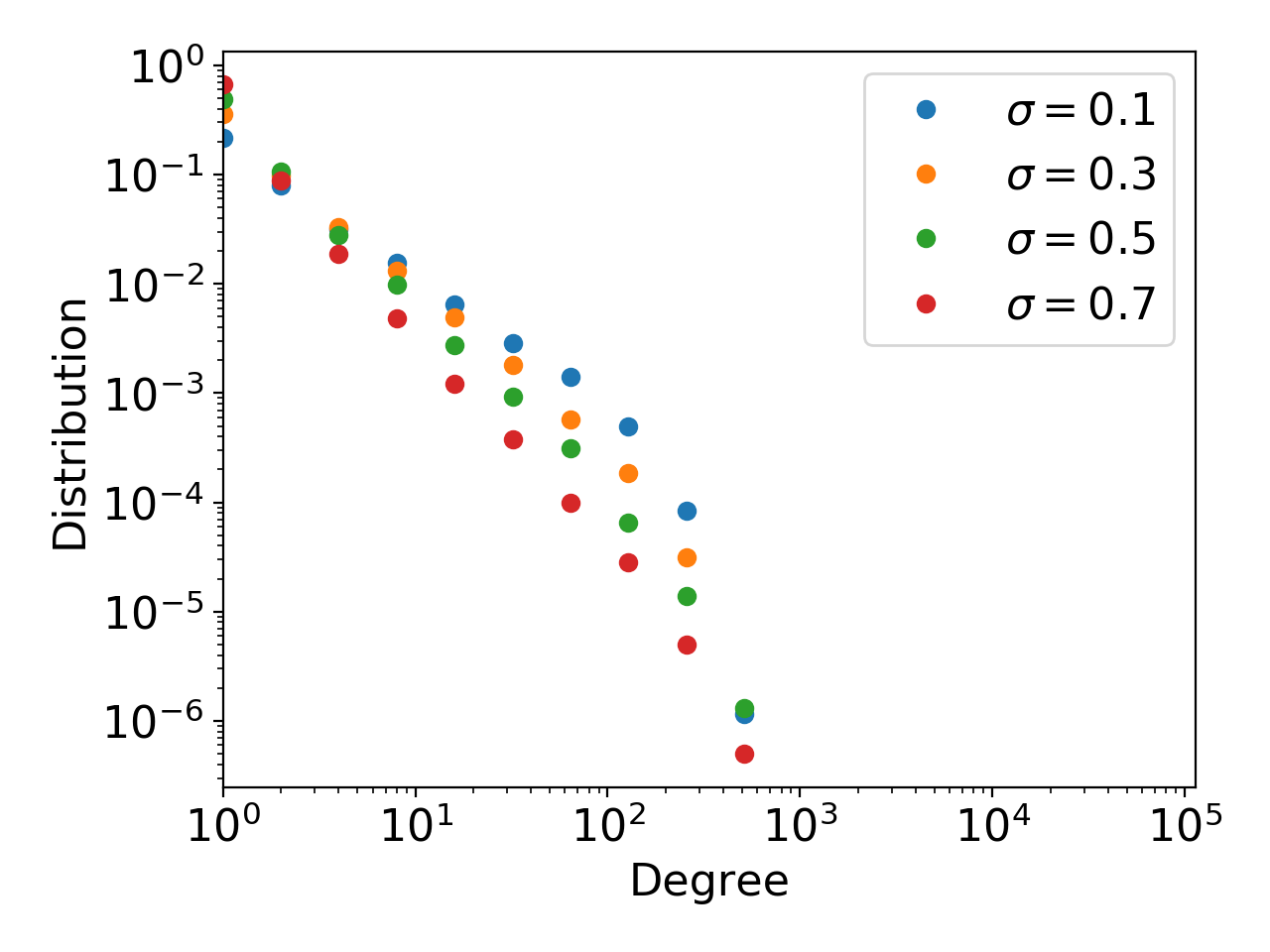}}

\caption{Degree distributions over time, for a network simulated from the GG model with $T=4$, $\alpha = 200$, $\tau = 1$, $\phi = 1$ and varying values of $\sigma$.}
\label{fig:ggp_simulation_sigma_degree}
\end{figure*}
\begin{figure}[t]
\centering
\subfloat[$\phi = 20$\label{fig:phi_20_weights}] {\includegraphics[width=0.4\textwidth]{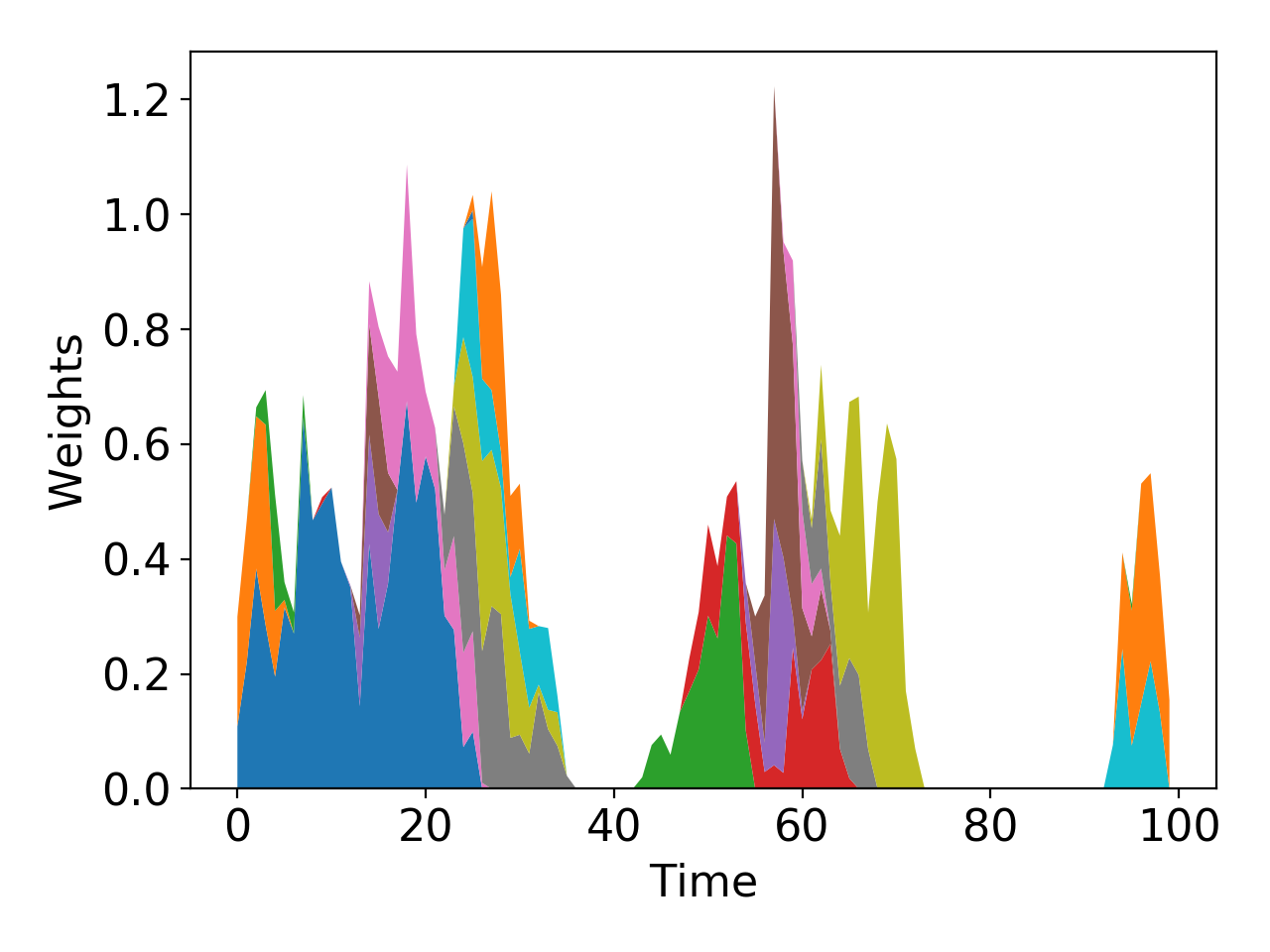}}
\hspace{0.01\textwidth}
\subfloat[$\phi = 2000$ \label{fig:phi_2000_weights}]{\includegraphics[width=0.4\textwidth]{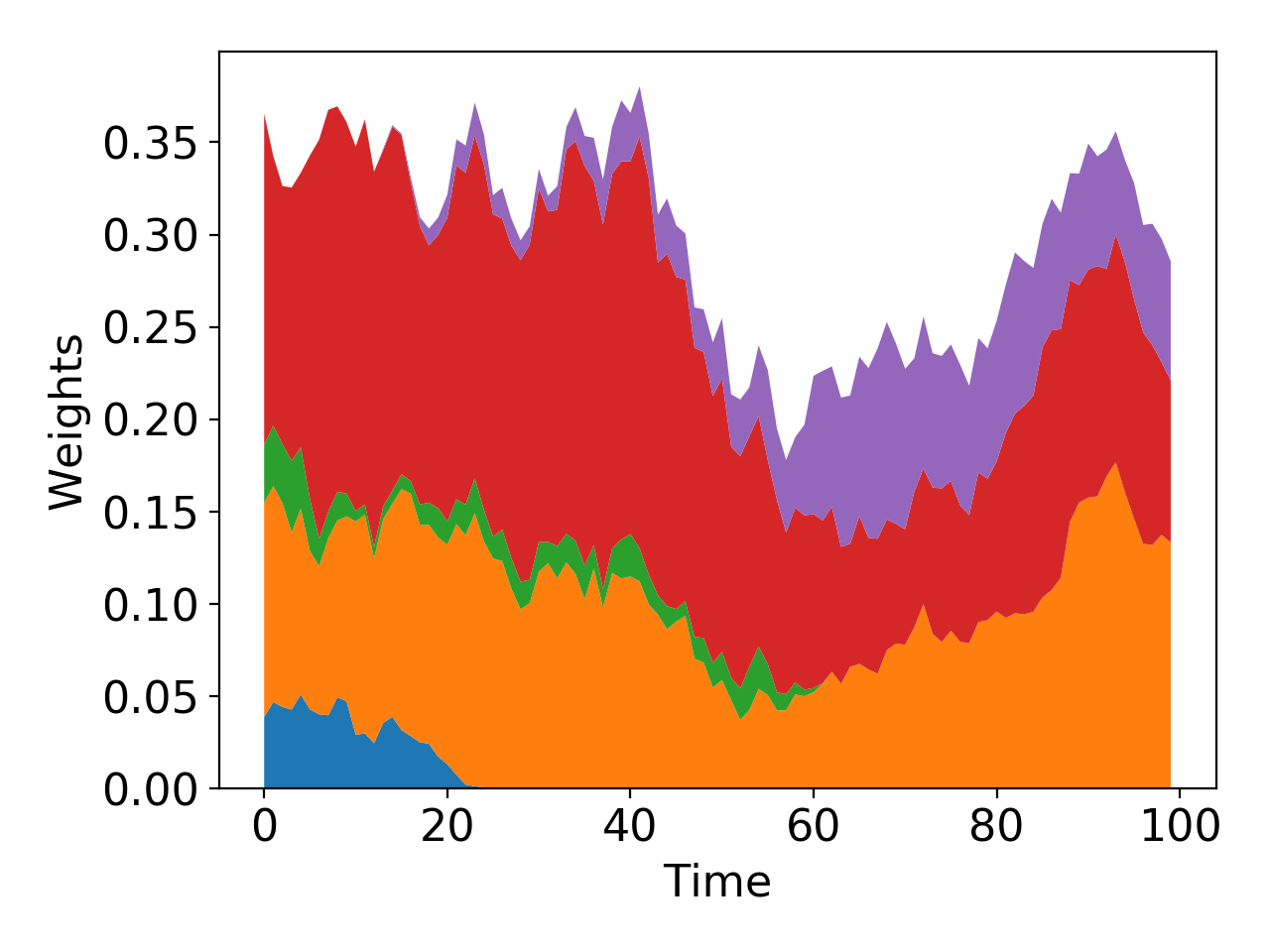}}

\caption{Evolution of weights over time, for a network simulated from the GG model with $T=100$, $\alpha = 1$, $\sigma = 0.01$, $ \tau = 1$ and (a) $\phi = 20$ (b) $\phi=2000$.}
\label{fig:ggp_simulation_phi_weights}
\end{figure}
We consider here that the sequence of random measures $(W_t)_{t=1,2,\ldots}$ follows a Markov model, such that $W_t$ is marginally distributed as $\GG(\alpha,\sigma,\tau)$. To this aim, we build on the generic construction of \citep{Pitt2005}. A similar model has been derived by \citep{CarTeh2012a} for dependent gamma processes (corresponding to the case $\sigma=0$ here). As in \citep{Caron2017}, we use the generalised gamma process here because of the flexibility the sparsity parameter $\sigma$ gives us. In particular, this setup allows us to capture power-law degree distributions, unlike with the gamma process.

For a sequence of additional latent variables $(C_t)_{t=1,2,\ldots}$, we consider a Markov chain $W_{t}%
\rightarrow C_{t}\rightarrow W_{t+1}$ starting with $W_{1}\sim\GG(\alpha
,\sigma,\tau)$ that leaves $W_{t}$ marginally $\GG(\alpha
,\sigma,\tau)$. For $t=1,\ldots,T-1$, define
\begin{equation}
C_{t}=\sum_{i=1}^{\infty}c_{ti}\delta_{\theta_{i}}\quad c_{ti}|W_{t}%
\sim\text{Poisson}(\phi w_{ti})\label{eq:latentCt}
\end{equation}
where $\phi>0$ is a parameter tuning the correlation. Given $C_t$, the measure $W_{t+1}$ is then constructed as a combination of masses defined
by $C_{t}$ and GG innovation:
\begin{align}
W_{t+1}=W_{t+1}^{\ast}+\sum_{i=1}^{\infty}w_{t+1,i}^{\ast}\delta_{\theta_{i}%
}
\end{align}
with
\begin{align}
W_{t+1}^{\ast}&\sim\GG(\alpha,\sigma,\tau+\phi)\\
w_{t+1,i}^{\ast}|C_{t}&\sim\text{Gamma}(\max(c_{ti}%
-\sigma,0),\tau+\phi).
\end{align}
By convention, $\text{Gamma}(0,\tau)=\delta_0$, hence $w_{t+1,i}^{\ast}=0$ if $c_{ti}=0$ else $w_{t+1,i}^{\ast}>0$,  Because the conditional laws of
$W_{t+1}|C_{t}$ coincide with that of $W_{t}|C_{t}$ \citep{Prunster2002,James2002,James2009},
the construction guarantees that $W_{t+1}$ has the same marginal distribution as $W_{t}$,
i.e., they are both distributed as $\GG(\alpha,\sigma,\tau)$. Moreover,
as proved in Section \ref{sec:app:proofs} of the Appendix,  
the
conditional mean of $W_{t+1}$ given $W_{t}=\sum_{i}w_{ti}\delta_{\theta_i}$ has the form
\begin{align}
E[W_{t+1}&|W_{t}]=\left (\frac{\tau}{\tau+\phi} \right )^{1-\sigma}E[W_t] \nonumber\\
&~+\frac{1}{\tau+\phi}\sum_{i=1}^{\infty}[\phi
w_{ti}-\sigma(1-e^{-\phi w_{ti}})]\delta_{\theta_{i}}\label{eq:conditionalmean}
\end{align}
In the gamma process case $(\sigma=0)$, the above expression reduces to
\[
E[W_{t+1}|W_{t}]=\frac{\tau}{\tau+\phi}E[W_{t}]+\frac{\phi}{\tau+\phi}%
W_{t}.
\]
\subsection{Summary of the model's hyperparameters}

The model is parameterised by $(\alpha, \sigma, \tau, \phi)$, where:
\begin{itemize}
\item $\alpha$ tunes the overall size of the networks, with a larger value of $\alpha$ corresponding to larger networks.
\item $\sigma$ controls the sparsity and power-law properties of the graph, as will be shown in Section~\ref{sec:properties}. In Figure \ref{fig:ggp_simulation_sigma_degree} we see that different values of $\sigma$ give rise to different power-law degree distributions.
\item $\tau$ induces an exponential tilting of large degrees in the degree distribution.
\item $\phi$ tunes the correlation of the sociabilities of each node over time. As we see in Figure \ref{fig:ggp_simulation_phi_weights}, larger values correspond to higher correlation and smoother evolution of the weights.\\
\end{itemize}

\section{Sparsity and power-law properties of the model}
\label{sec:properties}

By construction,
the interactions at time $t$, $n_{tij}$, are drawn from the same (static) model as in \citep{Caron2017}, using a generalised gamma process for the L\'evy intensity, and so applying Proposition 18 in \citep{Caron2017a} we obtain the following asymptotic properties
\begin{proposition}
Let $N_{t,\alpha}$ be the number of active nodes at time $t$, $N_{t,\alpha}^{(e)} = \sum_{i\leq j} z_{tij}$ be the number of edges and $N_{t,\alpha,j}$ the number of nodes of degree $j$ in the graph at time $t$, then
 as $\alpha$ tends to infinity, almost surely, we have for any $t$:
if $\sigma >0$,
$
N_{t,\alpha}^{(e)} \asymp N_{t,\alpha}^{2/(1+\sigma)},
$ if $\sigma=0$, $N_{t,\alpha}^{(e)} \asymp N_{t,\alpha}^{2}/\log^2(N_{t,\alpha})$ and if $\sigma<0$,  $N_{t,\alpha}^{(e)} \asymp N_{t,\alpha}^{2}$.
Also,  almost surely, for any $t\geq 1$ and $j\geq 1$, if $\sigma \in (0,1)$,
$$\frac{N_{t,\alpha,j}}{N_{t,\alpha}}\rightarrow p_j,  \quad p_j=\frac{\sigma\Gamma(j-\sigma)}{j!\Gamma(1-\sigma)},$$
 while if $\sigma\leq 0$, $N_{t,\alpha,j}/N_{t,\alpha} \rightarrow 0$ for all $j\geq 1$.
 \end{proposition}
Hence the graphs are sparse if $\sigma\geq 0$ and dense if $\sigma<0$.

\section{Approximate inference}
\label{sec:inference}

\subsection{Finite-dimensional approximation}
Performing exact inference using this model is quite challenging, and we consider instead an approximate inference method, using a finite-dimensional approximation to the GG prior, introduced by \citep{lee2016finite}. This approximation gives rise to a particularly simple conjugate construction, enabling posterior inference to be performed.

Let $\text{BFRY}\left(\eta,\tau,\sigma\right)$ denote a (scaled and exponentially tilted) BFRY random variable\footnote{The name was coined by \citep{Devroye2014} after \citep{Bertoin2006}.} on $(0,\infty)$ with probability density function
\begin{align*}
    g_{\eta,\tau,\sigma}(w) = \frac{\sigma w^{-1-\sigma}e^{-\tau w}\left( 1-e^{- (\sigma /\eta)^{1/\sigma}w} \right)  }{\Gamma(1-\sigma)\left\{ \left( \tau +(\sigma /\eta)^{1/\sigma} \right)^{\sigma} -\tau^{\sigma}\right\} }
\end{align*}
with parameters $\sigma\in(0,1)$, $\tau>0$ and $\eta>0$. 

At time 1, consider the finite-dimensional measure
\[
W_{1}=\sum_{i=1}^{K}w_{1i}\delta_{\theta_{i}}
\]
where $w_{1i} \sim \text{BFRY}\left(\alpha/K,\tau,\sigma\right)$ and $K<\infty$ is the truncation level. As shown by~\citep{lee2016finite}, for $\sigma\in(0,1)$
$$W_{1}\overset{d}{\to}\GG(\alpha,\sigma,\tau)$$
as the truncation level $K$ tends to infinity. By this we mean that, if $W \sim GG(\alpha,\sigma,\tau)$, then 
$$
\lim_{K \to \infty} \mathcal{L}_f(W') = \mathcal{L}_f(W)
$$
for an arbitrary measurable and positive $f$, where $\mathcal{L}_f(W) \coloneqq \mathbb{E}\left[e^{-W(f)}\right]$ is the Laplace functional of $W$ as defined by \citep{lee2016finite}.

To use this finite approximation with our dynamic model, we consider Poisson latent variables as in Equation~\eqref{eq:latentCt}. The measure $W_{t+1}$ is then constructed as:
\begin{align*}
W_{t+1}&=\sum_{i=1}^{K}w_{t+1,i}\delta_{\theta_{i}}\\
w_{t+1,i}|C_{t}&\sim\text{BFRY}\left(\alpha'_t/K,\tau+\phi,\sigma - c_{ti}\right).
\end{align*}
where
\begin{align*}
    \alpha'_t = K(\sigma - c_{ti})(\sigma K/\alpha)^{\frac{c_{ti}-\sigma}{\sigma}}.
\end{align*}

This construction mirrors that of Section \ref{sec:sociability}, with the key difference being that we now obtain a stationary $\text{BFRY}\left(\alpha/K,\tau,\sigma\right)$ distribution for the $w_{ti}$. We can easily see this by noting that if
\begin{align*}
    w_{ti} &\sim \text{BFRY}\left(\alpha/K,\tau,\sigma\right)\\
    c_{ti}|w_{ti} &\sim\text{Poisson}(\phi w_{ti})
\end{align*}
then
\begin{align}\label{eq:t_conditional}
    p(w_{ti}|c_{ti}) \propto w_{ti}^{-1-\sigma+c_{ti}}e^{-(\tau + \phi)w_{ti}}\left( 1-e^{- (\sigma K/\alpha)^{1/\sigma}w_{ti}} \right)
\end{align}
which we then recognise as a $\text{BFRY}\left(\alpha'_t/K,\tau+\phi,\sigma - c_{ti}\right)$ with $\alpha'_t$ as above.

Thus, the use of a finite-dimensional approximation BFRY random variables gives us the simple conjugate construction that we desired. The reason that we introduce this non-standard distribution is that, as far as we know, it is not possible to approximate the generalised gamma process using a finite measure with i.i.d. gamma random weights. In the specific case $\sigma=0$, we could use the simpler finite approximation using $Gamma(\alpha / K, \tau)$ random variables. However, this would preclude us from modelling networks with power-law degree distributions, as discussed previously. 

\begin{figure*}[t]
\centering
\subfloat[$t=1$\label{fig:sim_post_pred_t_0}]{\includegraphics[width=0.25\textwidth]{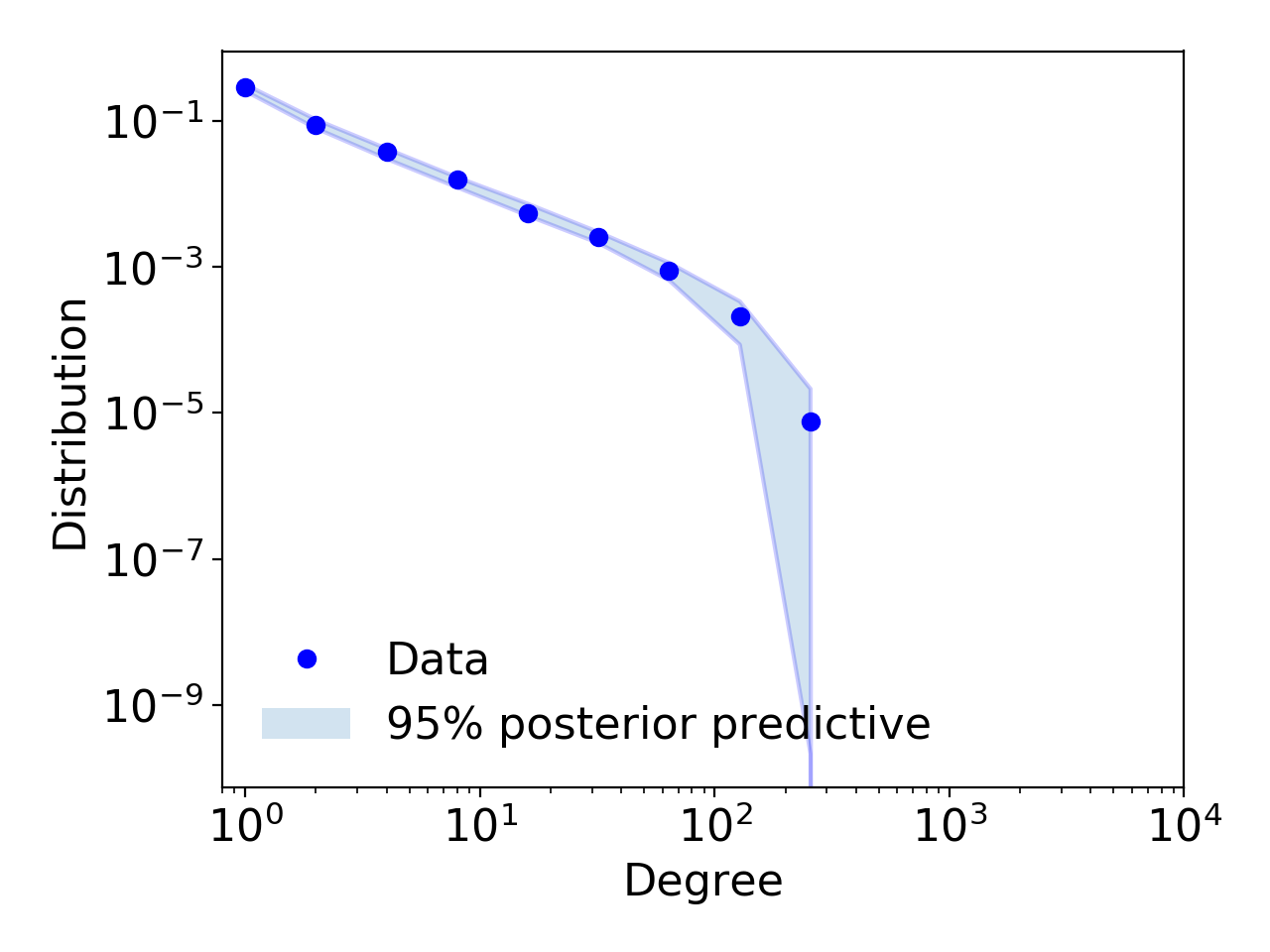}}
\subfloat[$t=2$ \label{fig:sim_post_pred_t_1}]{\includegraphics[width=0.25\textwidth]{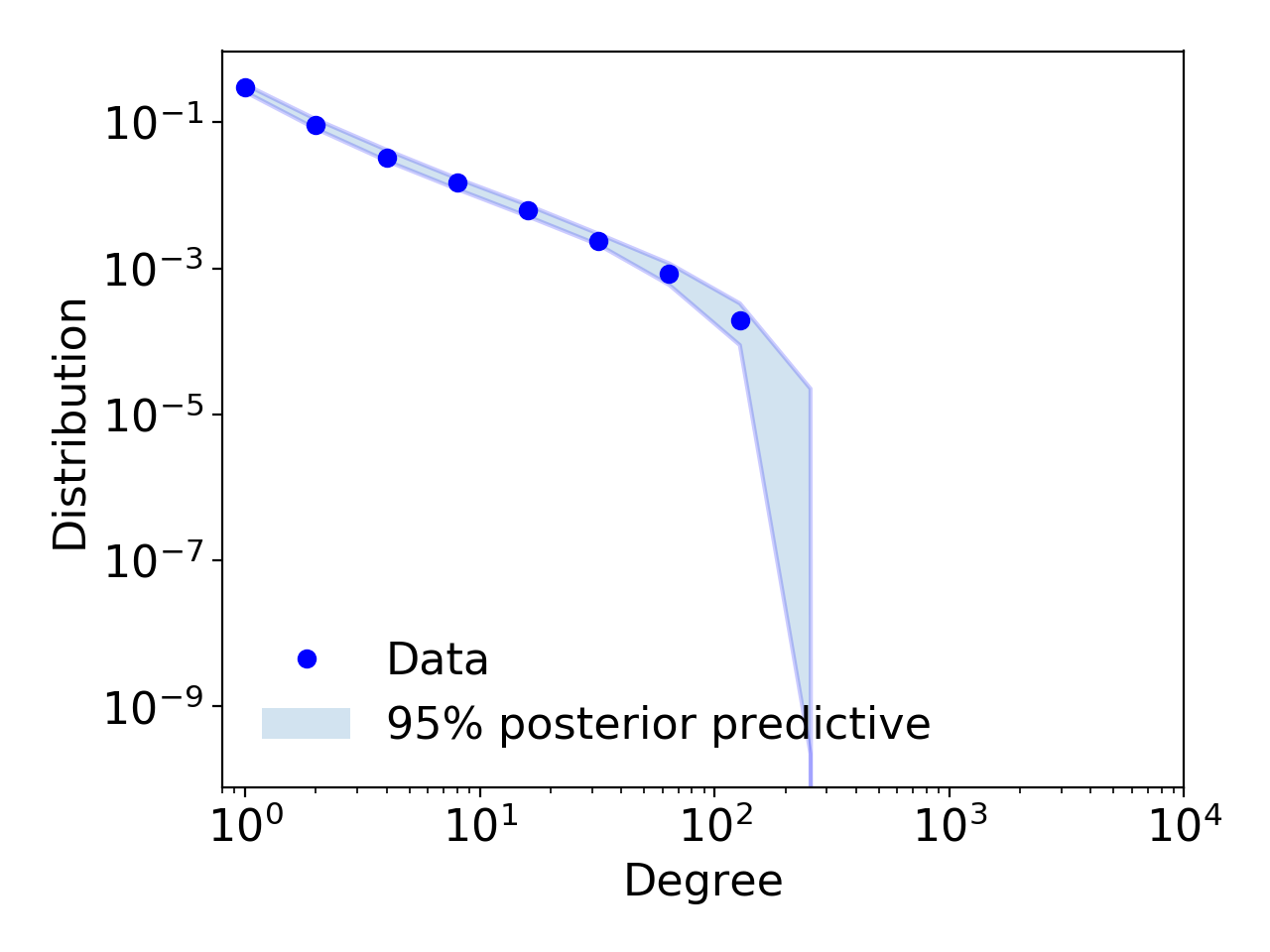}}
\subfloat[$t=3$ \label{fig:sim_post_pred_t_2}]{\includegraphics[width=0.25\textwidth]{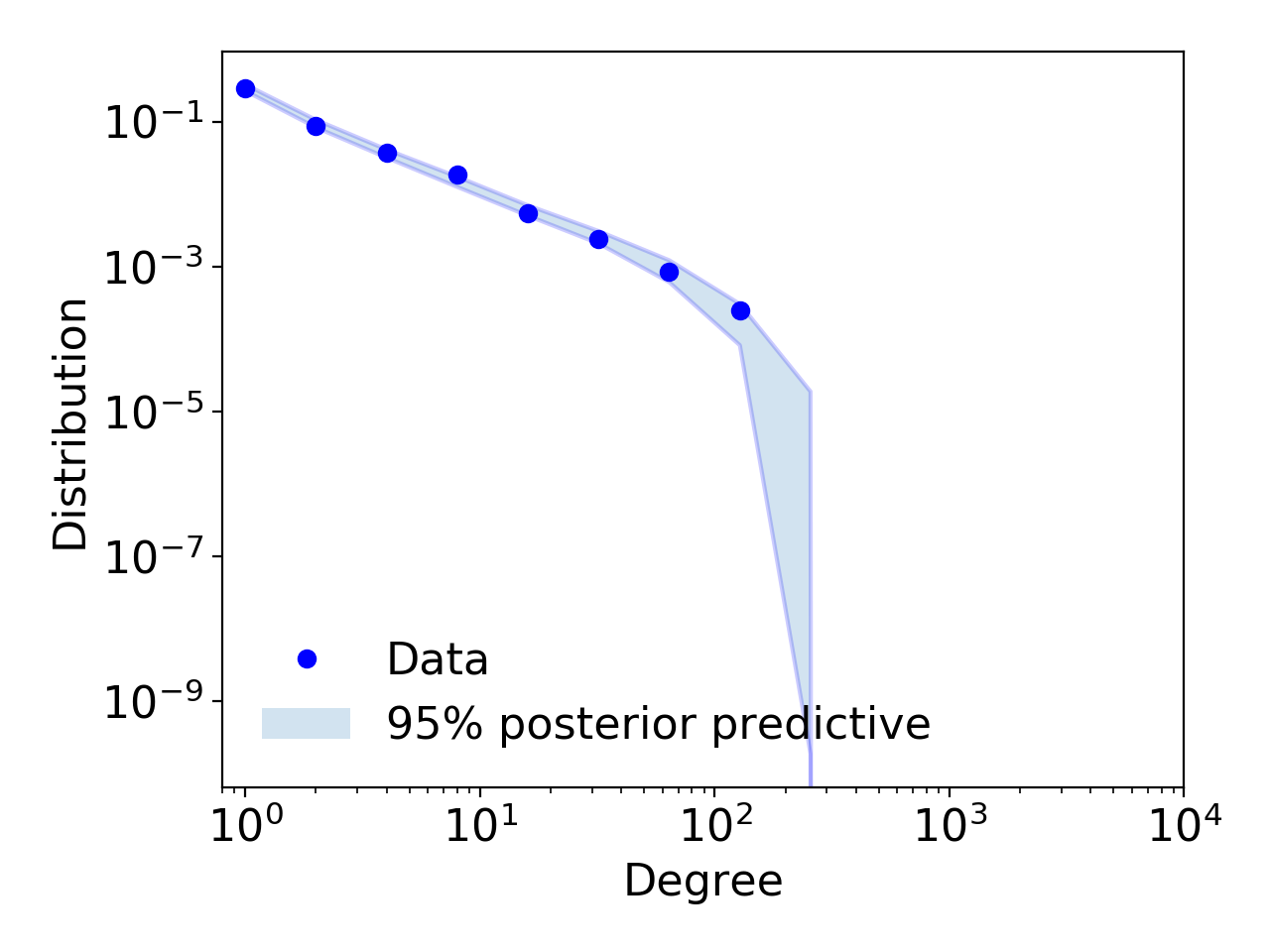}}
\subfloat[$t=4$ \label{fig:sim_post_pred_t_3}]{\includegraphics[width=0.25\textwidth]{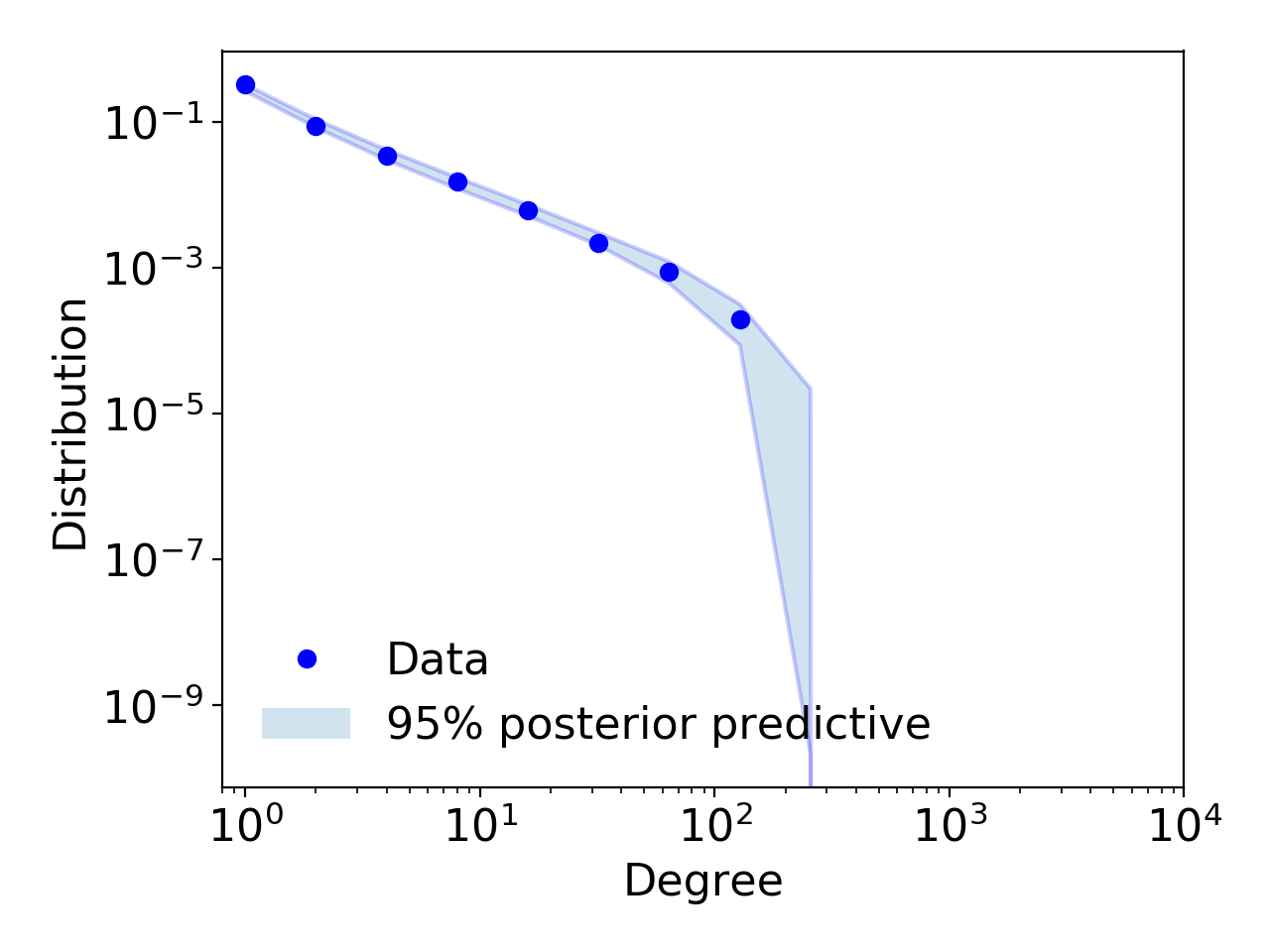}} \\
\subfloat[$t=1$\label{fig:sim_high_weights_t_0}] {\includegraphics[width=0.25\textwidth]{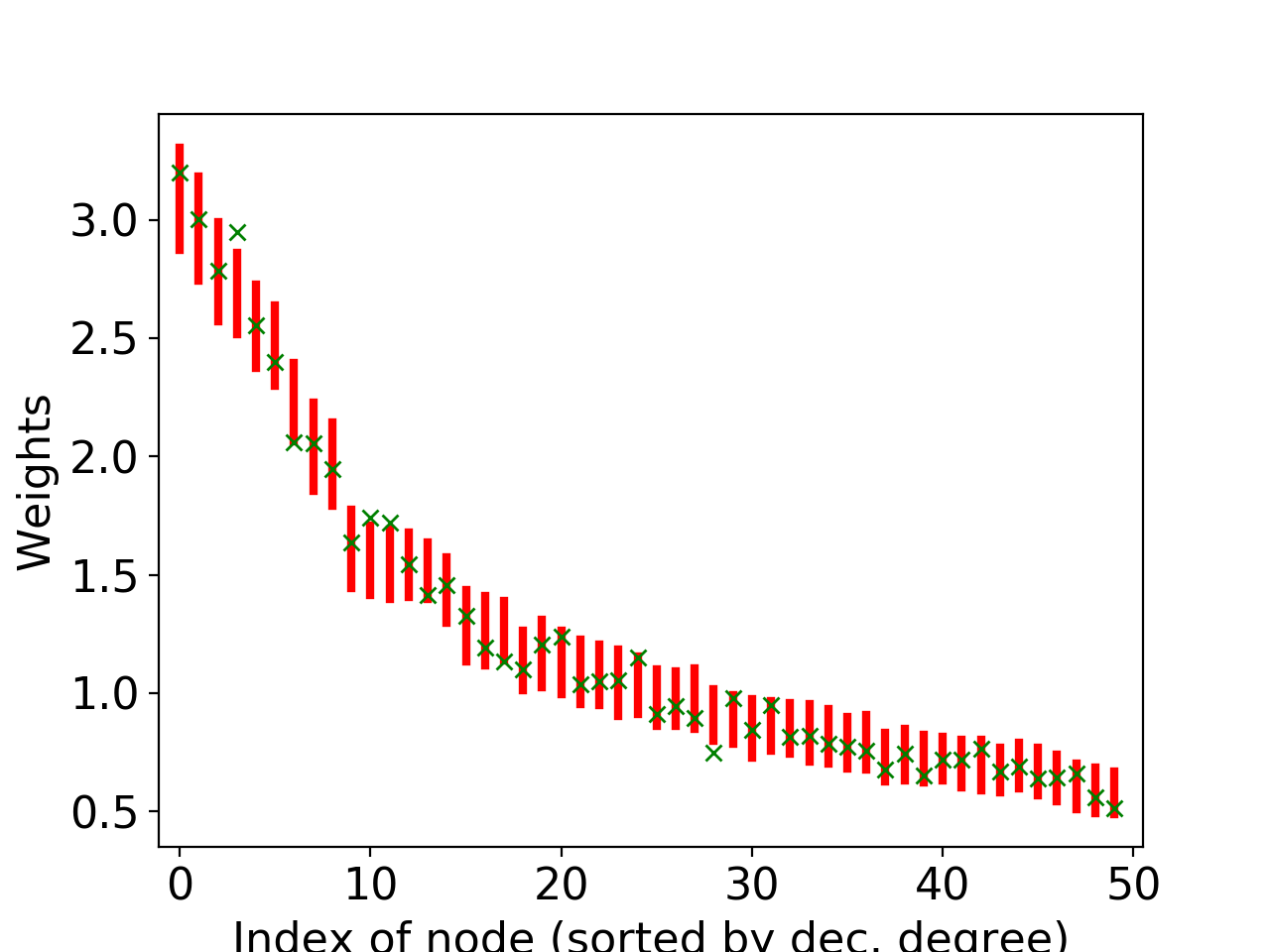}}
\subfloat[$t=2$ \label{fig:sim_high_weights_t_1}]{\includegraphics[width=0.25\textwidth]{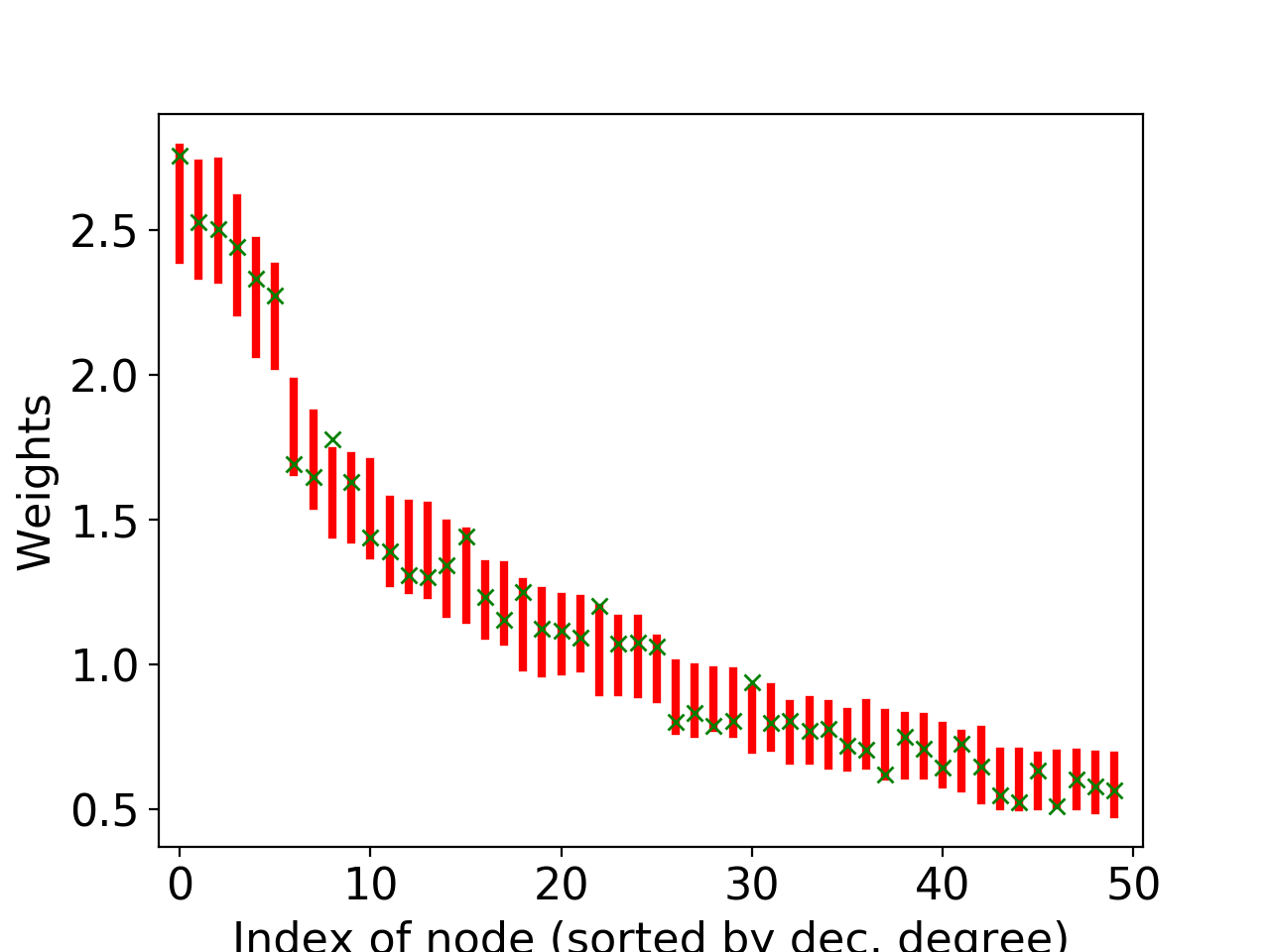}}
\subfloat[$t=3$ \label{fig:sim_high_weights_t_2}]{\includegraphics[width=0.25\textwidth]{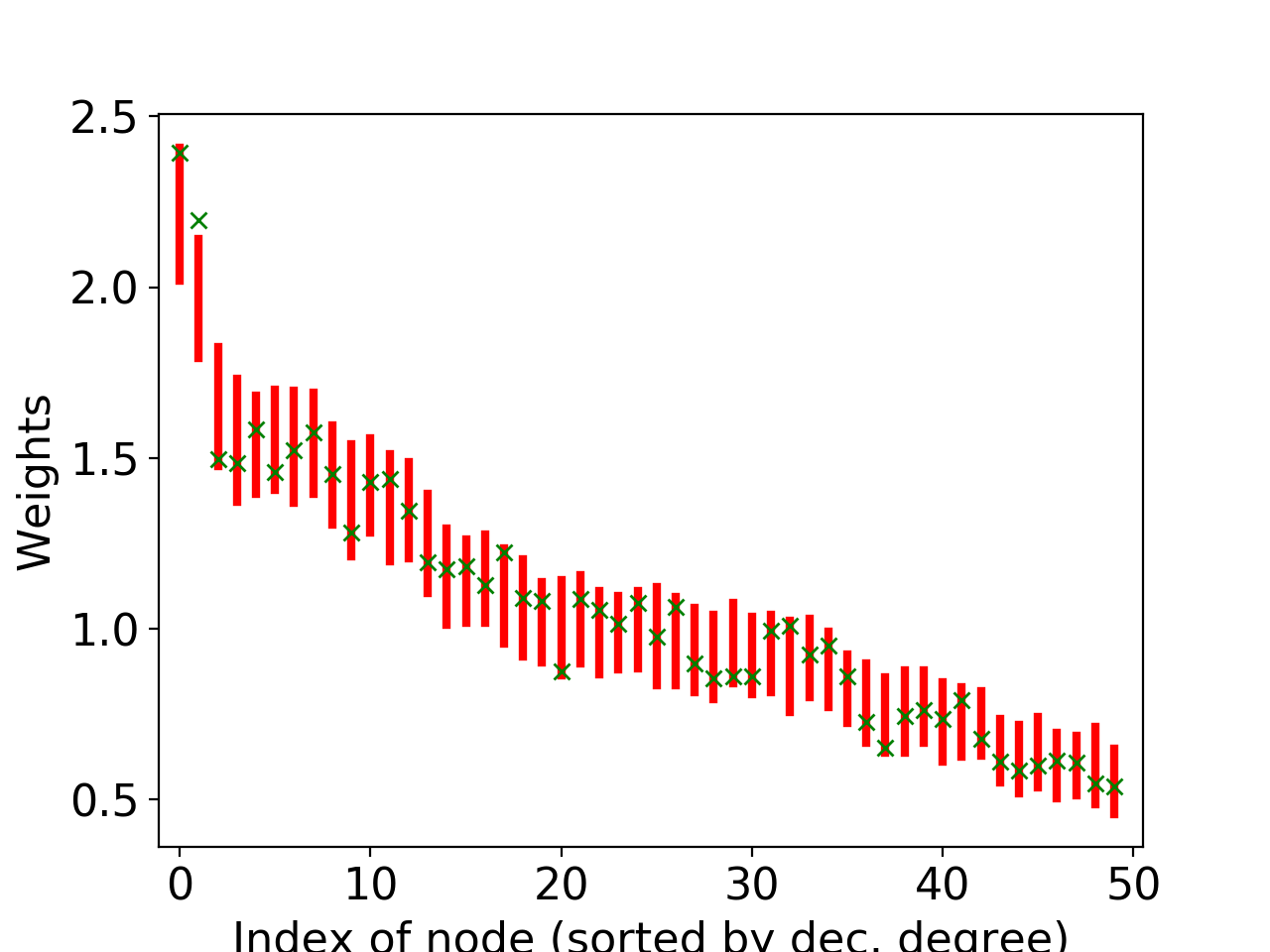}}
\subfloat[$t=4$ \label{fig:sim_high_weights_t_3}]{\includegraphics[width=0.25\textwidth]{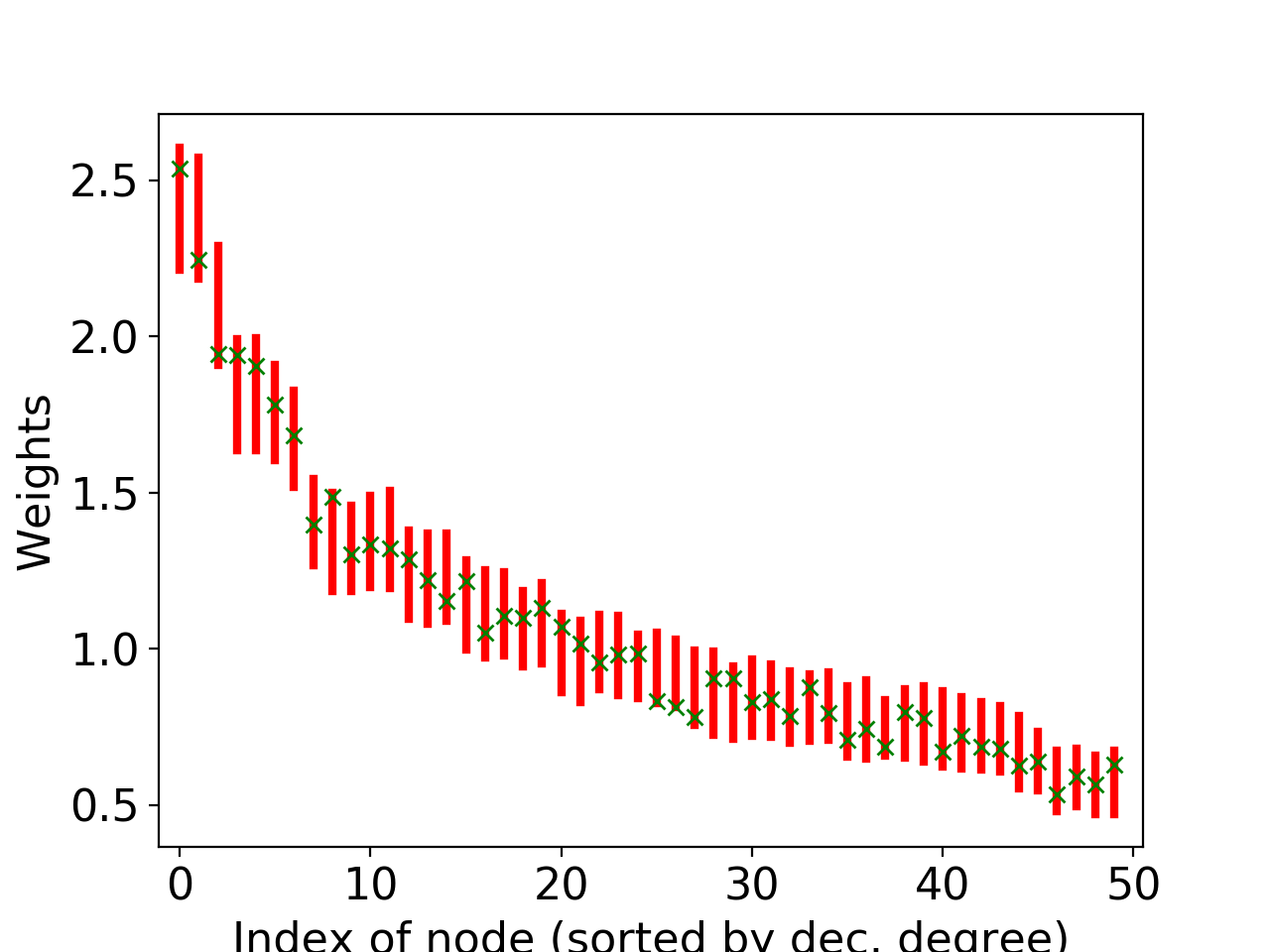}}

\caption{(Top row) Posterior Predictive Degree Distribution and (bottom row) 95\% credible intervals for sociabilites of the nodes with the highest degrees, for a network simulated from the GG model. True weights are represented by a green cross.}
\label{fig:ggp_simulation_post_pred}
\end{figure*}

\begin{figure}[t]
\centering
\label{fig:sim_high_weights}
\includegraphics[width=0.6\textwidth]{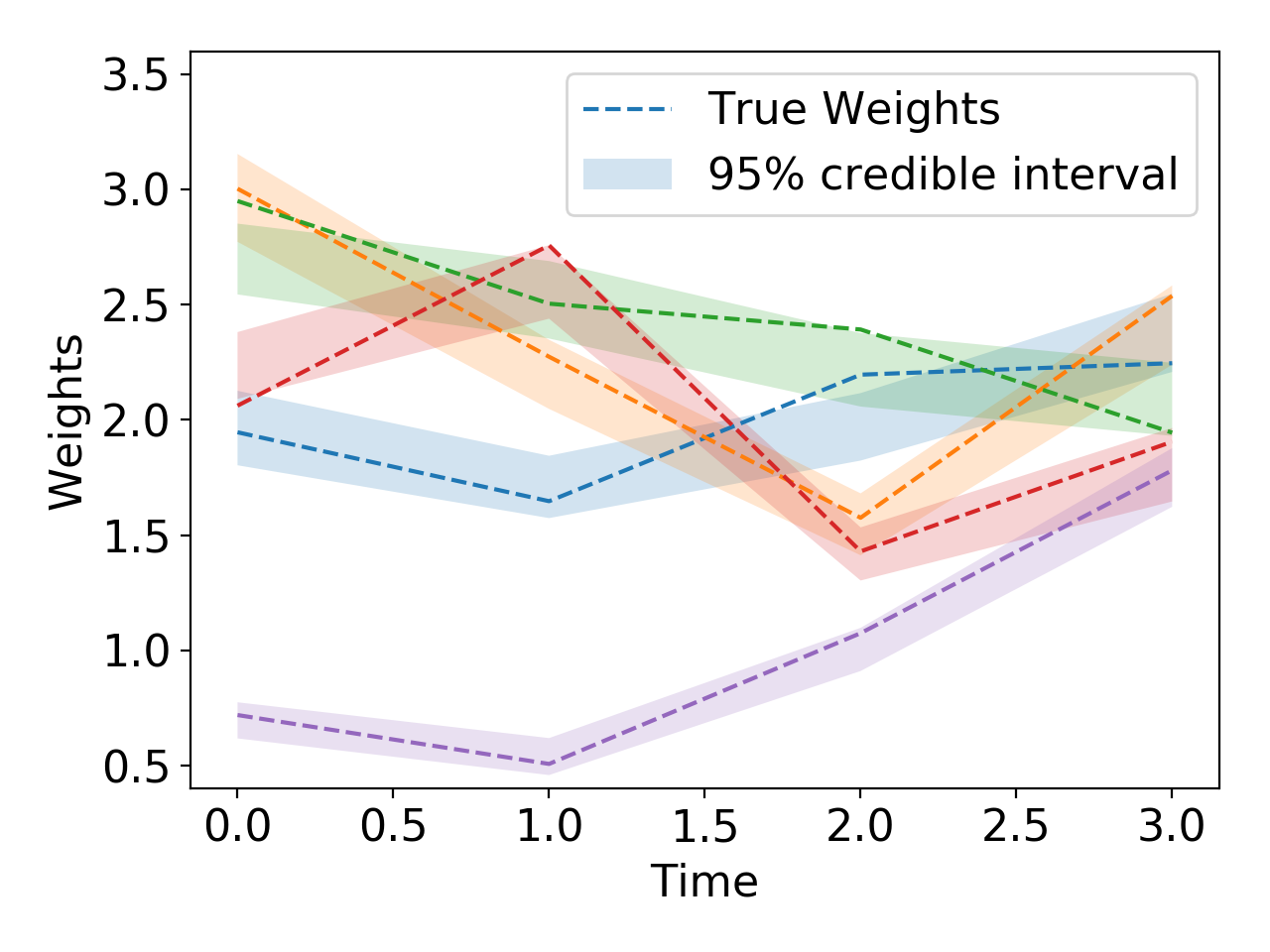}
\caption{Evolution of weights of high degree nodes for the network simulated from the GG model. The dotted line shows the true value of the weights in each case.}
\label{fig:sim_weights}
\end{figure}
\subsection{Posterior Inference Algorithm}
In order to perform posterior inference with the approximate method, we use a Gibbs sampler. We introduce auxiliary variables $\{u_{ti}\}^K_{i=1}, t=1,\ldots,T$ following a truncated exponential distribution. The overall sampler is as follows  (see details in Section \ref{sec:app:mcmc_alg} of the Appendix):

$1.$ Update the weights $w_{ti}$ given the rest using Hamiltonian Monte Carlo (HMC).\\
$2.$ Update the latent $c_{ti}$ given the rest using Metropolis Hastings (MH).\\
$3.$ Update the hyperparameters $\alpha$, $\sigma$, $\phi$ and $\tau$ and the latent variables $u_{ti}$ given the rest

For the hyperparameters, we place gamma priors on $\alpha$, $\tau$ and $\phi$, and a Beta prior on $\sigma$.

\section{Experiments}
\label{sec:experiments}
\begin{figure}[t]
\centering
\subfloat[$\sigma$\label{fig:sigma_comparison}]{\includegraphics[width=0.4\textwidth]{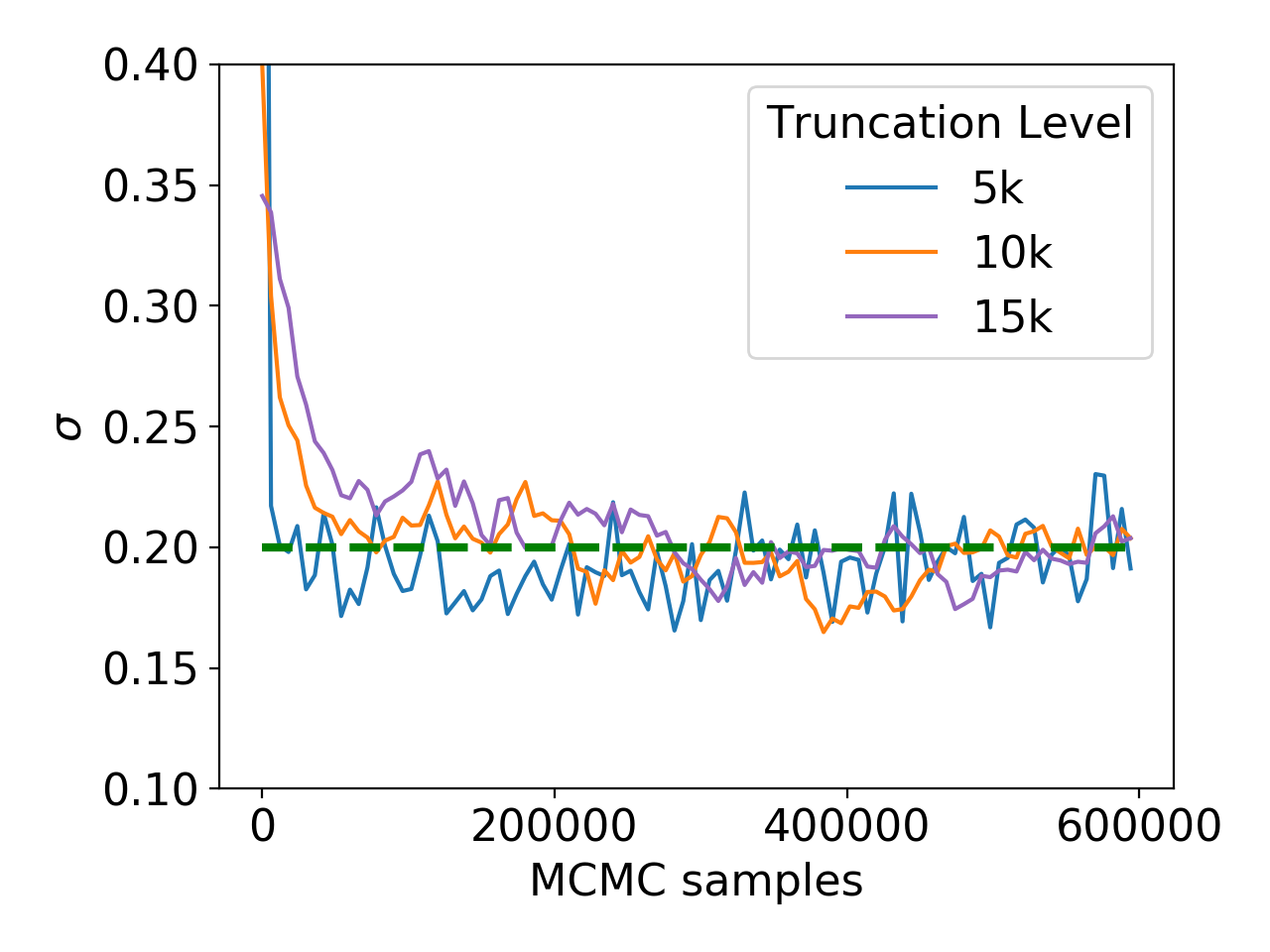}}
\subfloat[$\phi$ \label{fig:phi_comparison}]{\includegraphics[width=0.4\textwidth]{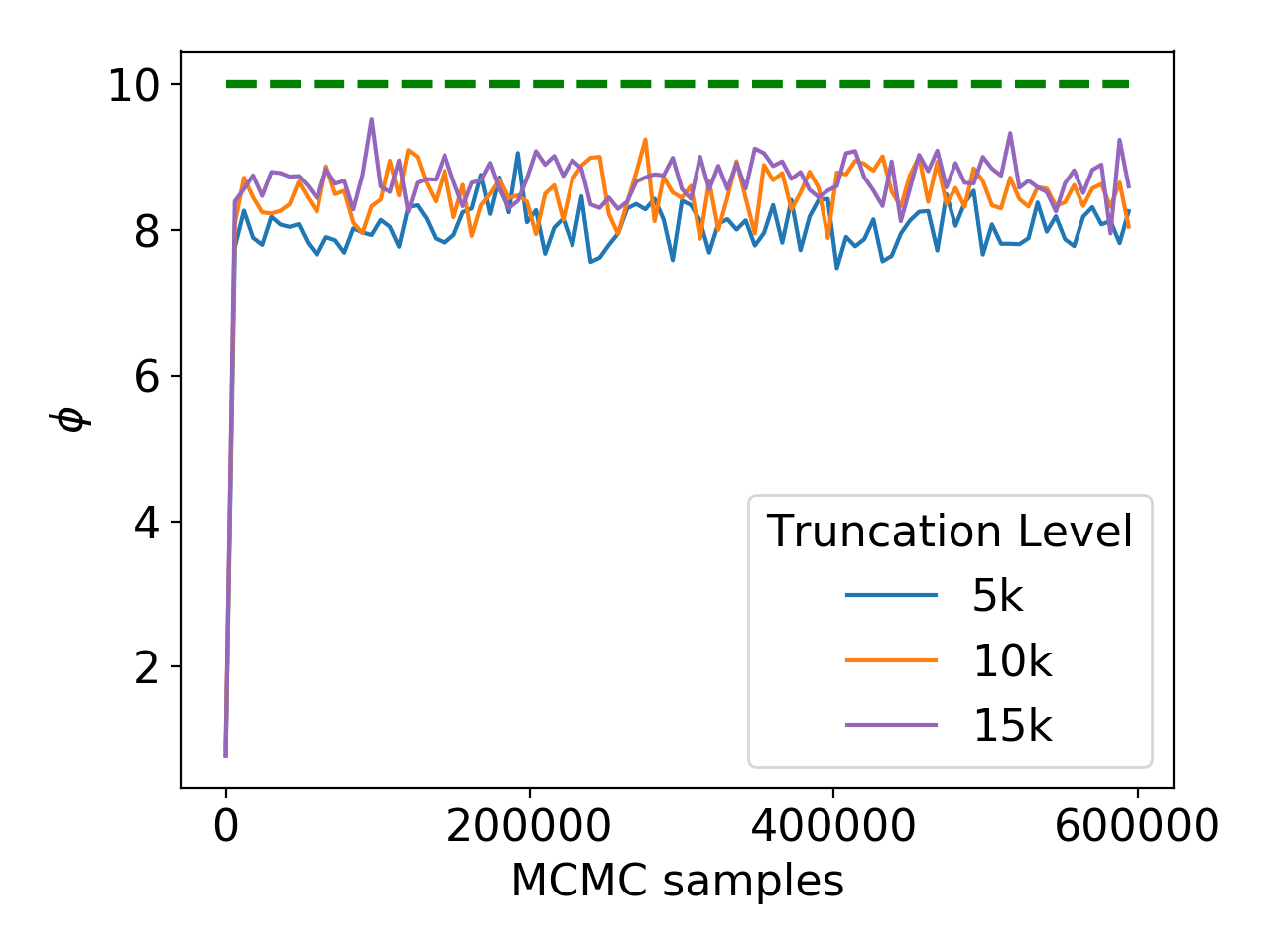}}\\
\subfloat[$\tau$\label{fig:tau_comparison}]{\includegraphics[width=0.4\textwidth]{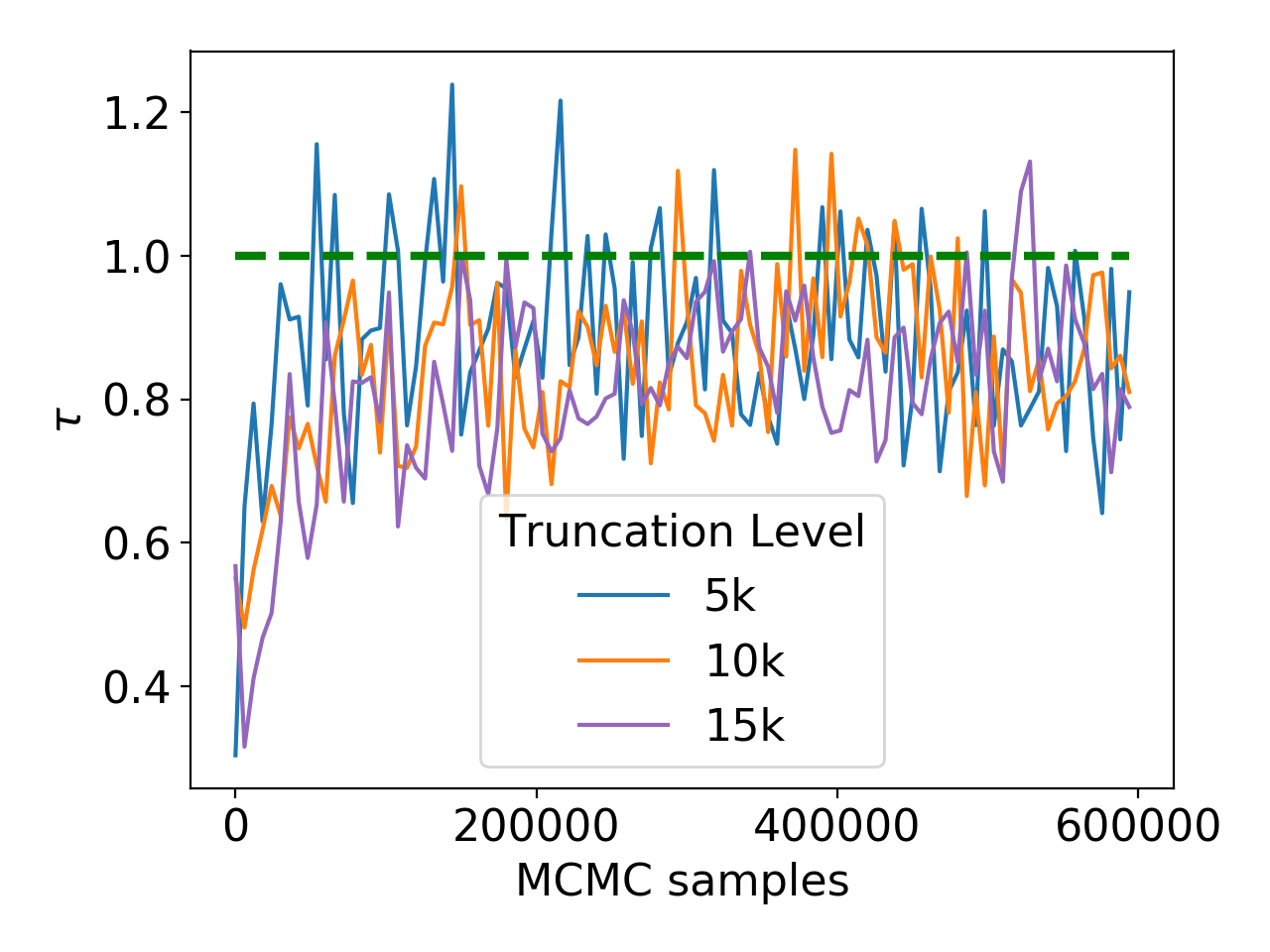}}
\subfloat[$\alpha$ \label{fig:alpha_comparison}]{\includegraphics[width=0.4\textwidth]{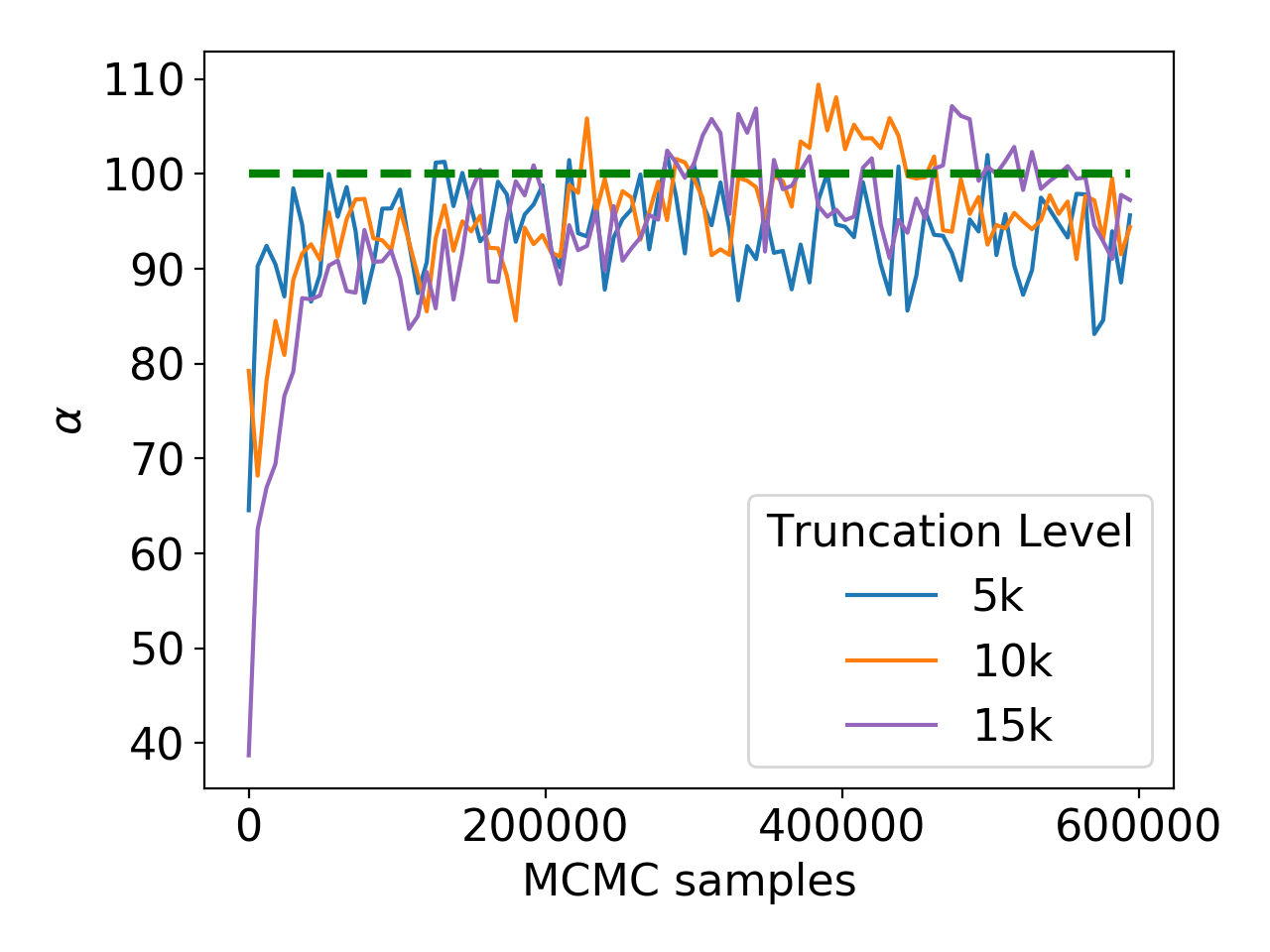}}
\caption{Trace plots for (a) $\sigma$, (b) $\phi$, (c) $\tau$ and (b) $\alpha$ for different truncation levels, when fitting to a network simulated from the GG model}
\label{fig:sim_truncation_comparison}
\end{figure}
\subsection{Simulated Data}
In order to assess the approximate inference scheme, we first consider a synthetic dataset. We simulate a network with $T=4$, $\alpha = 100$, $\sigma = 0.2$, $\tau = 1$, $\phi = 10$ from the exact model (see Section~\ref{sec:model}), and then estimate it using our approximate inference scheme described in Section~\ref{sec:inference}. The generated network has $3,096$ nodes and $101,571$ edges. We run an MCMC chain of length $600,000$, with the first $300,000$ samples discarded as burn-in. We set the truncation threshold to $K=15,000$.

The approximate inference scheme estimates the model's parameters well. This can be seen in terms of the fit of the posterior predictive degree distribution to the empirical as seen in Figures \ref{fig:sim_post_pred_t_0}-\ref{fig:sim_post_pred_t_3}, and the coverage of the credible intervals for the weights, as we see in Figures \ref{fig:sim_high_weights_t_0}-\ref{fig:sim_high_weights_t_3} for the 50 nodes with the highest degree, and in Figure~\ref{fig:sim_weights} for the evolution over time of some high degree nodes.

\begin{figure*}[t]
\centering
\subfloat[$t=2$ \label{fig:reddit_post_pred_t_1}]{\includegraphics[width=0.25\textwidth]{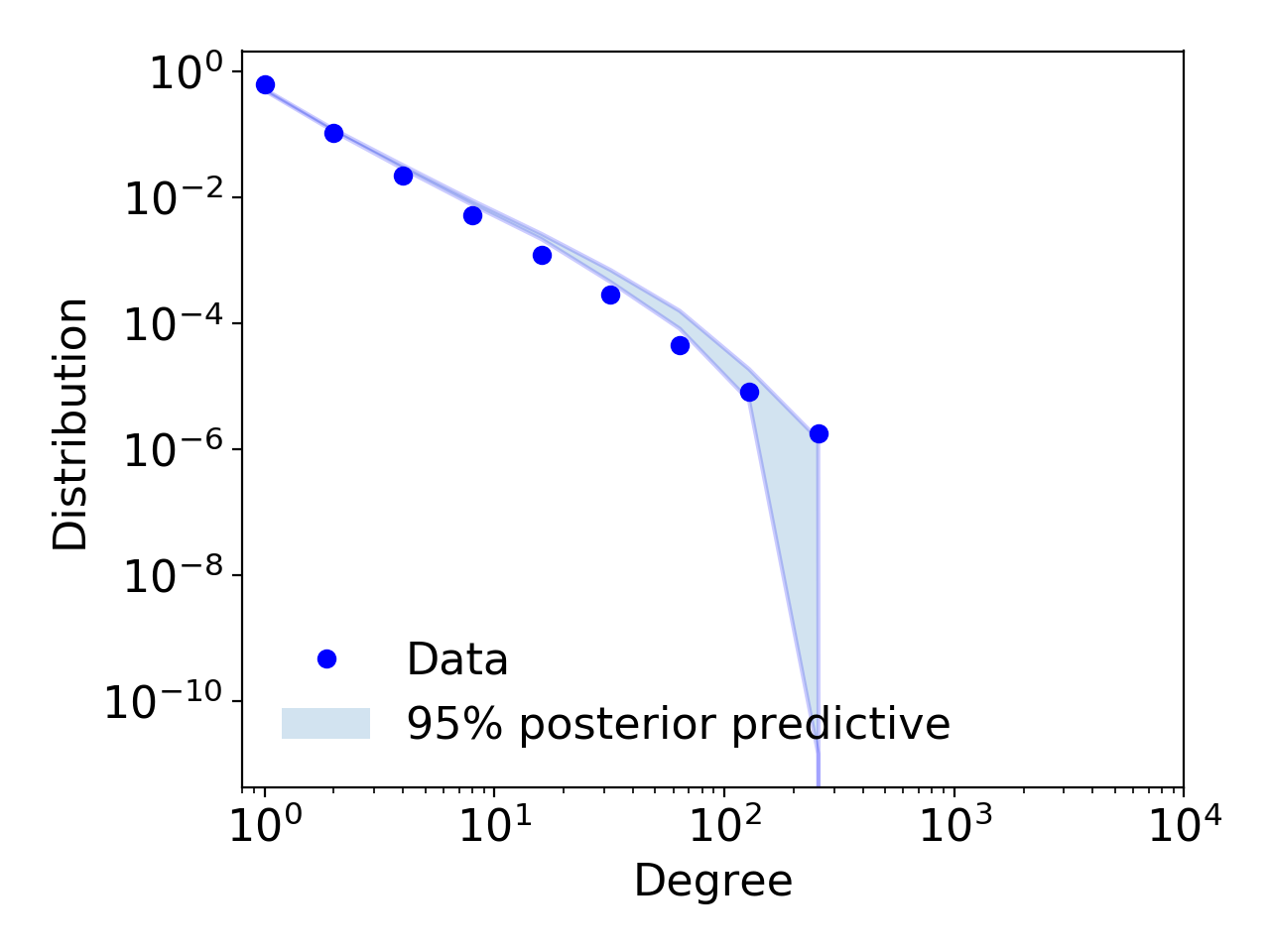}}
\subfloat[$t=6$ \label{fig:reddit_post_pred_t_5}]{\includegraphics[width=0.25\textwidth]{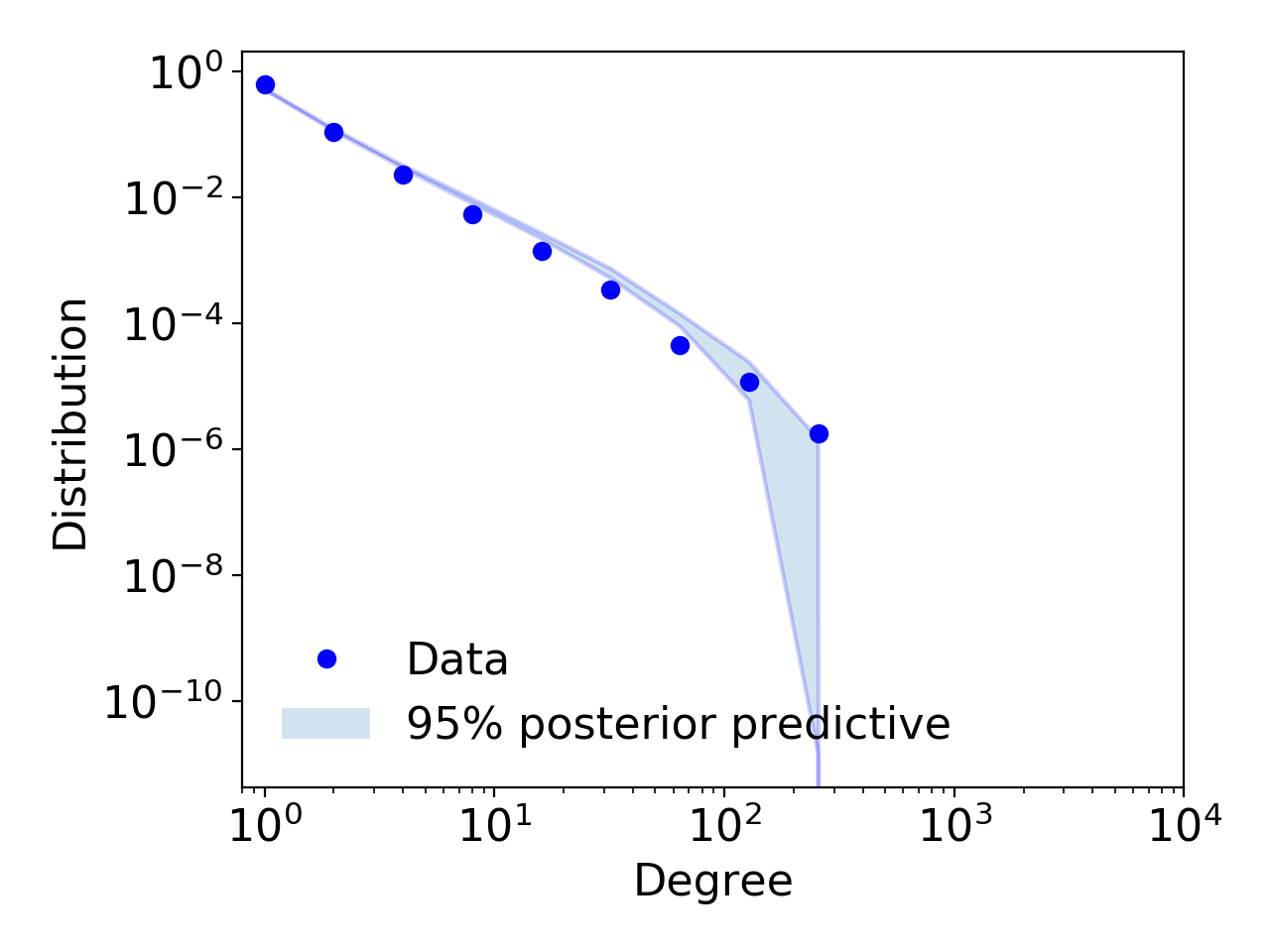}}
\subfloat[$t=10$ \label{fig:reddit_post_pred_t_9}]{\includegraphics[width=0.25\textwidth]{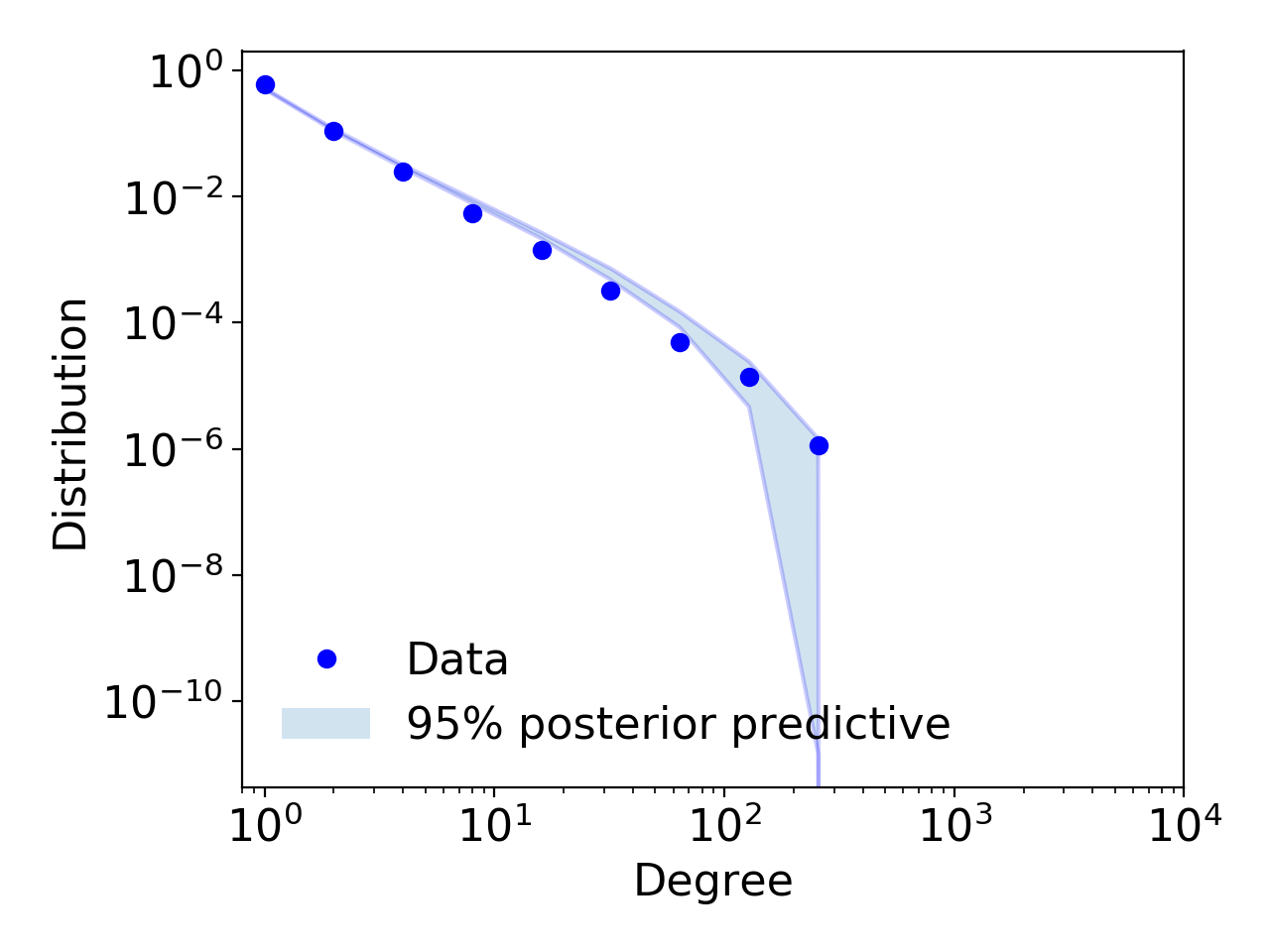}}
\subfloat[$t=12$ \label{fig:reddit_post_pred_t_11}]{\includegraphics[width=0.25\textwidth]{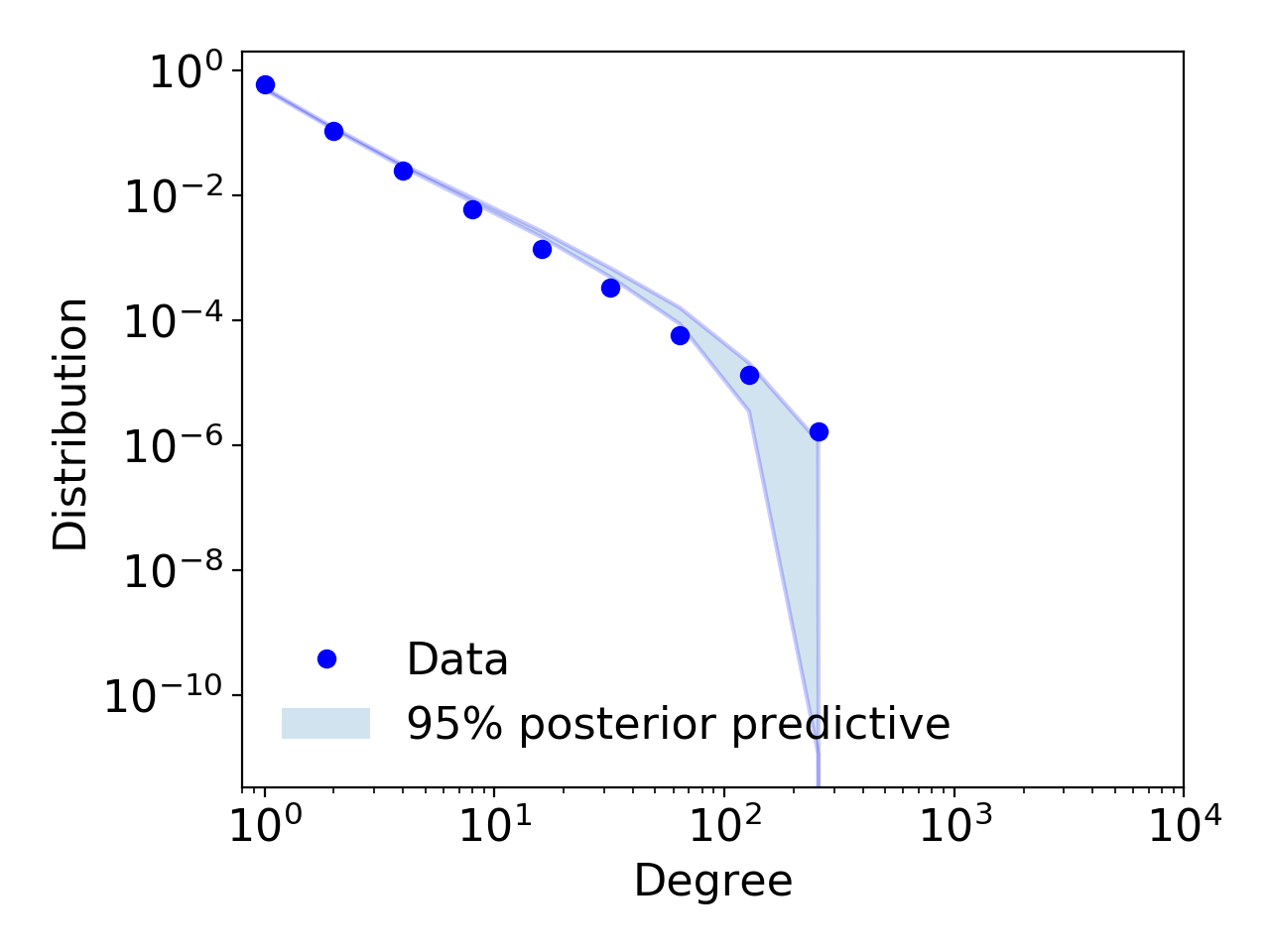}}
\caption{Posterior Predictive Degree Distribution over time for the Reddit hyperlink network}
\label{fig:reddit_post_pred}
\end{figure*}

To see the influence of the truncation level $K$, we rerun the algorithm for different truncation levels $K=5,000$ and $K=10,000$. Trace plots of the hyperparameters are shown in Figure \ref{fig:sim_truncation_comparison}. The approximate posterior distributions are quantitatively similar, but we see that increasing the value of $K$ leads to a more correlated Markov chain. However, we note that $\phi$ is slightly under-estimated in our model, and this problem is more severe the lower the truncation level is taken to be. Further comparison is given in Section \ref{sec:app:mcmc_plots} of the Appendix.

\subsection{Real Data}

We illustrate the use of our model on two more dynamic network datasets: the Reddit Hyperlink Network \citep{kumar2018community}\footnote{\url{https://snap.stanford.edu/data/soc-RedditHyperlinks.html}} and the Reuters Terror dataset\footnote{\url{http://vlado.fmf.uni-lj.si/pub/networks/data/CRA/terror.htm}}.
\subsubsection{Reddit Hyperlink Network}

\begin{figure}[ht]
\centering
\subfloat[Degree evolution of high degree nodes\label{fig:reddit_high_degrees}] {\includegraphics[width=0.4\textwidth]{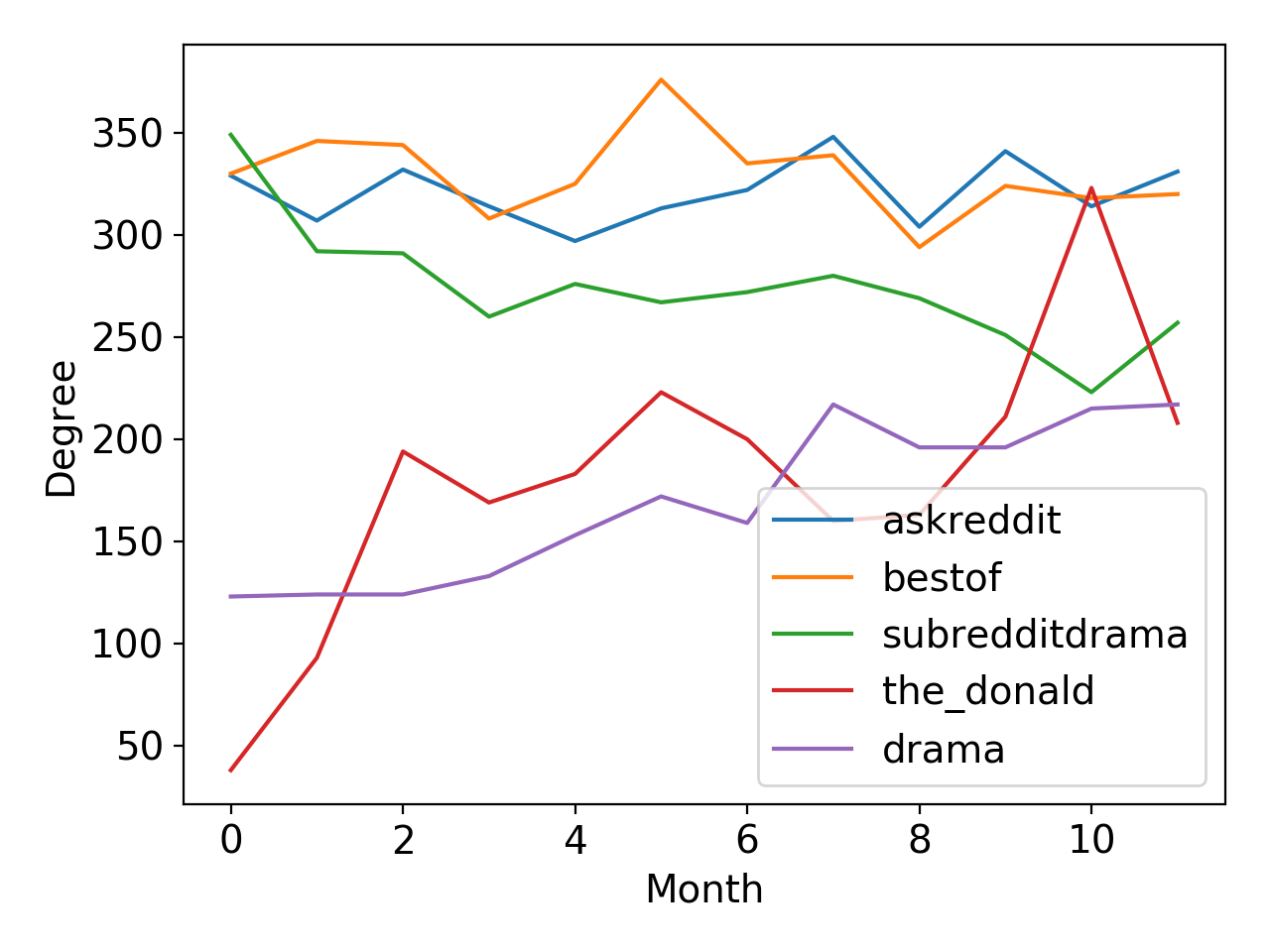}}
\hspace{0.01\textwidth}
\subfloat[Weight evolution of high degree nodes \label{fig:reddit_high_weights}]{\includegraphics[width=0.4\textwidth]{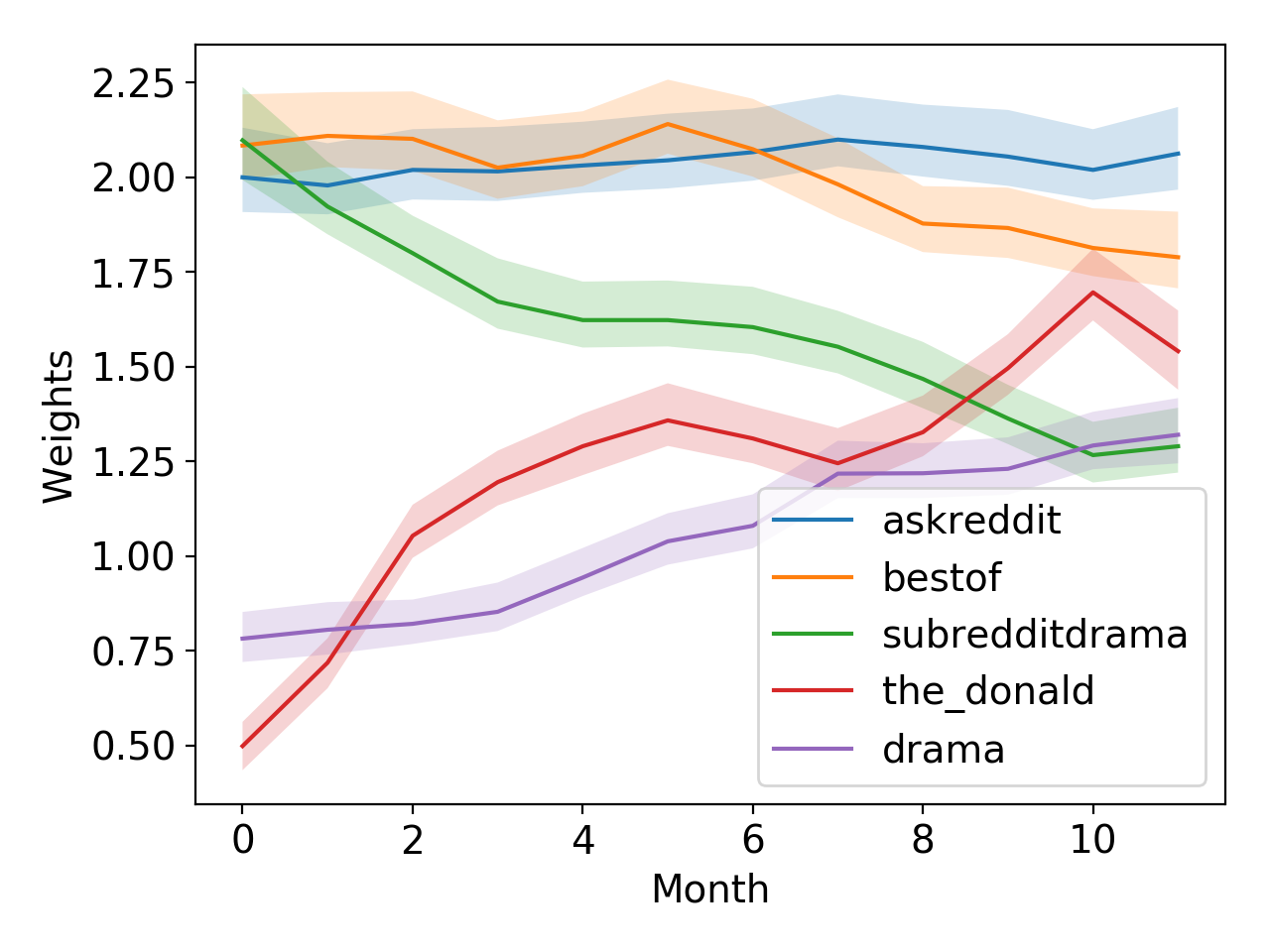}}
\caption{Evolution of (a) degrees and (b) weights of high degree nodes for the Reddit hyperlink network}
\label{fig:reddit_weights}
\end{figure}

\begin{figure*}[ht]
\centering
\subfloat[Weights evolution of the subreddits `the\_donald' and `sandersforpresident'\label{fig:reddit_word_weights_0}] {\includegraphics[width=0.3\textwidth]{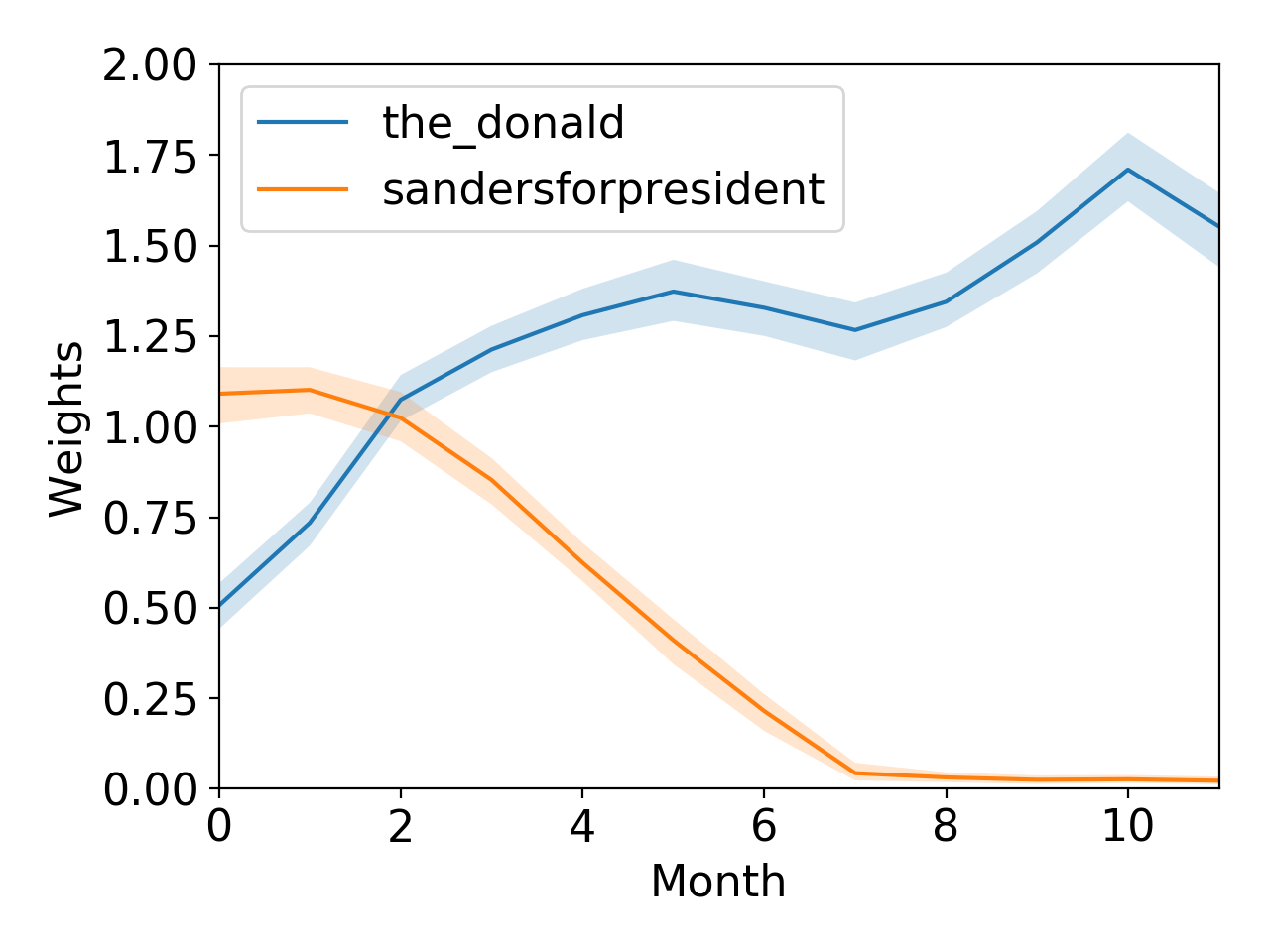}}
\hspace{0.01\textwidth}
\subfloat[Weights evolution of the subreddits `worldnews' and `politics'\label{fig:reddit_word_weights_1}]{\includegraphics[width=0.3\textwidth]{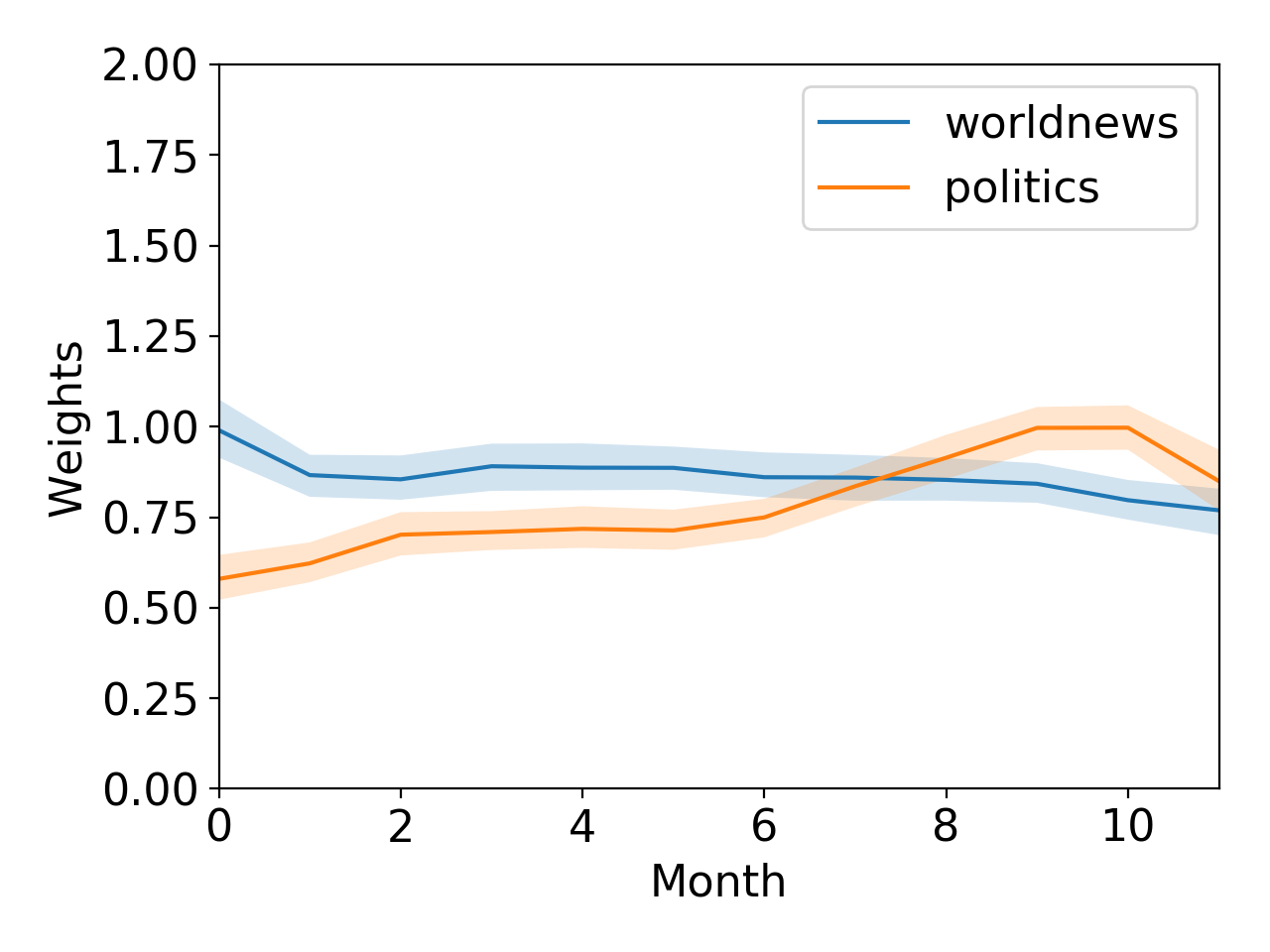}}
\hspace{0.01\textwidth}
\subfloat[Weights evolution of the subreddits `the\_donald' and `enoughtrumpspam'\label{fig:reddit_word_weights_2}]{\includegraphics[width=0.3\textwidth]{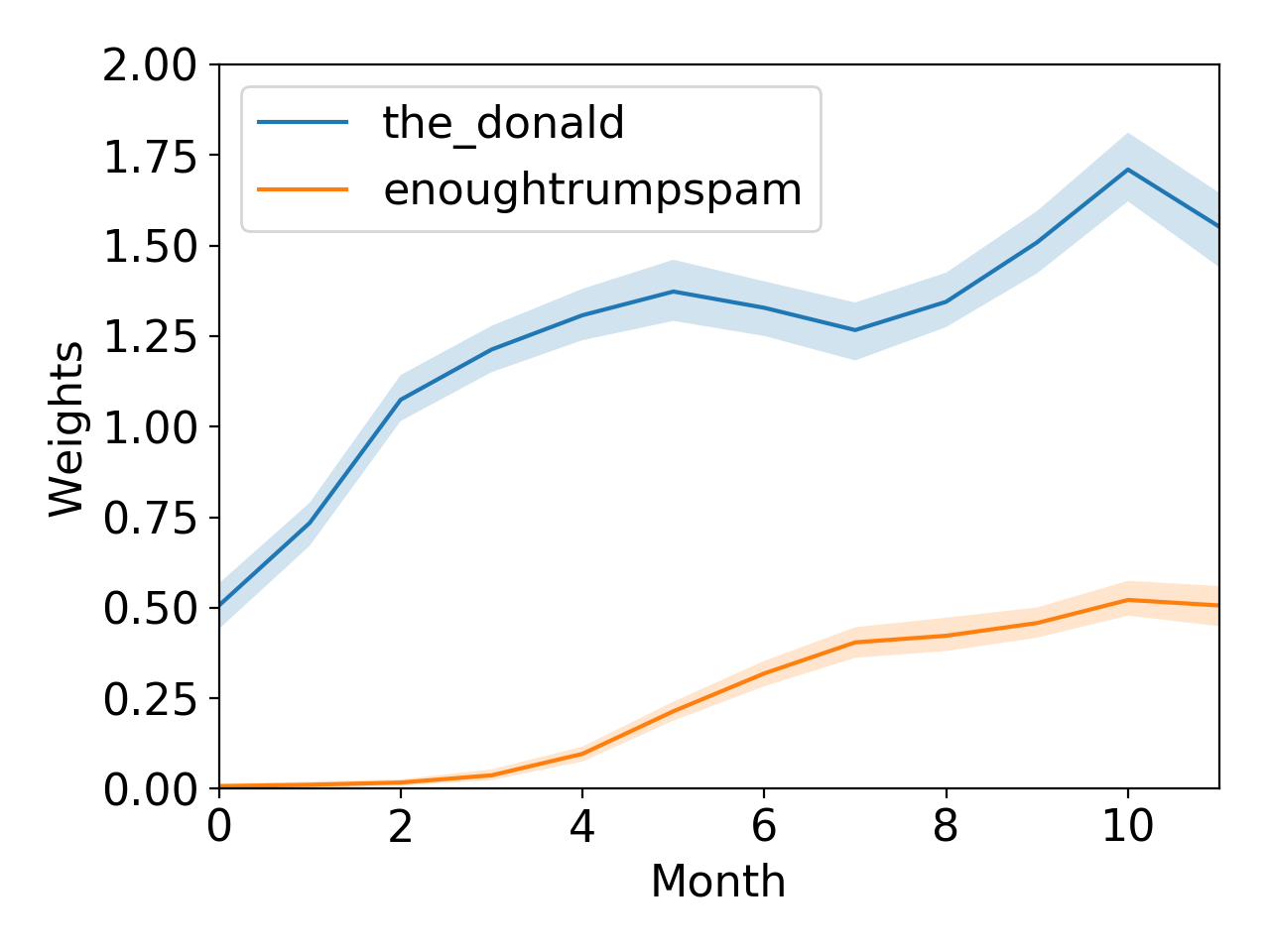}}
\caption{Evolution of the weights, for the Reddit hyperlink network.}
\label{fig:reddit_subreddit_weights}
\end{figure*}
The Reddit hyperlink network represents hyperlink connections between subreddits (communities on Reddit) over a period of $T=12$ consecutive months in $2016$. Nodes are subreddits (communities) and the symmetric edges represent hyperlinks originating in a post in one community and linking to a post in another community. The network has $N= 28,810$ nodes and $388,574$ interactions. The observations here are hyperlinks between the pair of subreddits $i,j$ at time $t$. The dataset has been made symmetric by placing an edge between nodes $i$ and $j$ if there is a hyperlink between them in either direction. We also assume that there are no loops in the network, that is $n_{tij} = 0$ for $i=j$. We run the Gibbs sampler $400,000$ samples, with the first $200,000$ discarded as burn in. In this case, we choose a truncation level of $K=40,000$.

From Figure \ref{fig:reddit_post_pred} we see that our model is capturing the empirical degree distribution well. Furthermore, in Figure \ref{fig:reddit_weights} we see that the model is able to capture the evolution of weights associated with each subreddit in a fashion that agrees with the observed frequency of interactions. The high degree nodes here are interpretable as either communities with a very large number of followers - such as ``askreddit'' or ``bestof'', or communities which frequently link to others - such as ``drama'' or ``subredditdrama''.

In particular, we see in Figure \ref{fig:reddit_word_weights_0} that the weights of the controversial but popular political subreddit ``The Donald'' increases to a peak in November, corresponding to the 2016 U.S. presidential election. Conversely, the weights of the subreddit ``Sandersforpresident'' decrease as the year goes on, corresponding to the end of Senator Bernie Sanders' presidential campaign, a trend that again agrees with the evolution of the corresponding observed degrees.

\subsubsection{Reuters Terror Dataset}

\begin{figure*}[ht]
\centering
\subfloat[$t=1$\label{fig:reuters_post_pred_t_0}] {\includegraphics[width=0.25\textwidth]{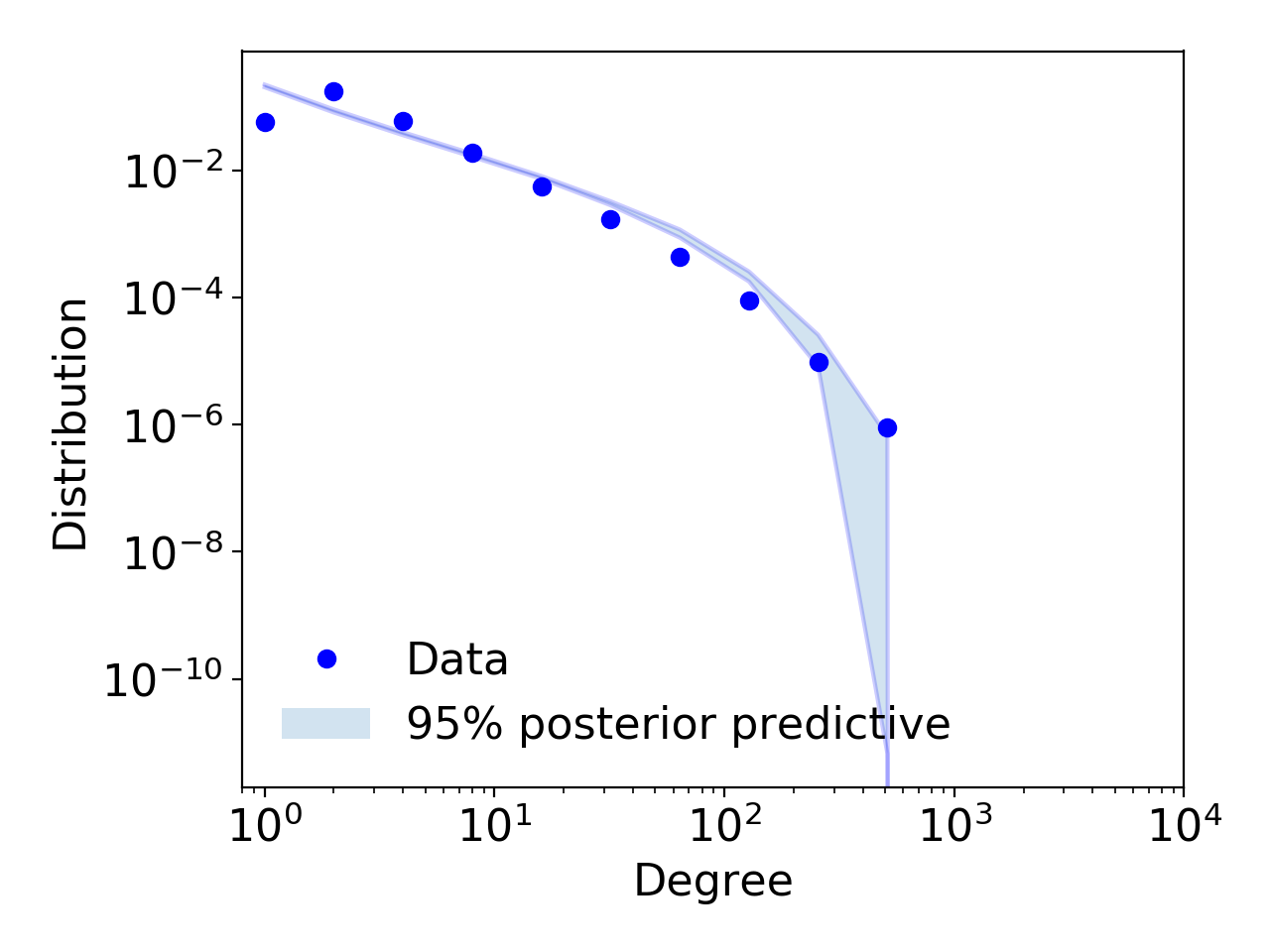}}
\subfloat[$t=3$ \label{fig:reuters_post_pred_t_2}]{\includegraphics[width=0.25\textwidth]{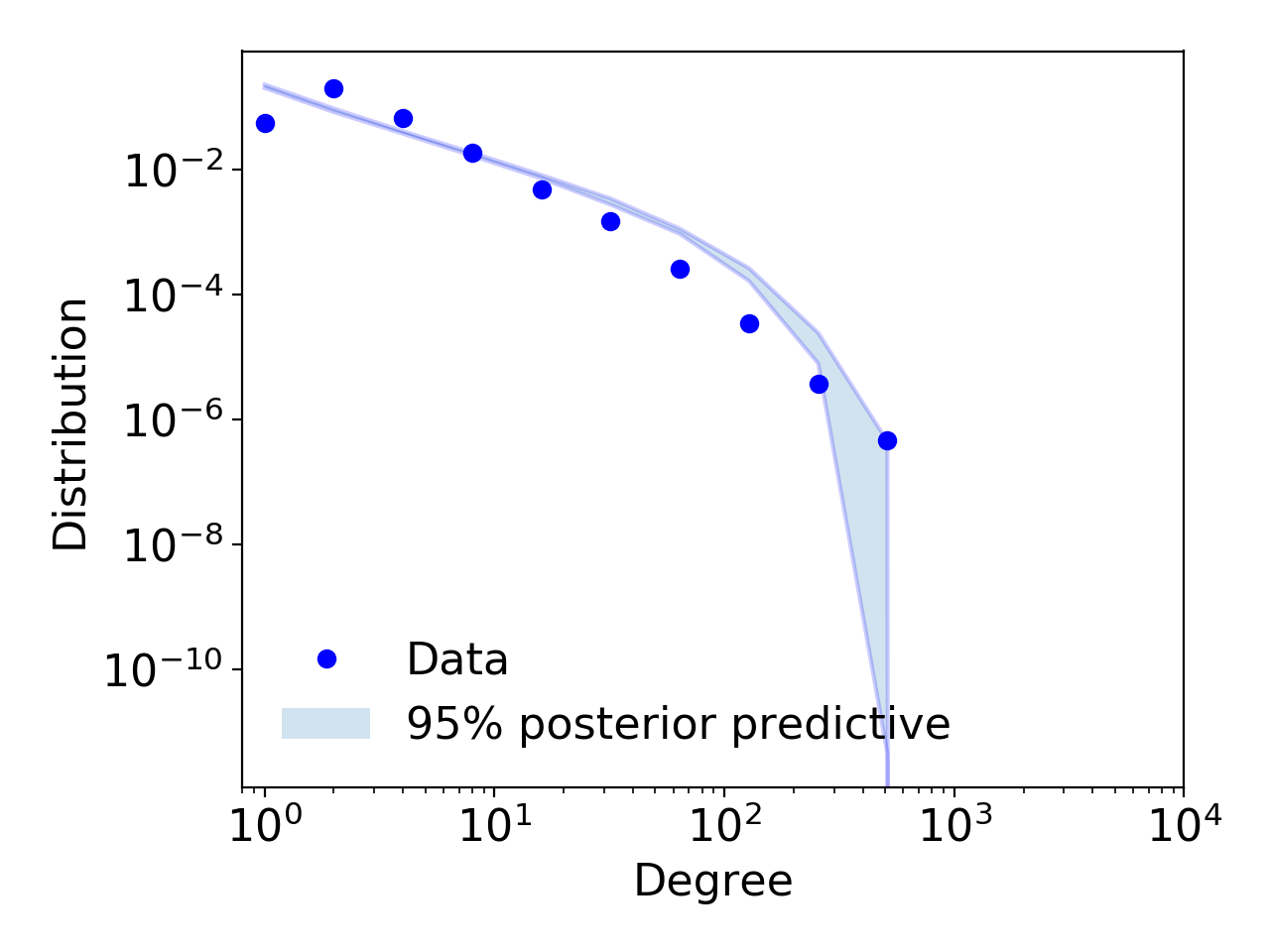}}
\subfloat[$t=5$ \label{fig:reuters_post_pred_t_3}]{\includegraphics[width=0.25\textwidth]{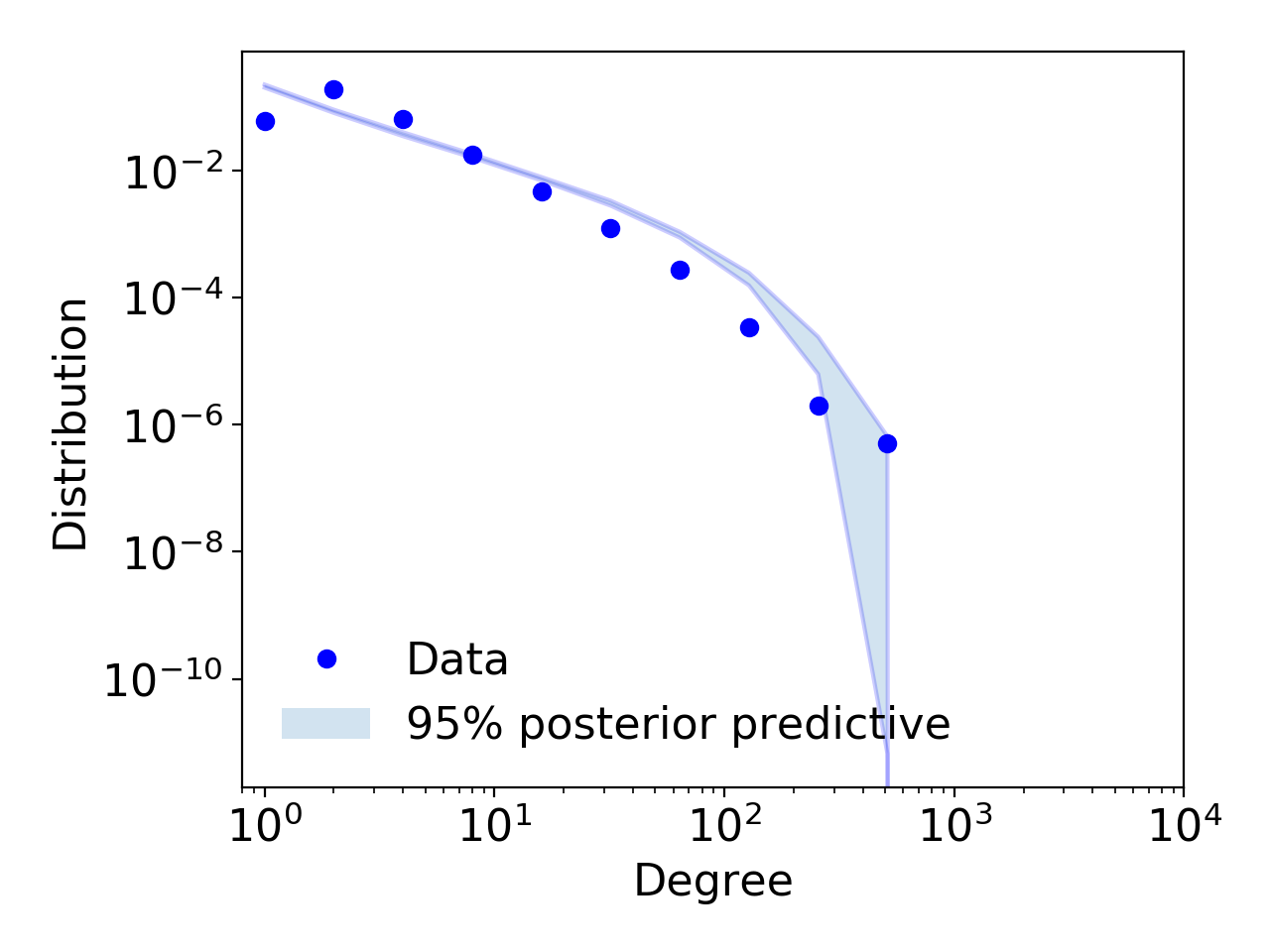}}
\subfloat[$t=7$ \label{fig:reuters_post_pred_t_4}]{\includegraphics[width=0.25\textwidth]{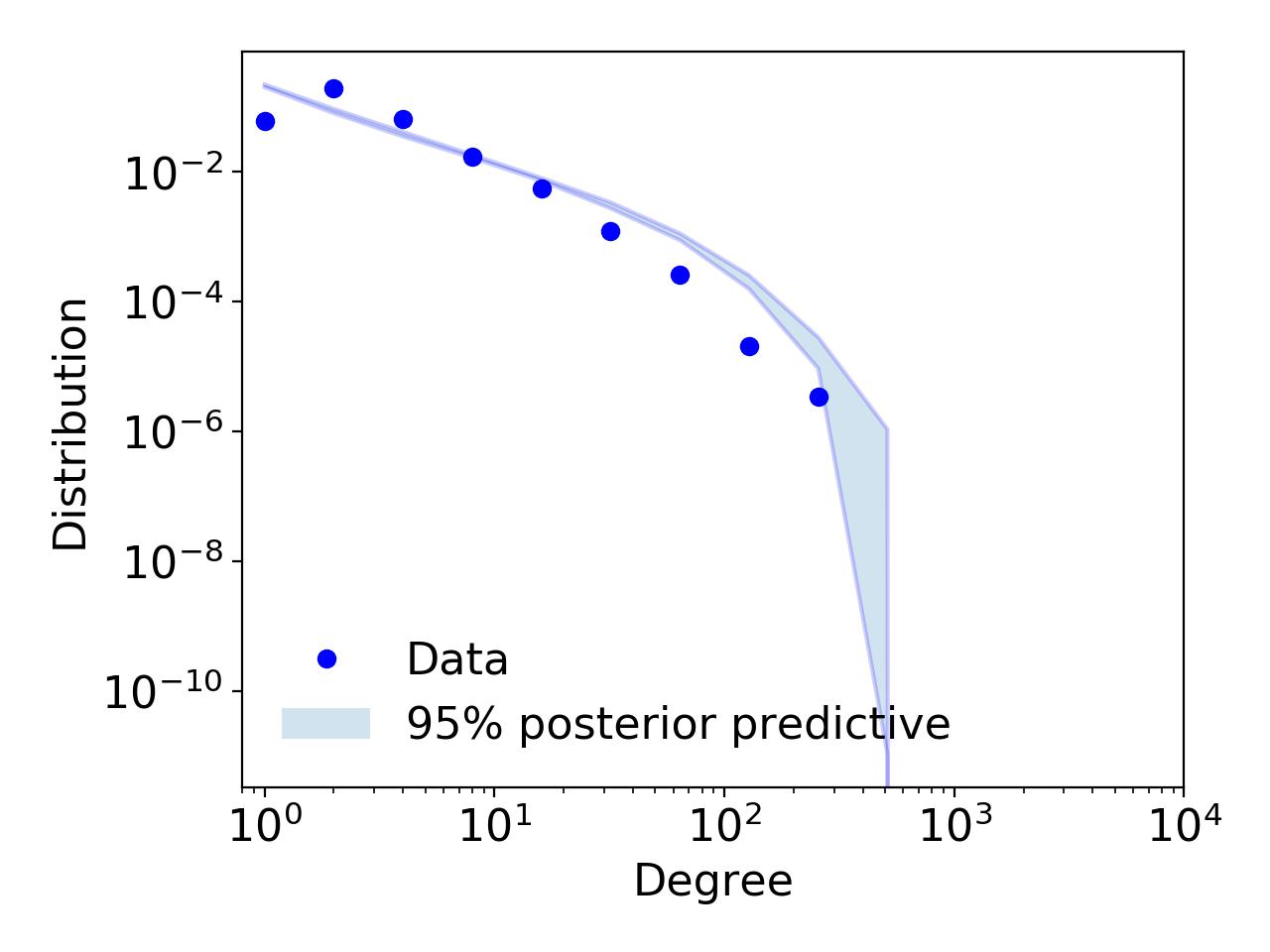}}
\caption{Posterior Predictive Degree Distribution for the Reuters terror dataset}
\label{fig:reuters_post_pred}
\end{figure*}
The final dataset we consider is the Reuters terror news network dataset. It is based on all stories released during $T=7$ consecutive weeks (the original data was day-by-day, but was shortened and collated for our purposes) by the Reuters news agency concerning the $09/11/01$ attack on the U.S.. Nodes are words and edges represent co-occurence of words in a sentence, in news. The network has $N= 13,332$ nodes (different words) and $473,382$ interactions. The observations here are the frequency of co-occurence between the pair of words $i,j$ at time $t$. We assume that there are no loops in the network, that is $n_{tij} = 0$ for $i=j$. We run the Gibbs sampler with $200,000$ samples, with the first $100,000$ discarded as burn in.  In this case, we choose a truncation level of $K=20,000$.

\begin{figure*}[ht]
\centering
\subfloat[Weights evolution of the words `plane' and `attack'\label{fig:reuters_word_weights_0}] {\includegraphics[width=0.3\textwidth]{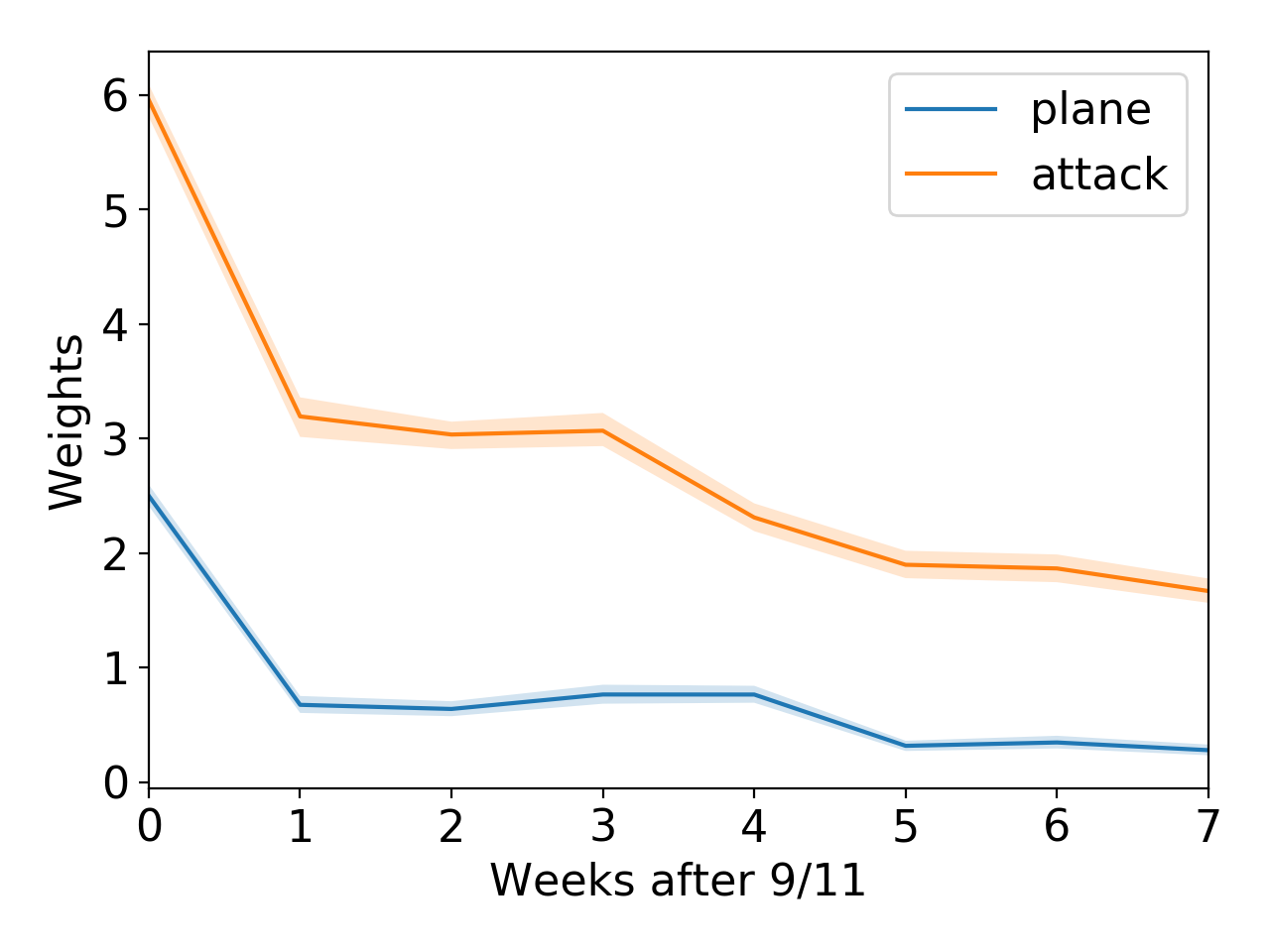}}
\hspace{0.01\textwidth}
\subfloat[Weights evolution of the words `al quaeda', `taliban' and `bin laden'\label{fig:reuters_word_weights_1}]{\includegraphics[width=0.3\textwidth]{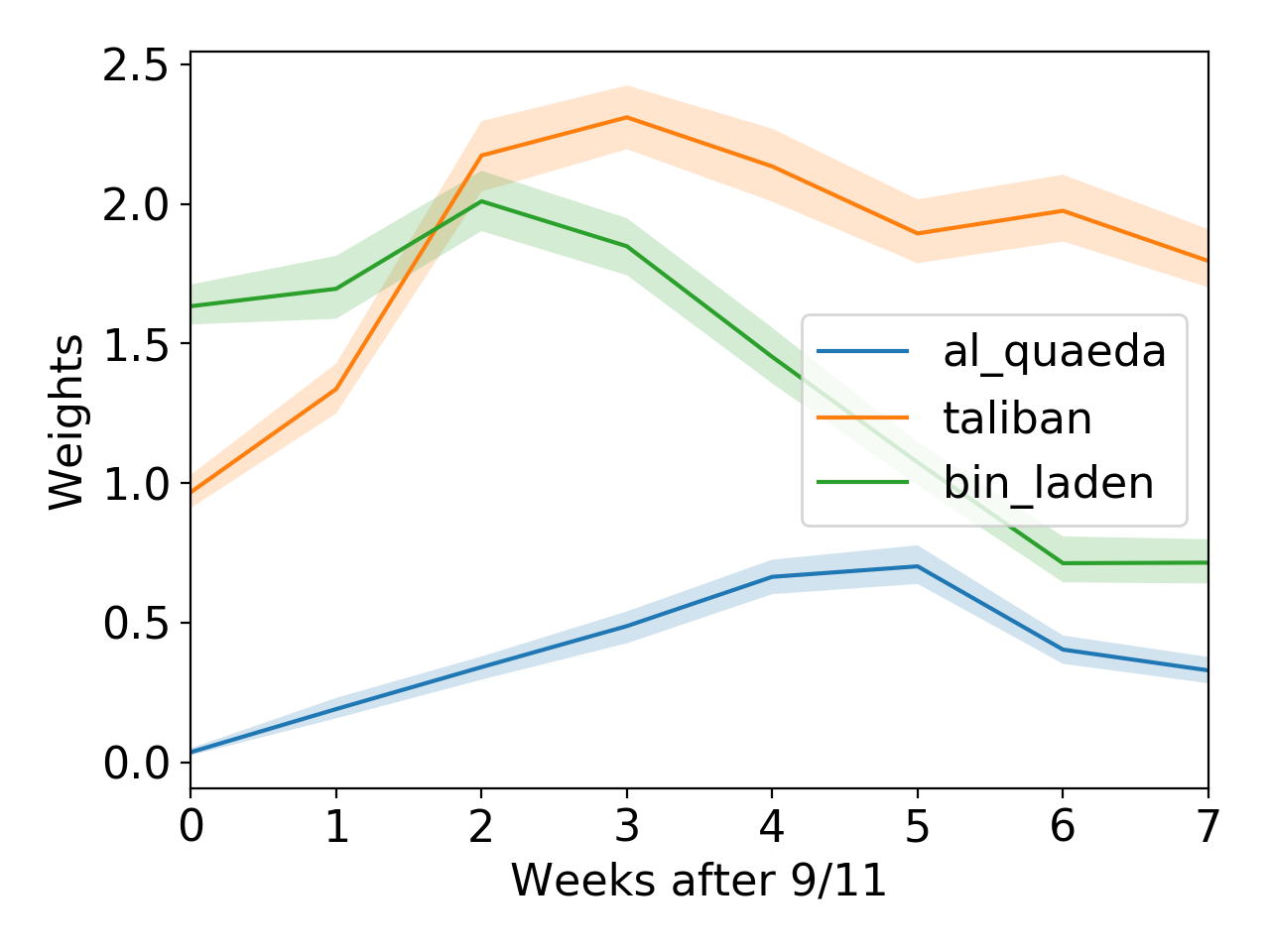}}
\hspace{0.01\textwidth}
\subfloat[Weights evolution of the words `anthrax' and `letter'\label{fig:reuters_word_weights_2}]{\includegraphics[width=0.3\textwidth]{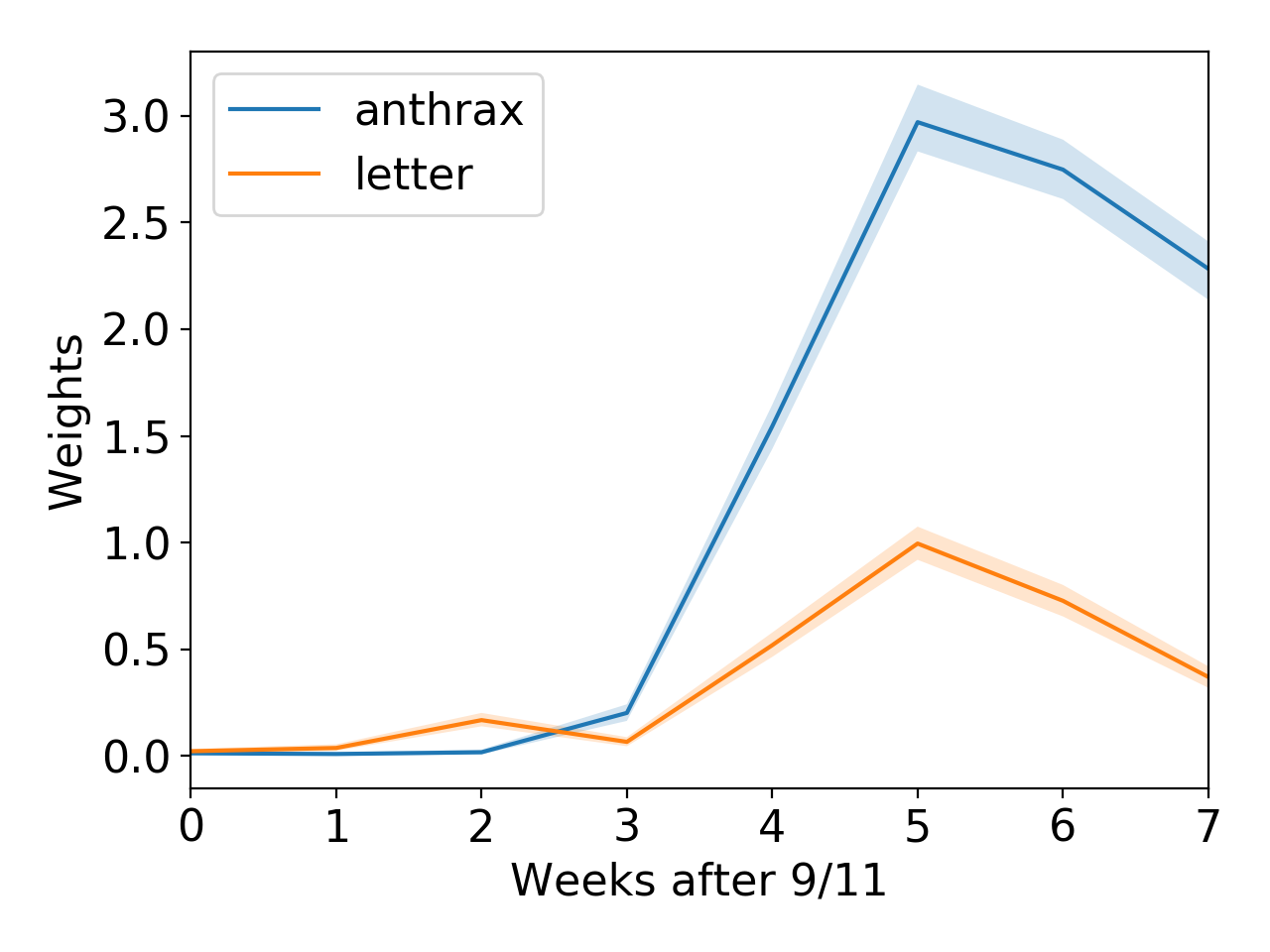}}
\caption{Evolution of the weights, for the Reuters terror dataset}
\label{fig:reuters_words_weights}
\end{figure*}

Figure \ref{fig:reuters_post_pred} suggests that the empirical degree distribution does not follow a power law distribution, and our model therefore provides a moderate fit to the empirical degree distribution. The model is however able to capture the evolution of the popularity of the different words, as shown in Figure \ref{fig:reuters_word_weights_0}. For example, the weights of the words ``plane'' and ``attack'' decrease over time after 9/11, while the words ``letter" and ``anthrax" show a peak a few weeks after the attack. These correspond to the anthrax attacks that occurred over several weeks starting a week after 9/11.

Due to the empirical degree distribution not following a power-law, the estimated value of $\sigma$ is very close to $0$, which causes a slow convergence of the MCMC algorithm.

\section{Discussion and Extensions}
A lot of work has been done on modelling dynamic networks; we restrict ourselves to focussing on Bayesian approaches. Much of this work has centred around extending static models. For example, \citep{Xu2014} and \citep{Durante2014} extend the stochastic block model to the dynamic setting by allowing for parameters that evolve over time. There has also been work on extending the mixed membership stochastic blockmodel \citep{Fu2009, ho2011evolving, xing2010state}, the infinite relational model \citep{ng2017dynamic} and the latent feature relational model \citep{foulds2011dynamic, heaukulani2013dynamic, kim2013nonparametric}. The problem inherent in these models is that they lead to networks which are dense almost surely \citep{Aldous1981, Hoover1979}, a property considered unrealistic for many real-world networks \citep{Orbanz2015} such as social and internet networks.

In order to build models for sparse dynamic networks, \citep{Ghalebi2019} build on the framework of edge-exchangeable networks \citep{cai2016edge, crane2018edge, williamson2016nonparametric}, in which graphs are constructed based on an infinitely exchangeable sequence of edges. As in our case, this framework allows for sparse networks with power-law degree distributions. This work, along with others in this framework \citep{ng2017dynamic, ghalebi2018dynamic}, utilises the mixture of Dirichlet network distributions (MDND) to introduce struture in the networks.

Conversely, our works builds on a different notion of exchangeability \citep{Kallenberg1990, Caron2017}. Within this framework, \citep{Miscouridou2018} use mutually-exciting Hawkes processes to model temporal interaction data. The difference between our work and theirs is that the sociabilities of the nodes are constant throughout time, with the time evolving element driven by previous interactions via the Hawkes process. Their work also builds on that of \citep{Todeschini2016}, incorporating community structure to the network. Exploring how communities appear, evolve and merge could have many practical uses. Thus, expanding our model to capture both evolving node popularity and dynamically changing community structure could be a useful extension.

Furthermore, our model assumes that the time between observations of the network is constant. If this is not the case, it may be helpful to use a continuous-time version of our model. This could be done by considering a birth-death process for the interactions between nodes, where each interaction has a certain lifetime distribution. The continuously evolving node sociabilities could then by described by the Dawson-Watanabe superprocess \citep{Watanabe1968, Dawson1975}.

\section*{Acknowledgements}
C. Naik was supported by the Engineering and Physical Sciences Research Council and Medical Research Council [award reference 1930478]. F. Caron was supported by the Engineering and Physical Sciences Council under grant EP/P026753/1. J. Rousseau was supported by the European Research Council (ERC) under the European Union’s Horizon 2020 research and innovation programme (grant agreement No 834175). K. Palla was supported by the European Research Council under the European Union’s Seventh Framework Programme (FP7/2007-2013) ERC grant agreement no. 617411.

\bibliography{bnpnetwork}
\bibliographystyle{splncs04}

\newpage
\appendix

\section{Proofs}\label{sec:app:proofs}
\subsection{Proof of Equation (10)}
\begin{align*}
&E[W_{t+1}|W_{t}]\\&  =E[E[W_{t+1}|C_{t}]|W_{t}]\\ &  =E[W_{t+1}^{\ast}]+\frac
{1}{\tau+\phi}\sum_{i=1}^{\infty}E[\max(c_{ti}-\sigma,0)|W_{t}]\delta
_{\theta_{i}}\\
&  =E[W_{t+1}^{\ast}]+\frac{1}{\tau+\phi}\sum_{i=1}^{\infty}[\phi
w_{ti}-\sigma(1-e^{-\phi w_{ti}})]\delta_{\theta_{i}}
\end{align*}
Finally, note that, for any measurable set $A\subseteq [0,\alpha]$, using Campbell's theorem,
$$E[W_{t+1}(A)]=\lambda(A)\int_0^\infty w\rho(w)dw=\lambda(A)\tau^{\sigma-1}$$
and similarly
\begin{align*}
E[W^\ast_{t+1}(A)]&=\lambda(A)(\tau+\phi)^{\sigma-1}\\
&=\left(\frac{\tau}{\tau+\phi}\right)^{1-\sigma}E[W_{t+1}(A)]
\end{align*}
where $\lambda$ denotes the Lebesgue measure. Thus, 
\begin{align}
E[W_{t+1}&|W_{t}]=\left (\frac{\tau}{\tau+\phi} \right )^{1-\sigma}E[W_t] \nonumber\\
&~+\frac{1}{\tau+\phi}\sum_{i=1}^{\infty}[\phi
w_{ti}-\sigma(1-e^{-\phi w_{ti}})]\delta_{\theta_{i}}
\end{align}

\section{MCMC algorithm details}\label{sec:app:mcmc_alg}

For $t=1,\ldots,T$, let $\mathbf n_t=(n_{tij})_{1\leq i,j\leq K}$, $\mathbf w_t=(w_{ti})_{i=1,\ldots,K}$, $\mathbf u_t=(u_{ti})_{i=1,\ldots,K}$ and $\mathbf c_t=(c_{ti})_{i=1,\ldots,K}$.

\subsection{Conditional distribution for the interactions $n_{tij}$} \label{sec:multigraph_conditional}
The conditional distribution of $(\mathbf n_{t})_{t=1,\ldots, T}$ given the other variables can be found by recalling Equation (4) in the main article. We have
\begin{align*}
p((\mathbf n_{t})_{t=1,\ldots, T}\mid \text{rest})=\prod_{t=1}^T p(\mathbf n_t \mid \mathbf w_t)
\end{align*}
where 
\begin{align*}
p(\mathbf n_t \mid \mathbf w_t)  &=  \prod_{1\leq i<j\leq K} \frac{(2w_{ti}w_{tj})^{n_{tij}}e^{-2w_{ti}w_{tj}} }{n_{tij}!} \prod_{i=1}^K \frac{(w_{ti}^2)^{n_{tii}}e^{-w_{ti}^2} }{n_{tii}!}\\
&\propto \Big[\prod^{K}_{i=1} w^{m_{ti}}_{ti}\Big] e^{- (\sum^{K}_{i=1} w_{ti})^2 }
\end{align*}
where $m_{ti}=n_{tii}+\sum_{j=1}^K n_{tij}$.

\subsection{Update of $u_{ti}$ and $w_{ti}$}
The auxiliary variables $u_{ti}$ are introduced in order to help with numerical instability problems with the update of the weights $w_{ti}$. These generally stem from the $\left( 1-e^{- (\sigma K/\alpha)^{1/\sigma}w_{ti}}\right)$ term in the BFRY distribution, which can be unstable when  $(\sigma K/\alpha)^{1/\sigma}$ is large. Noting that the truncation level $K$ is a fixed value that we choose, we denote this term by
\begin{align}\label{eq:t_alphasigma_definition}
    t_{\alpha,\sigma} = (\sigma K/\alpha)^{1/\sigma}
\end{align}
in everything that follows.

When defining $u_{ti}$, we want to cancel this problematic term out from the distribution of $w_{ti}$, and thus we sample $u_{ti}$:
\begin{align}
    u_{ti}\mid w_{ti} \sim \text{tExp}(w_{ti},t_{\alpha,\sigma})
\end{align}
where $\text{tExp}(\lambda,a)$ denotes an Exponential distribution with rate parameter $\lambda$, truncated on $[0,a]$. The density function of $u_{ti}|w_{ti}$ is thus given by:

\begin{align}
    p(u_{ti}|w_{ti}) = \frac{w_{ti}e^{-u_{ti} w_{ti}}}{\left( 1-e^{- t_{\alpha,\sigma}w_{ti}}\right)}, \qquad 0 \leq u_{ti} \leq t_{\alpha,\sigma}
\end{align}
We can sample $u_{ti}$ directly using inverse transform sampling. However, due to non-conjugacy we  cannot sample directly from the posterior of $w_{ti}$. For that reason we use Hamiltonian Monte Carlo (HMC) \citep{Neal2000}.
For a given value of $t$, the posterior of $\mathbf w_t$ is given by:
\begin{align*}
&p(\mathbf w_t|\mathbf n_t, \mathbf c_t, \mathbf c_{t-1}, \mathbf u_t) \\
&\propto p(\mathbf n_t |\mathbf w_t)p(\mathbf w_t| \mathbf c_t, \mathbf c_{t-1})p(\mathbf u_t|\mathbf w_t)\\
&\propto \Big[\prod^{K}_{i=1} w^{m_i}_{ti}\Big] e^{- (\sum^{K}_{i=1} w_{ti})^2 } \Big[ \prod^{K}_{i=1} w^{ c_{ti} +\mathds{1}_{t>1}c_{t-1i}-1 -\sigma}_{ti} e^{- (\phi+\mathds{1}_{t>1}\phi+\tau)w_{ti}}\left( 1-e^{- t_{\alpha,\sigma}w_{ti}}\right)\Big]\\
&~~~~\times \Big[\prod^{K}_{i=1} \frac{w_{ti}e^{-u_{ti} w_{ti}}}{\left( 1-e^{- t_{\alpha,\sigma}w_{ti}}\right)} \Big]\\
&\propto \Big[ \prod^{K}_{i=1} w^{ c_{ti} +\mathds{1}_{t>1}c_{t-1i}+1+m_i-1 -\sigma}_{ti} e^{- (\phi+\mathds{1}_{t>1}\phi+\tau + u_{ti} )w_{ti}}\Big] e^{- (\sum^{K}_{i=1} w_{ti})^2 }
\end{align*}
where some of the terms only appear for $\mathds{1}_{t>1}$ because $\mathbf c_0$ does not exist, and thus
$p(w_{1i}|c_{1i}, c_{0,i})=p(w_{1i}|c_{1i})$. Here, we used that
\begin{align}
p(\mathbf w_t| \mathbf c_t, \mathbf c_{t-1}) \propto \prod^{K}_{i=1} \bigg[ \text{BFRY}\left(\alpha''_t/K,\tau+2\phi,\sigma - c_{ti} -c_{t-1i}\right) \bigg]
\end{align}
while noting that the form of $\alpha''_t$ is defined such that
\begin{align*}
 t_{\alpha,\sigma} = (\sigma K/\alpha)^{1/\sigma} = \left(\frac{(\sigma-c_{ti}-c_{t-1i}) K}{\alpha''_t}\right)^{\frac{1}{\sigma-c_{ti}-c_{t-1i}}}
\end{align*}

We use change of variables $y_{ti} = \log w_{ti}$. The HMC algorithm requires computing
the gradient of the log-posterior which is:
\begin{align}
\frac{\partial \log p(\mathbf y_t | \mathbf n_t, \mathbf c_t, \mathbf c_{t-1}, \mathbf u_{t})}{dy_{ti}} &= ( c_{ti} +\mathds{1}_{t>1}c_{t-1i}+1+m_{ti} -\sigma) \nonumber \\
&~~~~-w_{ti}\left(\phi+\mathds{1}_{t>1}\phi+\tau + u_{ti} + 2\sum^{K}_{j=1} w_{tj} \right)
\end{align}
The algorithm then proceeds as follows, for each $t=1,\ldots,T$:

\begin{enumerate}
\item Sample momentum variables $\mathbf p_t=(p_{ti})_{i=1,\ldots,K}$ as:
\begin{eqnarray*}
p_{ti} & \overset{iid}{\sim} & \mathcal{N}(0,1)\quad i=1,\ldots,K
\end{eqnarray*}
\item Simulate $L$ steps of the discretized Hamiltonian via
\begin{eqnarray*}
\log\tilde{\mathbf w}_{t}^{(0)} & = & \log \mathbf w_t\\
\tilde{p}_{ti}& = & p_{ti}+\frac{\varepsilon}{2}\frac{\partial \log p(\log \mathbf w_t | \mathbf n_t, \mathbf c_t, \mathbf c_{t-1}, \mathbf u_{t})}{d(\log w_{ti})} \quad i=1,\ldots,K
\end{eqnarray*}
and for $l=1,\ldots,L-1$
\begin{eqnarray*}
\log\tilde{\mathbf w}_{t}^{(l)} & = & \log\tilde{\mathbf w}_{t}^{(l-1)}+\varepsilon\tilde{\mathbf p}_{t}^{(l-1)}\\
\tilde{p}_{ti}^{(l)} & = & \tilde{p}_{ti}^{(l-1)}+\varepsilon\frac{\partial \log p(\log  \tilde{\mathbf w}_t^{(l)} | \mathbf n_t, \mathbf c_t, \mathbf c_{t-1}, \mathbf u_{t})}{d(\log \tilde{w}^{(l)}_{ti})}
\end{eqnarray*}
where $\tilde{\mathbf p}_{t}=(\tilde{p}_{ti})_{i=1,\ldots,K}$. Finally, set
\begin{eqnarray*}
\log\tilde{\mathbf w}_{t} & = & \log\tilde{\mathbf w}_{t}^{(L-1)}+\varepsilon\tilde{\mathbf p}_{t}^{(L-1)}\\
\tilde{p}_{ti}& = & -\left[\tilde{p}_{ti}^{(L-1)}+\frac{\varepsilon}{2}\frac{\partial \log p(\log  \tilde{\mathbf w}_t | \mathbf n_t, \mathbf c_t, \mathbf c_{t-1}, \mathbf u_{t})}{d(\log \tilde{w}_{ti})}\right] \quad i=1,\ldots,K
\end{eqnarray*}

\item Accept $\tilde{\mathbf w}_{t},$ with probability $a = \min(1, r)$, where $r$ is given by:
\begin{align}
r &= \Big[\prod^{K}_{i=1} \Big(\frac{\tilde{w}_{ti}   } {w_{ti}} \Big)^{c_{ti} +\mathds{1}_{t>1}c_{t-1i}+1+m_{ti} -\sigma} \Big] \nonumber \\
&\times e^{-(\sum^{K}_{i=1} \tilde{w}_{ti})^2 + (\sum^{K}_{i=1} w_{ti})^2 - (\phi+\mathds{1}_{t>1}\phi+\tau)(\sum^{K}_{i=1} \tilde{w}_{ti} - \sum^{K}_{i=1} w_{ti}) }\nonumber \\
&\times e^{-\sum^{K}_{i=1}u_{ti} \tilde{w}_{ti} - \sum^{K}_{i=1} u_{ti}w_{ti}} e^{-\frac{1}{2} \sum^{K}_{i=1}\tilde{p}_{ti}^2 - p_{ti}^2}
\end{align}

\end{enumerate}
In our case, we take $L=10$, and tune $\varepsilon$ on the first $10,000$ samples in order to achieve an acceptance rate of $0.65$.

\subsection{Update of $c_{ti}$}
We update $\mathbf c_t |\mathbf w_{t}, \mathbf w_{t+1}$ for each $t$. The conditional distribution is:
\begin{align}
p(c_{tk} | w_{tk}, w_{t+1k}) \propto& p(c_{tk}| w_{tk})p(w_{t+1k} | c_{tk})
\end{align}
where
\begin{align}
p(c_{tk}| w_{tk}) =& \Poisson{c_{tk}; \phi w_{tk}}\\
p(w_{t+1, k}| c_{tk}, w_{tk})  =&\text{BFRY}\left(w_{t+1, k};\alpha'_t/K,\tau+\phi,\sigma - c_{ti}\right)
\end{align}

For each $t=1, \dots, T$ and $k=1, \dots K$. We use Metropolis-Hastings to sample $c_{tk}$ with a Poisson as the proposal distribution, i.e. $\tilde{c}_{tk} \sim \Poisson{\phi w_k}$. Accept with probability $\min(1, r)$ where

\begin{align}
r = \frac{p(w_{t+1, k} | \tilde{c}_{tk})}{p(w_{t+1, k} | c_{tk})} = \frac{\text{BFRY}\left(w_{t+1, k};\tilde{\alpha}'_t/K,\tau+\phi,\sigma - \tilde{c}_{ti}\right)}{\text{BFRY}\left(w_{t+1, k};\alpha'_t/K,\tau+\phi,\sigma - c_{ti}\right)}
\end{align}
where we remember that $\alpha'_t$ depends on $c_t$.

\subsection{Update of $\alpha$, $\sigma$, $\tau$, $\phi$}

The joint posterior is given by
\begin{align}\label{eq:jointpost}
p(\alpha, \sigma, \tau, \phi|C,W)\propto& \prod^{K}_{k=1} \bigg(p(w_{1k}) p(c_{1k}| \phi, w_{1k})\bigg) \prod^{T}_{t=2}
\bigg[\prod^{K}_{k=1} \bigg(p(w_{tk} |c_{t-1k}) p(c_{tk}| \phi, w_{tk})\bigg)\bigg] \nonumber \\
& \times p(\alpha) p(\sigma) p(\tau) p(\phi)\nonumber \\
\propto& \prod^{K}_{k=1} \bigg(\text{BFRY}\left(w_{1k};\alpha/K,\tau,\sigma \right) \frac{(w_{1k}\phi)^{c_{1k}}e^{-\phi w_{1k}}}{c_{1k}!}\bigg) \nonumber \\
& \times \prod^{T}_{t=2}
\bigg[\prod^{K}_{k=1} \bigg(\text{BFRY}\left(w_{tk};\alpha'_t/K,\tau+\phi,\sigma - c_{t-1,k}\right) \frac{(w_{tk}\phi)^{c_{tk}}e^{-\phi w_{tk}}}{c_{tk}!}\bigg)\bigg] \nonumber \\
&\times p(\alpha) p(\sigma) p(\tau)p(\phi)
\end{align}
We place Gamma priors on $\alpha$, $\tau$ and $\phi$, and a Beta prior on $\sigma$:

\begin{align}
p(\alpha) &=\GammaD{\alpha; a_1, a_2}\\
p(\sigma) &= \BetaD{\sigma; s_1, s_2}  = \frac{\sigma^{s_1 -1 } (1-\sigma)^{s_2-1} }{B(s_1, s_2)}\\
p(\tau) &= \GammaD{\tau; t_1, t_2} \\
p(\phi) &= \GammaD{\phi; f_1, f_2}
\end{align}
We then use a Metropolis Hastings sampler, with proposals given by:

\begin{align}
q(\tilde{\alpha}|\alpha ) &=\logn{\tilde{\alpha}; \log(\alpha), \sigma_{\alpha}}  = \frac{1}{\sigma_\alpha \sqrt{2\pi} \tilde{\alpha}} \exp{- \frac{(\log{\tilde{\alpha}} - \log{\alpha} )^2}{2\sigma^2_\alpha} } \\
q(\tilde{\tau}|\tau ) &=\logn{\tilde{\tau}; \log(\tau), \sigma_{\tau}}  = \frac{1}{\sigma_\tau \sqrt{2\pi} \tilde{\tau}} \exp{- \frac{(\log{\tilde{\tau}} - \log{\tau} )^2}{2\sigma^2_\tau} } \\
q(\tilde{\phi}|\phi ) &=\logn{\tilde{\phi}; \log(\phi), \sigma_{\phi}}  = \frac{1}{\sigma_\phi \sqrt{2\pi} \tilde{\phi}} \exp{- \frac{(\log{\tilde{\phi}} - \log{\phi} )^2}{2\sigma^2_\phi} } \\
q(\tilde{\sigma}|\sigma ) &= \frac{1}{\sigma_\sigma \sqrt{2\pi} \tilde{\sigma}(1- \tilde{\sigma}) }   \exp{- \frac{(\log{\frac{\tilde{\sigma}}{1 - \tilde{\sigma}}    } - \log{\frac{\sigma}{1-\sigma}} )^2}{2\sigma^2_\sigma} }
\end{align}
The parts of the MH ratio involving the priors and proposals of the hyperparameters are given by:
\begin{align}
\frac{p(\tilde{\alpha})q(\alpha| \tilde{\alpha} )}{p(\alpha)q(\tilde{\alpha}|\alpha )} &=e^{-a_2(\tilde{\alpha}-\alpha)}\left(\frac{\tilde{\alpha}}{\alpha}\right)^{a_1}\\
\frac{p(\tilde{\tau})q(\tau| \tilde{\tau} )}{p(\tau)q(\tilde{\tau}|\tau )} &=e^{-t_2(\tilde{\tau}-\tau)}\left(\frac{\tilde{\tau}}{\tau}\right)^{t_1}\\
\frac{p(\tilde{\phi})q(\phi| \tilde{\phi} )}{p(\phi)q(\tilde{\phi}|\phi )} &=e^{-f_2(\tilde{\phi}-\phi)}\left(\frac{\tilde{\phi}}{\phi}\right)^{f_1}\\
\frac{p(\tilde{\sigma})q(\sigma| \tilde{\sigma} )}{p(\sigma)q(\tilde{\sigma}|\sigma )} &=\left(\frac{\tilde{\sigma}}{\sigma}\right)^{s_1}\left(\frac{1-\tilde{\sigma}}{1-\sigma}\right)^{s_2}
\end{align}
The MH ratio without the priors and proposals on $\alpha, \sigma, \tau$ and $\phi$ is given by

\begin{align}
    \dfrac{\splitdfrac{ \prod^{K}_{k=1} \bigg(\text{BFRY}\left(w_{1k};\tilde{\alpha}/K,\tilde{\tau},\tilde{\sigma} \right) \frac{(w_{1k}\tilde{\phi})^{c_{1k}}e^{-\tilde{\phi} w_{1k}}}{c_{1k}!}\bigg)}{\times \prod^{T}_{t=2}
\bigg[\prod^{K}_{k=1} \bigg(\text{BFRY}\left(w_{tk};\tilde{\alpha}'_t/K,\tilde{\tau}+\tilde{\phi},\tilde{\sigma} - c_{t-1,k}\right) \frac{(w_{tk}\tilde{\phi})^{c_{tk}}e^{-\tilde{\phi} w_{tk}}}{c_{tk}!}\bigg)\bigg]}}
{\splitdfrac{\prod^{K}_{k=1} \bigg(\text{BFRY}\left(w_{1k};\alpha/K,\tau,\sigma \right) \frac{(w_{1k}\phi)^{c_{1k}}e^{-\phi w_{1k}}}{c_{1k}!}\bigg)}{\times \prod^{T}_{t=2}\bigg[\prod^{K}_{k=1} \bigg(\text{BFRY}\left(w_{tk};\alpha'_t/K,\tau+\phi,\sigma - c_{t-1,k}\right) \frac{(w_{tk}\phi)^{c_{tk}}e^{-\phi w_{tk}}}{c_{tk}!}\bigg)\bigg]}}
\end{align}
The logarithm of the ratio is then given by:
\begin{align}
    &K\left[\log(\tilde{\sigma}) - \log(\sigma) \right] + K\left[\log(\Gamma(1-\sigma)) - \log(\Gamma(1-\tilde{\sigma})) \right] + \left[- (\tilde{\tau} + \tilde{\phi}) + (\tau + \phi) \right]\sum_{k=1}^K w_{1k}\nonumber\\
    &+K\left[\log\left(\left\{ \left( \tau +t_{\alpha,\sigma} \right)^{\sigma} -\tau^{\sigma}\right\} \right) - \log\left(\left\{ \left( \tilde{\tau} +t_{\tilde{\alpha},\tilde{\sigma}} \right)^{\tilde{\sigma}} -\tilde{\tau}^{\tilde{\sigma}}\right\} \right) \right]\nonumber\\
    &+\left(\sigma - \tilde{\sigma}\right)\sum_{t=1}^T\sum_{k=1}^K\log(w_{tk}) +\left[\log(\tilde{\phi}) - \log\phi)\right]\sum_{t=1}^T\sum_{k=1}^Kc_{tk}\nonumber\\
    &+ \sum_{t=1}^T\sum_{k=1}^K \left[ \log\left(1-e^{-t_{\tilde{\alpha},\tilde{\sigma}}w_{tk}}\right) - \log\left(1-e^{-t_{\alpha,\sigma}w_{tk}}\right) \right]\nonumber\\
    &+\sum_{t=2}^T\sum_{k=1}^K\log\left( \frac{\Gamma(1-\sigma + c_{t-1k})}{\Gamma(1-\tilde{\sigma} + c_{t-1k})}\right) + \left[- (\tilde{\tau} + 2\tilde{\phi}) + (\tau + 2\phi) \right]\sum_{t=2}^T\sum_{k=1}^K w_{tk}\nonumber\\
    &+\sum_{t=2}^T\sum_{k=1}^K\log\left( \frac{\tilde{\sigma} - c_{t-1k}}{\sigma - c_{t-1k}}\right) + \sum_{t=2}^T\sum_{k=1}^K\log\left( \frac{\left\{ \left( \tau + \phi +t_{\alpha,\sigma} \right)^{\sigma - c_{t-1k}} -(\tau+\phi)^{\sigma- c_{t-1k}}\right\}}{\left\{ \left( \tilde{\tau}+\tilde{\phi} +t_{\tilde{\alpha},\tilde{\sigma}} \right)^{\tilde{\sigma}- c_{t-1k}} -\left(\tilde{\tau}+\tilde{\phi}\right)^{\tilde{\sigma}- c_{t-1k}}\right\}}\right)
\end{align}

\subsection{Log-posterior density}
\label{subsec:app:jointposterior}

The posterior probability density function, given the latent variables and up to a normalizing constant, thus takes the form:
\begin{align}
&p\left( \left(w_{tk},c_{tk}\right)_{k=1,\ldots,K, t= 1, \ldots, T},\sigma,\tau, \phi,\alpha, \rho \mid  (n_{tij})_{1\leq i,j\leq K, t=1,\ldots, T}\right)\nonumber\\
& \propto \prod^{T}_{t=1}\bigg[p(\mathbf n_t| \mathbf w_t)\bigg]\prod^{K}_{k=1} \bigg(p(w_{1k}) p(c_{1k}| \phi, w_{1k})\bigg) \prod^{T}_{t=2}
\bigg[\prod^{K}_{k=1} \bigg(p(w_{tk} |c_{t-1k}) p(c_{tk}| \phi, w_{tk})\bigg)\bigg] \nonumber \\
&~~~~\times p(\alpha) p(\sigma) p(\tau) p(\phi)\nonumber \\
& \propto \prod^{T}_{t=1}\bigg[\Big[\prod^{K}_{k=1} w^{m_{tk}}_{tk}\Big] e^{- (\sum^{K}_{k=1} w_{tk})^2 })\bigg] \prod^{K}_{k=1} \bigg(\text{BFRY}\left(w_{1k};\alpha/K,\tau,\sigma \right) \frac{(w_{1k}\phi)^{c_{1k}}e^{-\phi w_{1k}}}{c_{1k}!}\bigg)\nonumber\\
&~~~~\times \prod^{T}_{t=2} \bigg[\prod^{K}_{k=1} \bigg(\text{BFRY}\left(w_{tk};\alpha'_t/K,\tau+\phi,\sigma - c_{t-1,k}\right) \frac{(w_{tk}\phi)^{c_{tk}}e^{-\phi w_{tk}}}{c_{tk}!}\bigg)\bigg] \nonumber \\
&~~~~\times p(\alpha) p(\sigma) p(\tau)p(\phi).
\label{eq:app:jointposterior}
\end{align}

\section{MCMC plots}\label{sec:app:mcmc_plots}
In this section we give the MCMC trace plots for the simulated and real data experiments. In each case, the samples are thinned so that we have $100$ samples from the posterior.
\subsection{Simulated Data}
\begin{figure}[ht]
\centering
\subfloat[$\alpha$ trace]{\includegraphics[width=0.3\textwidth]{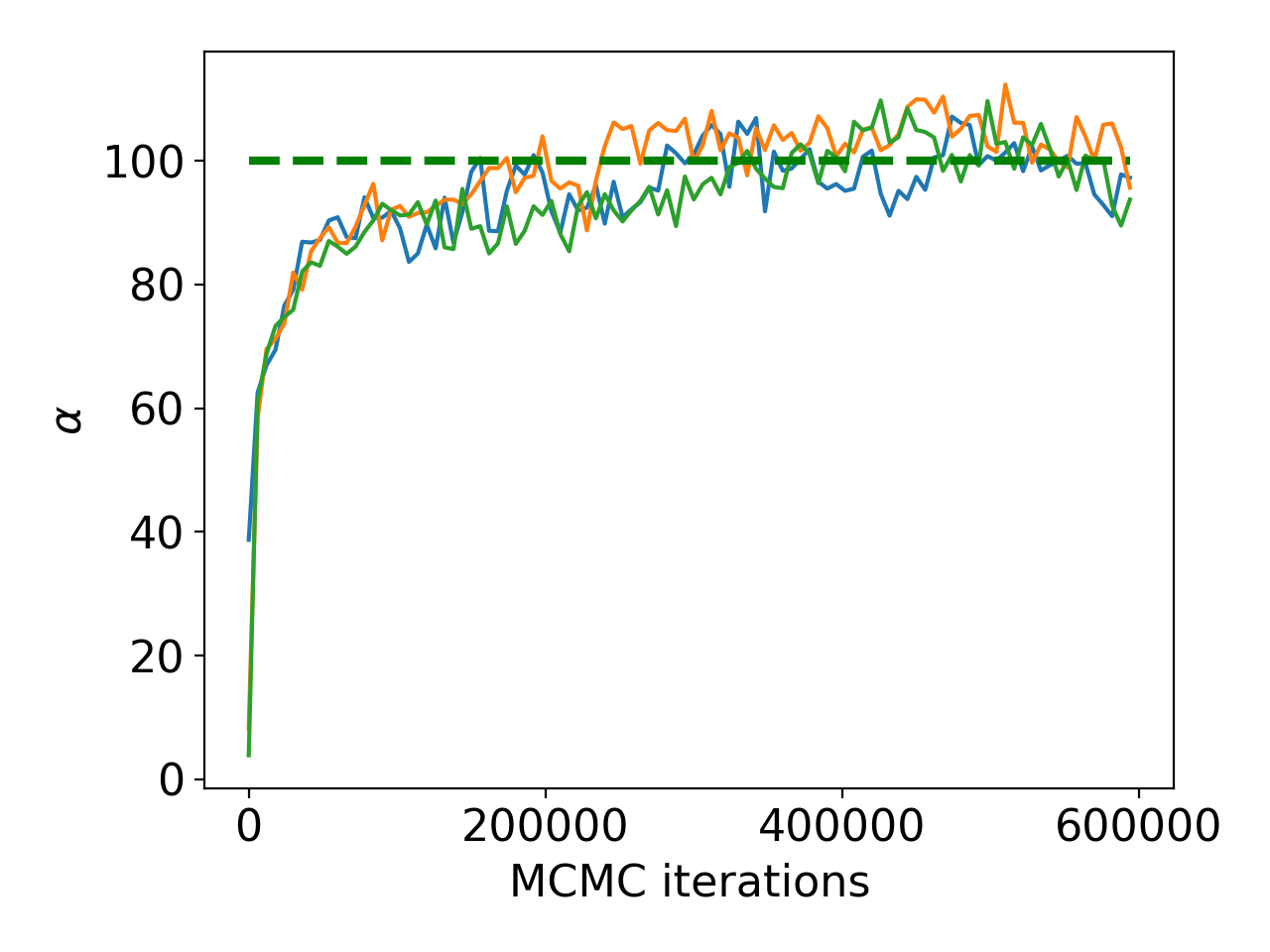}}
\subfloat[$\sigma$ trace]{\includegraphics[width=0.3\textwidth]{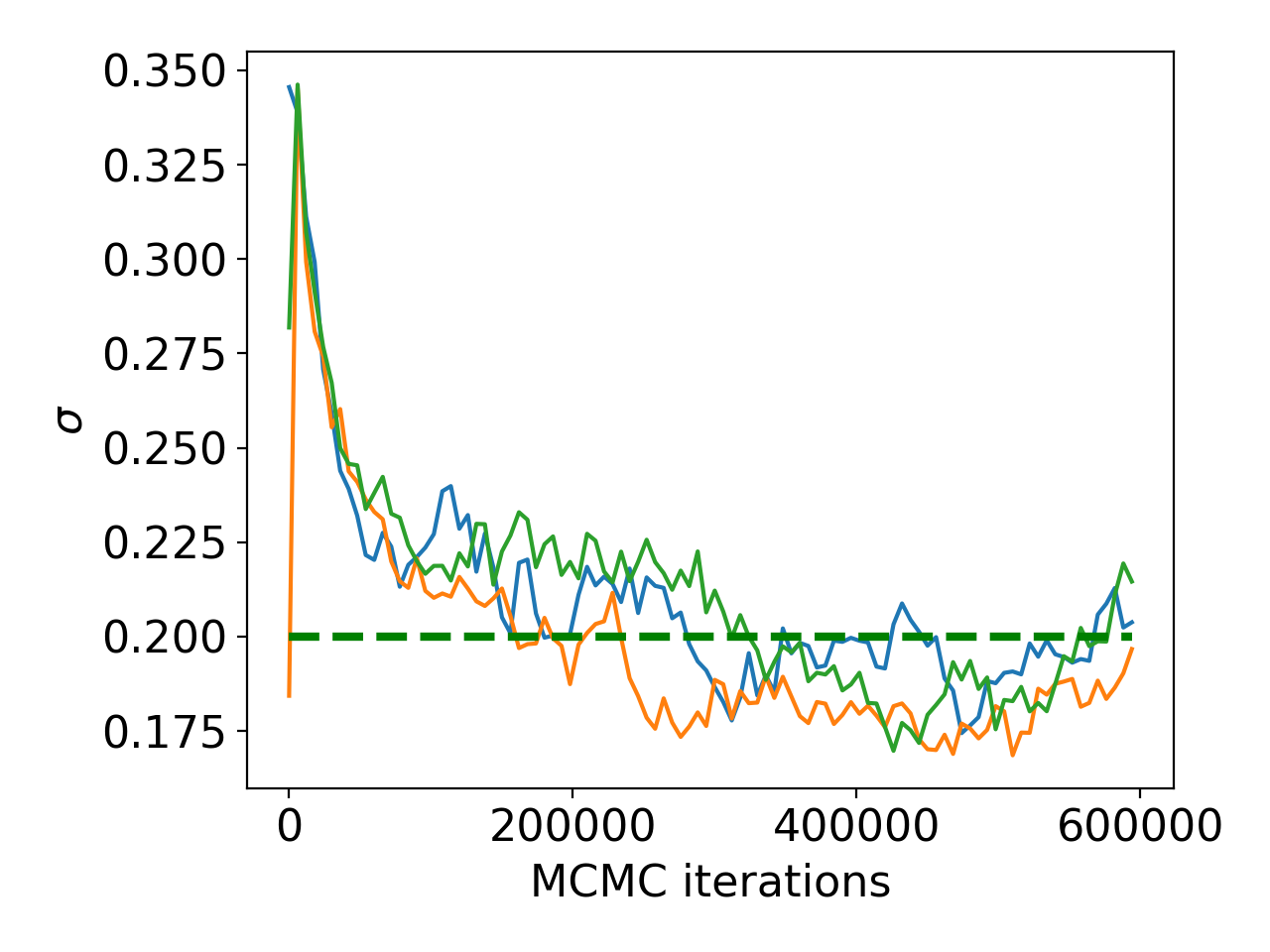}}
\subfloat[$\tau$ trace]{\includegraphics[width=0.3\textwidth]{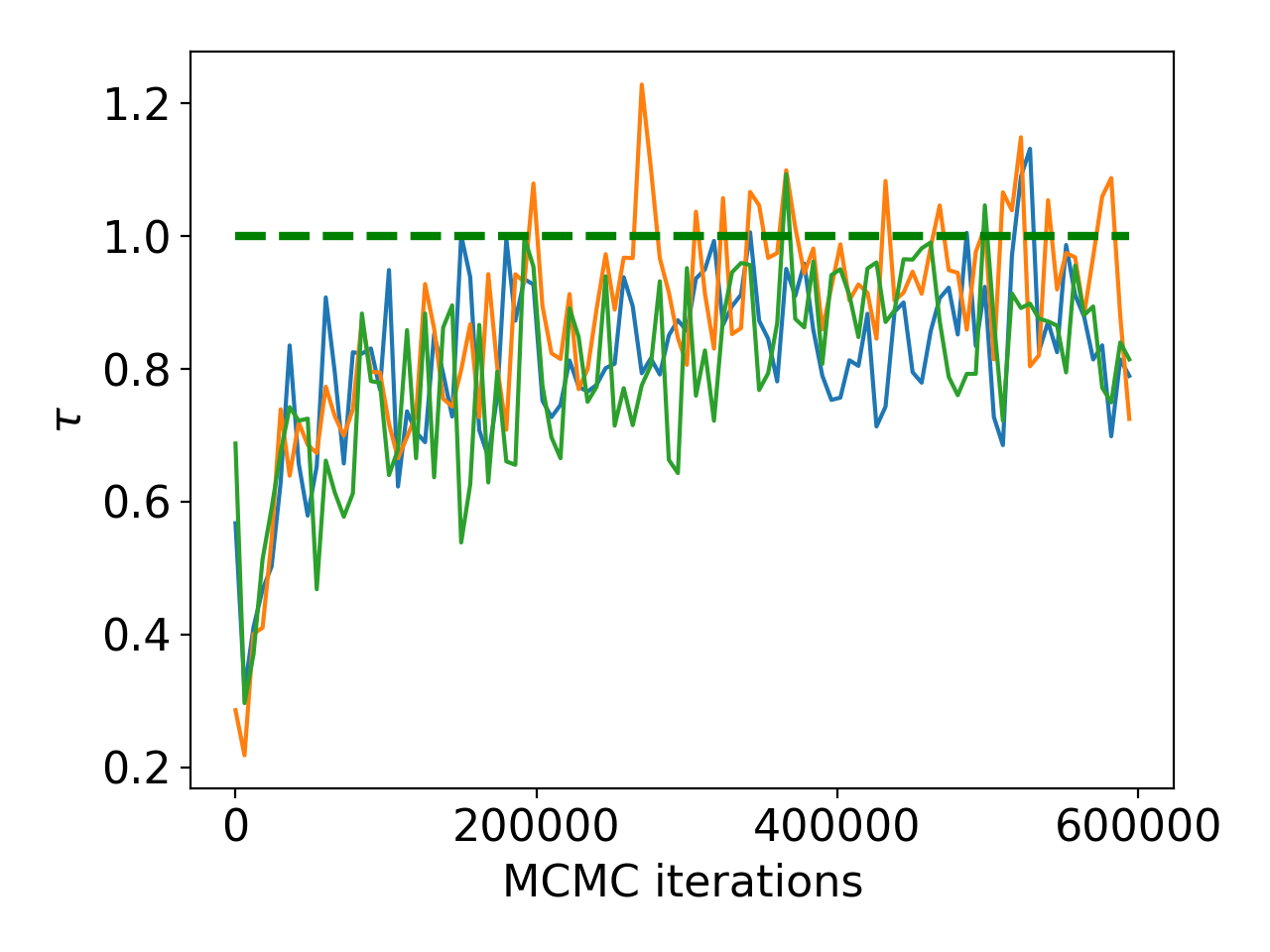}}\\
\subfloat[$\phi$ trace]{\includegraphics[width=0.3\textwidth]{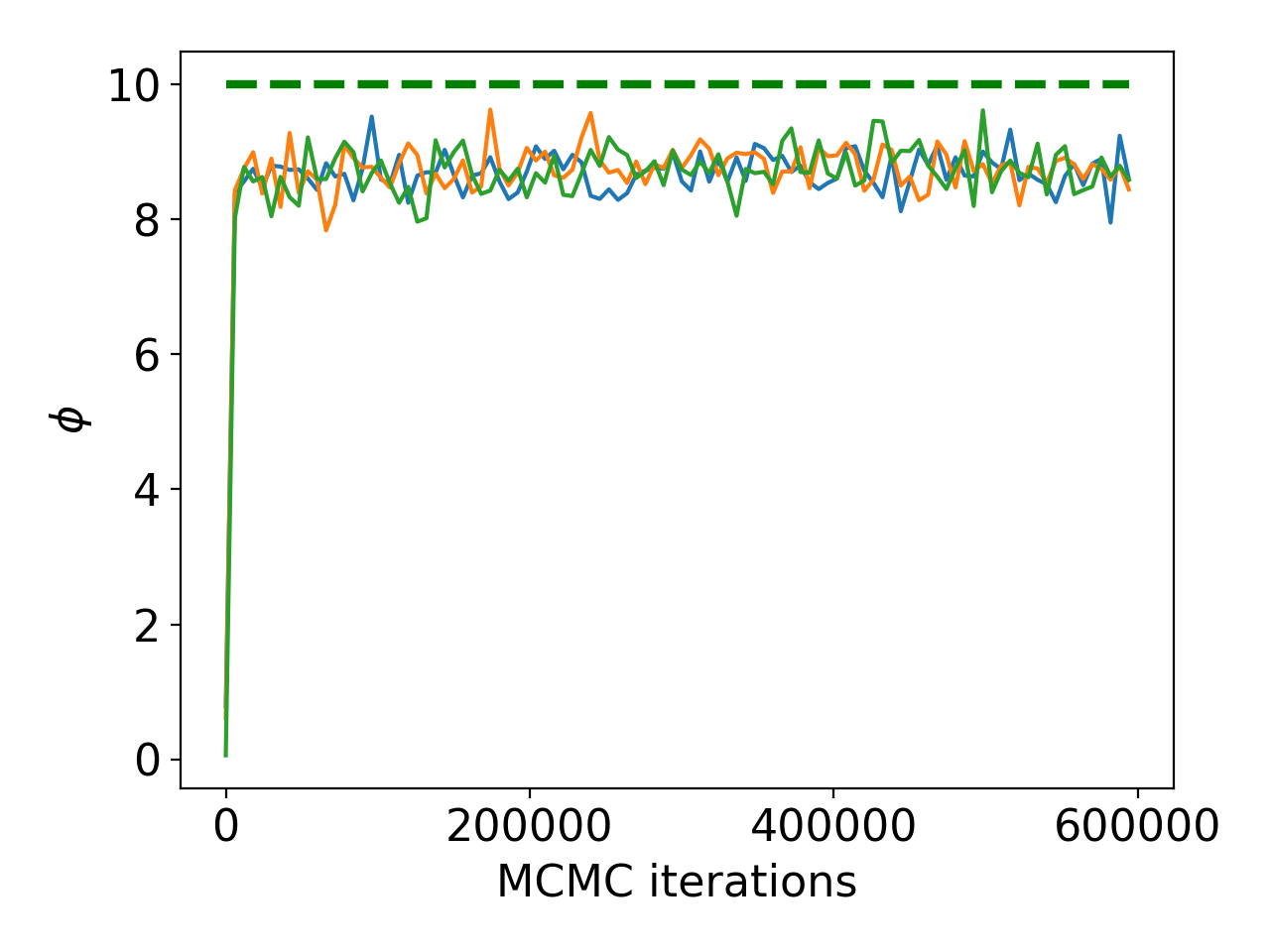}}
\subfloat[Logposterior trace]{\includegraphics[width=0.3\textwidth]{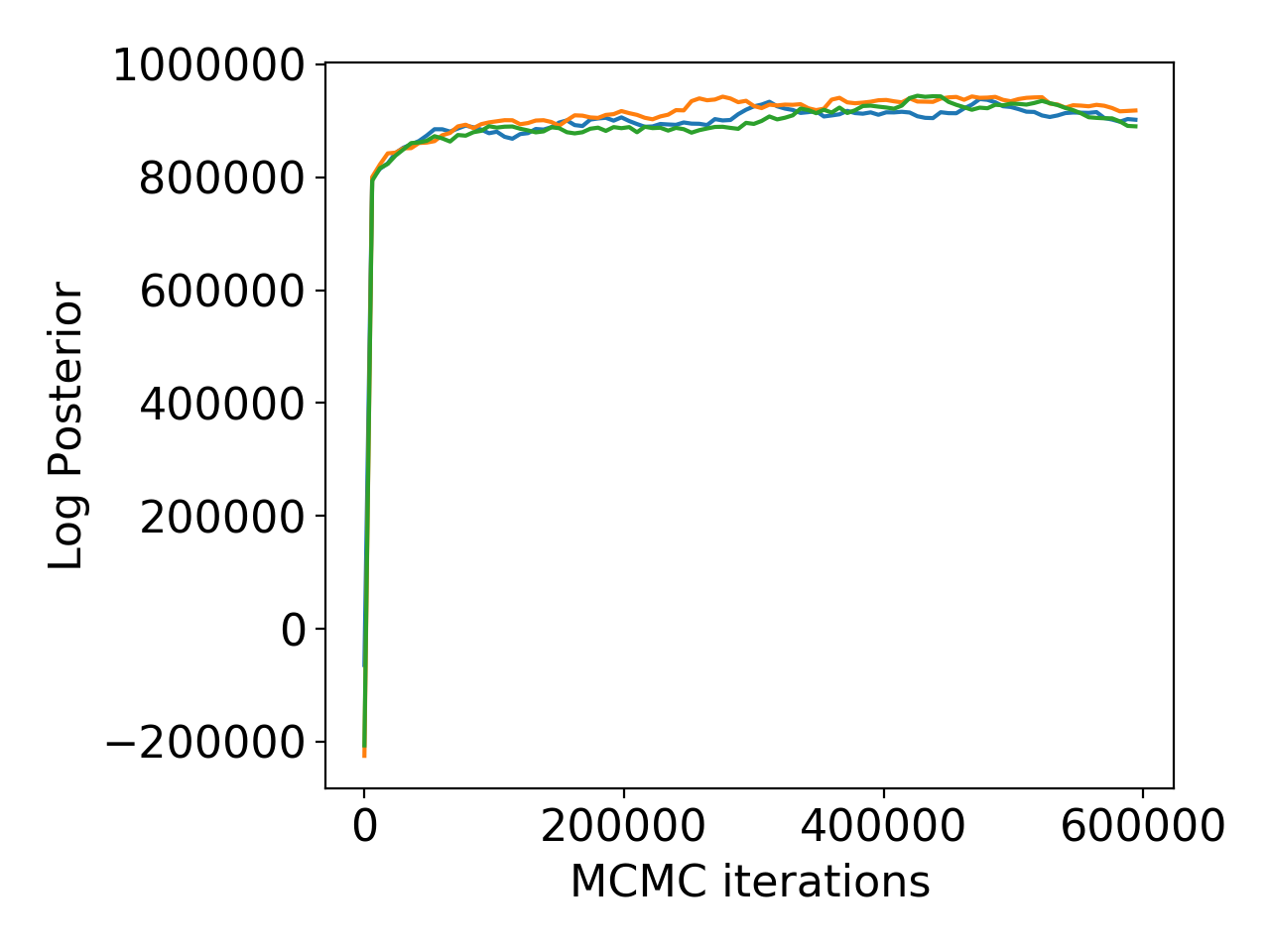}}

\caption{MCMC trace plots of hyperparameters and logposterior for the network simulated from the GG model, with $\alpha = 100$, $\sigma = 0.2$, $\tau = 1$ and $\phi= 10$. The true values in each case are denoted by the dotted green line.}
\label{fig:mcmc_trace_simulated_parameters}

\end{figure}

In Figure \ref{fig:mcmc_trace_simulated_parameters} we see the trace plots for the hyperparameters $\alpha, \sigma, \tau$ and $\phi$, as well at the logposterior (up to a constant), for the network simulated from the GG model. We can see that the parameters other than $\phi$ converge well to their true values, and that in each case the mixing of the three chains is good. Furthermore, we can see that the logposterior has converged to a stable value.

\subsection{Real Data}

\subsubsection{Reddit Hyperlink Network}
\begin{figure}[ht]
\centering
\subfloat[$\alpha$ trace]{\includegraphics[width=0.3\textwidth]{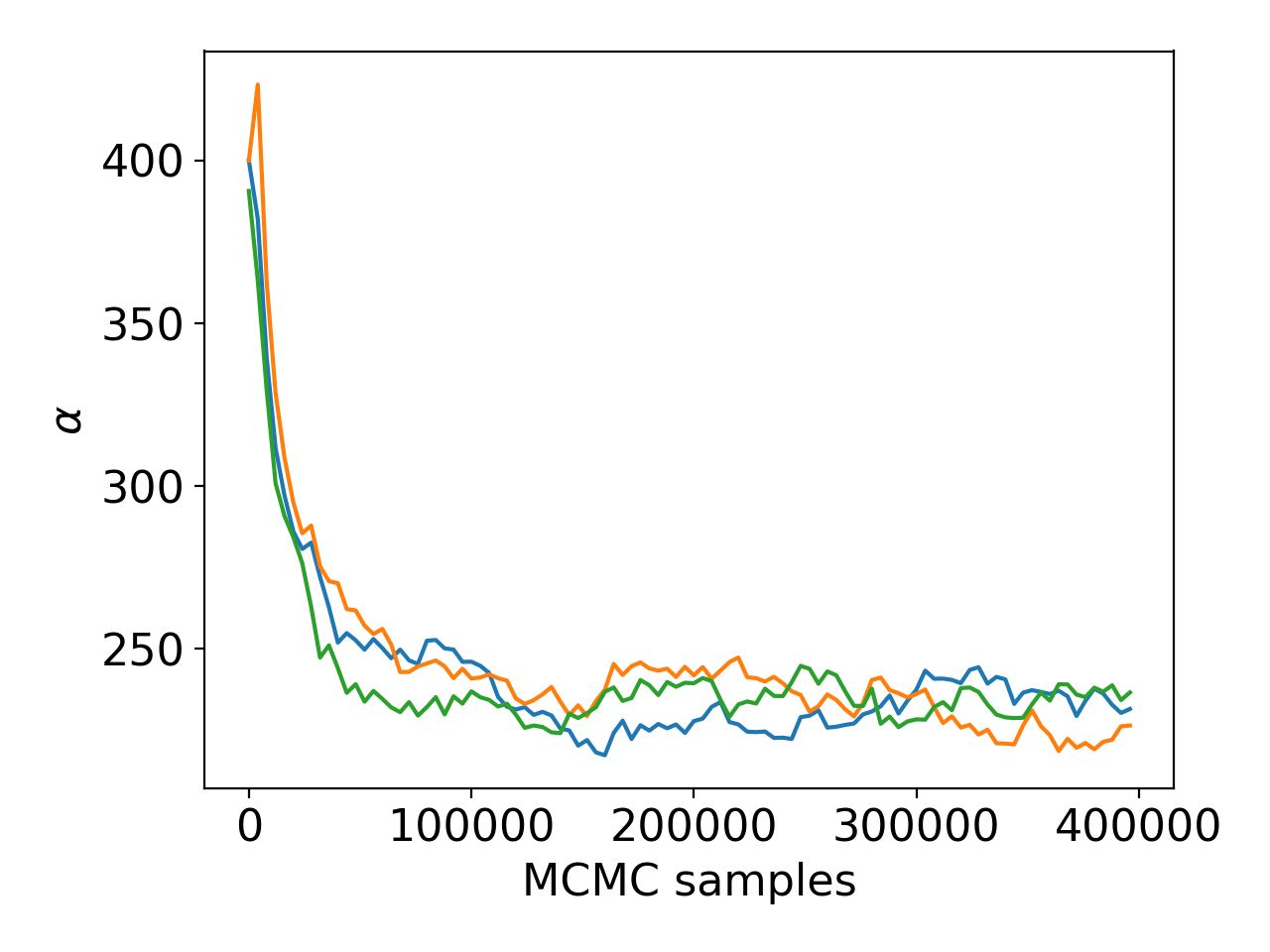}}
\subfloat[$\sigma$ trace]{\includegraphics[width=0.3\textwidth]{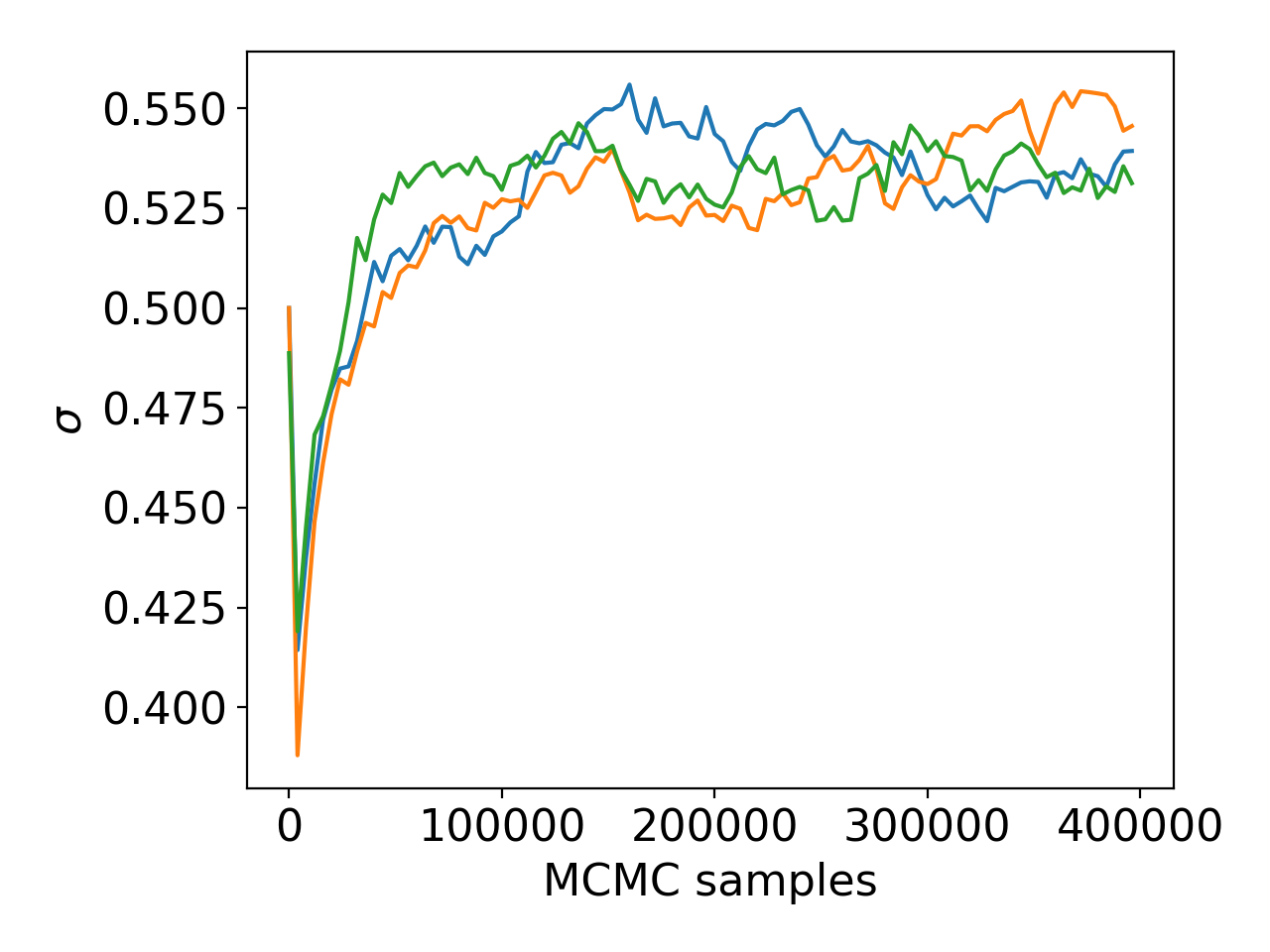}}
\subfloat[$\tau$ trace]{\includegraphics[width=0.3\textwidth]{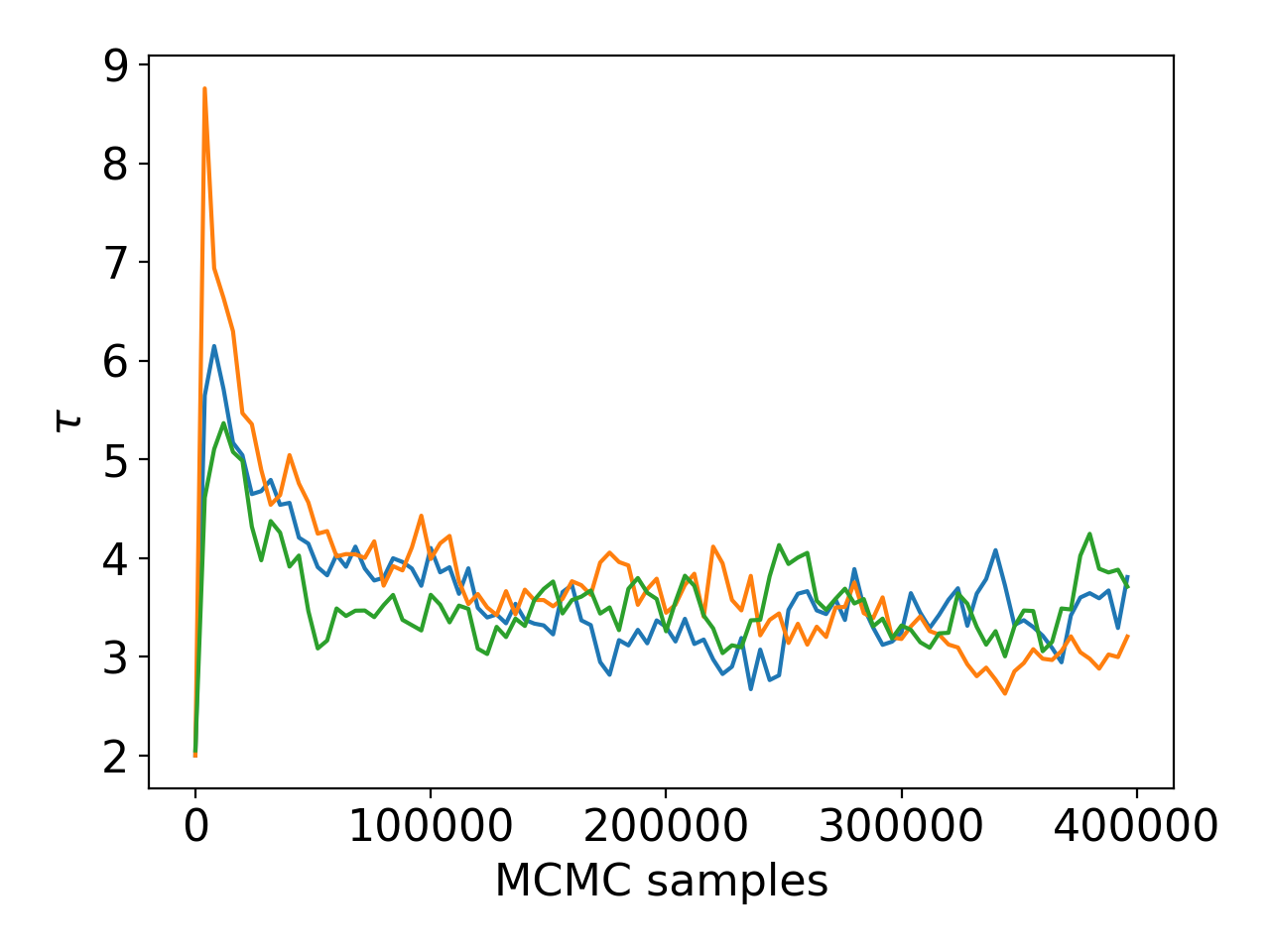}}\\
\subfloat[$\phi$ trace]{\includegraphics[width=0.3\textwidth]{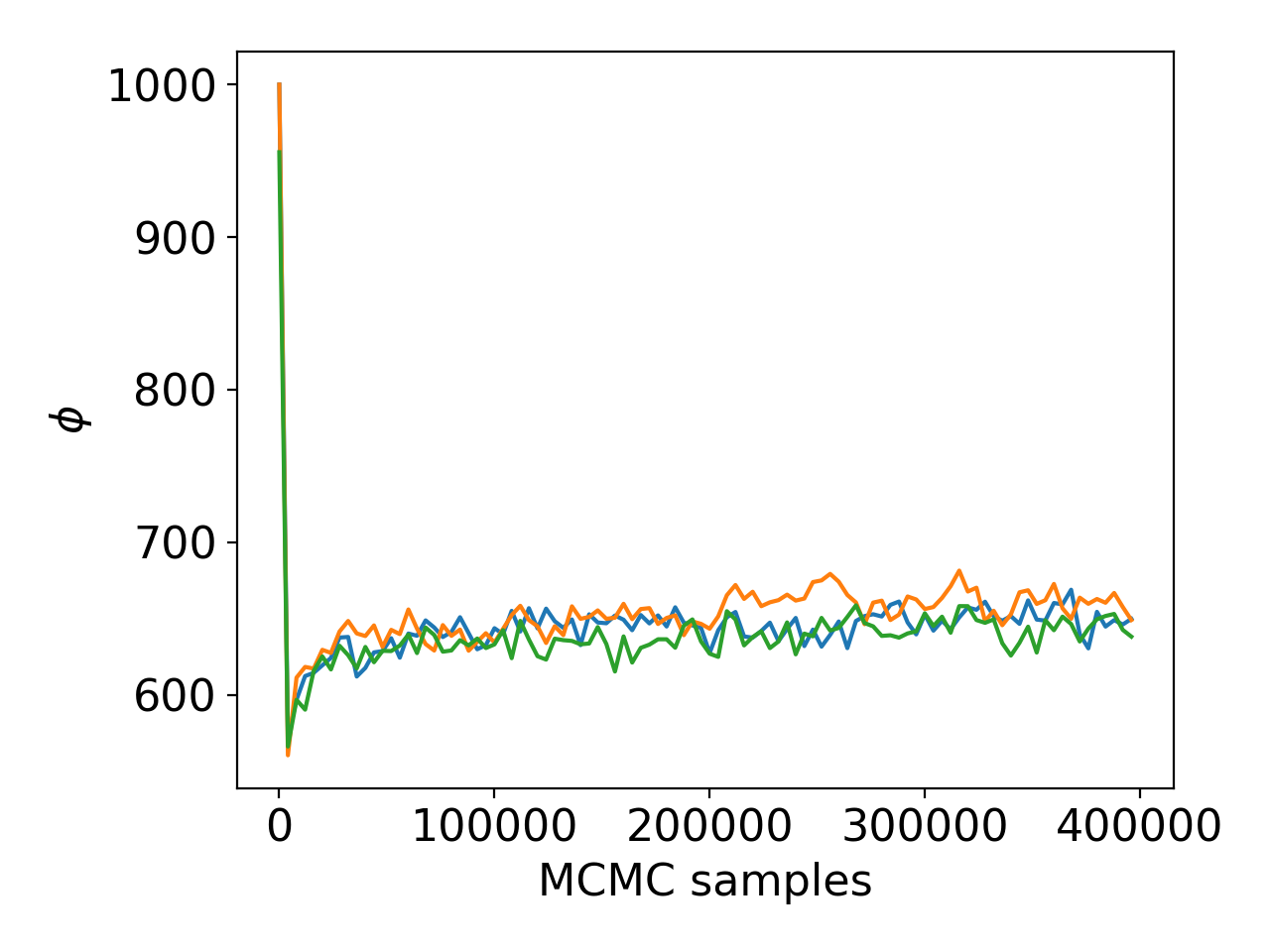}}
\subfloat[Logposterior trace]{\includegraphics[width=0.3\textwidth]{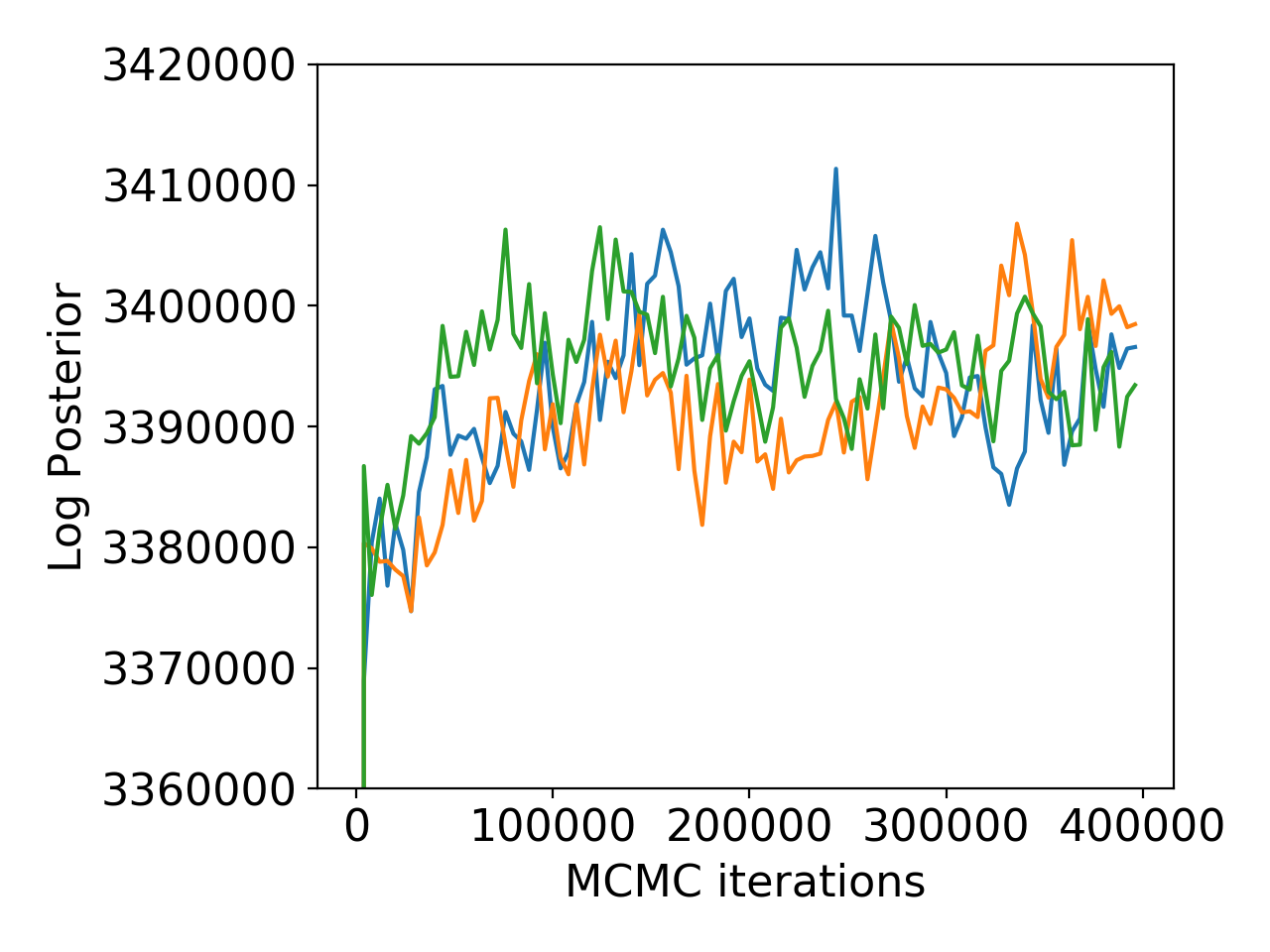}}

\caption{MCMC trace plots of hyperparameters and logposterior for the Reddit hyperlink network.}
\label{fig:mcmc_trace_reddit_parameters}

\end{figure}

In Figure \ref{fig:mcmc_trace_reddit_parameters} we see the trace plots for the hyperparameters $\alpha, \sigma, \tau$ and $\phi$, as well at the logposterior, for the Reddit hyperlink network. We can see that the parameters and the logposterior have converged to a stable value.

\subsubsection{Reuters Terror Dataset}
\begin{figure}[ht]
\centering
\subfloat[$\alpha$ trace]{\includegraphics[width=0.3\textwidth]{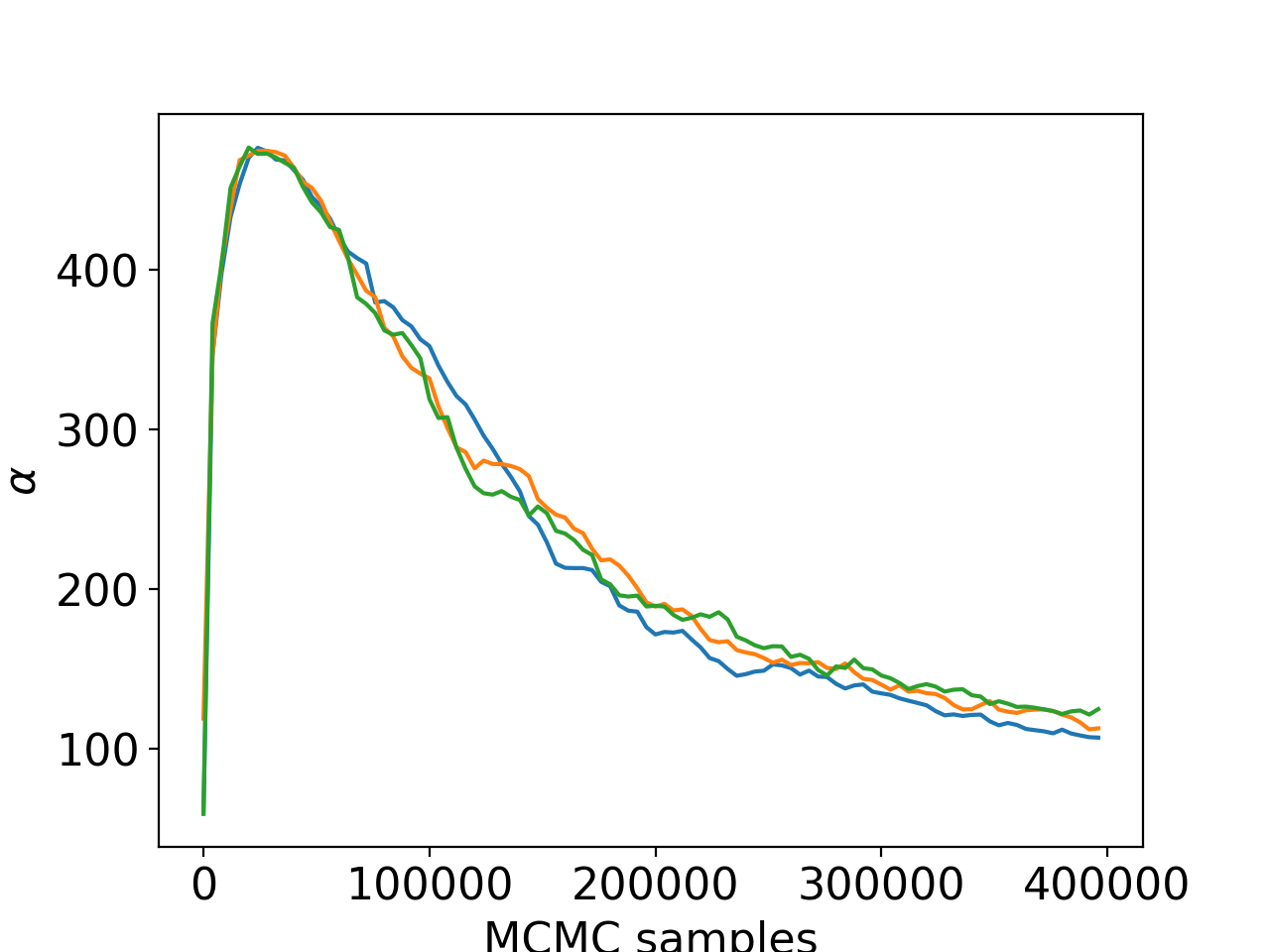}}
\subfloat[$\sigma$ trace]{\includegraphics[width=0.3\textwidth]{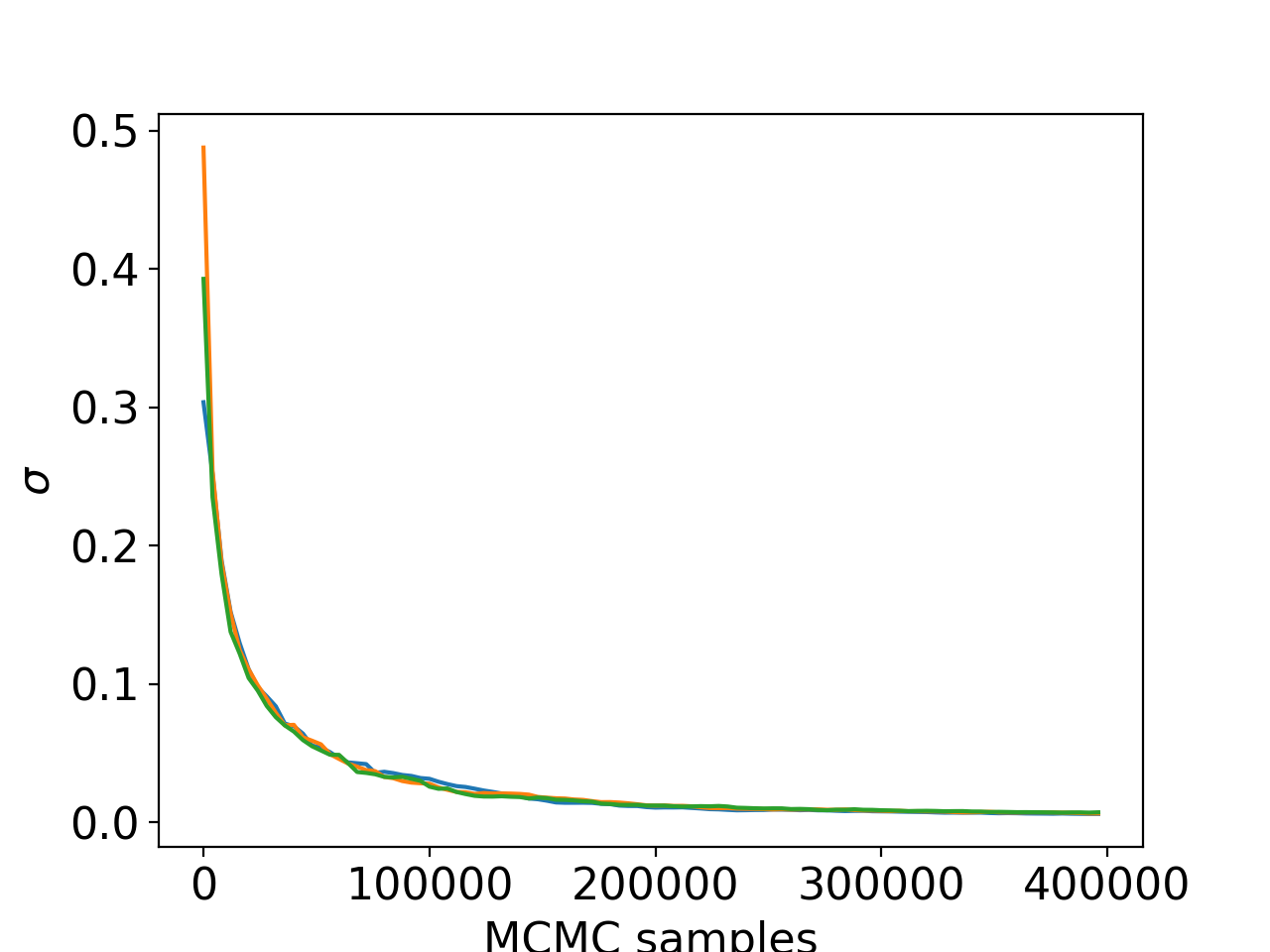}}
\subfloat[$\tau$ trace]{\includegraphics[width=0.3\textwidth]{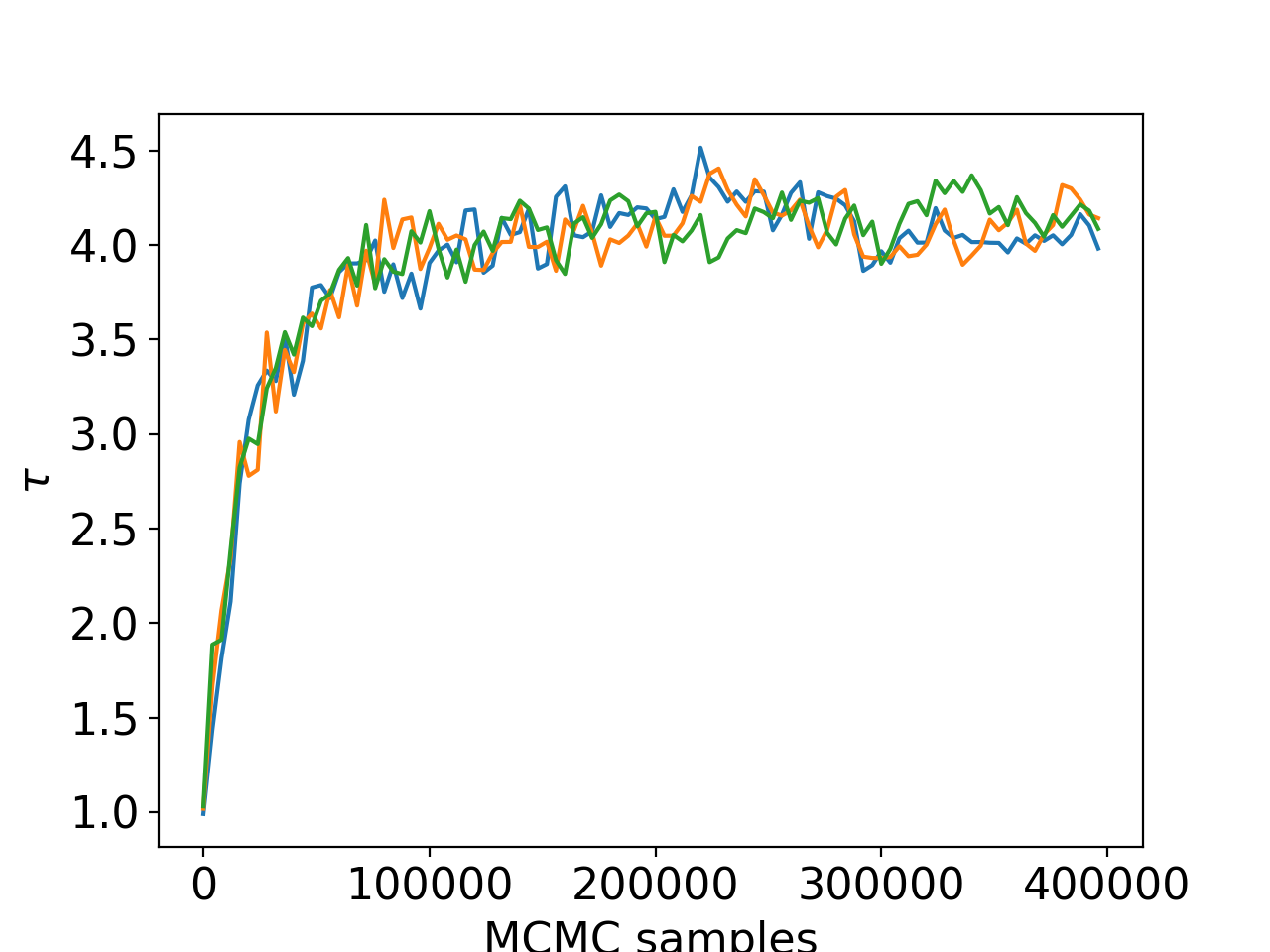}}\\
\subfloat[$\phi$ trace]{\includegraphics[width=0.3\textwidth]{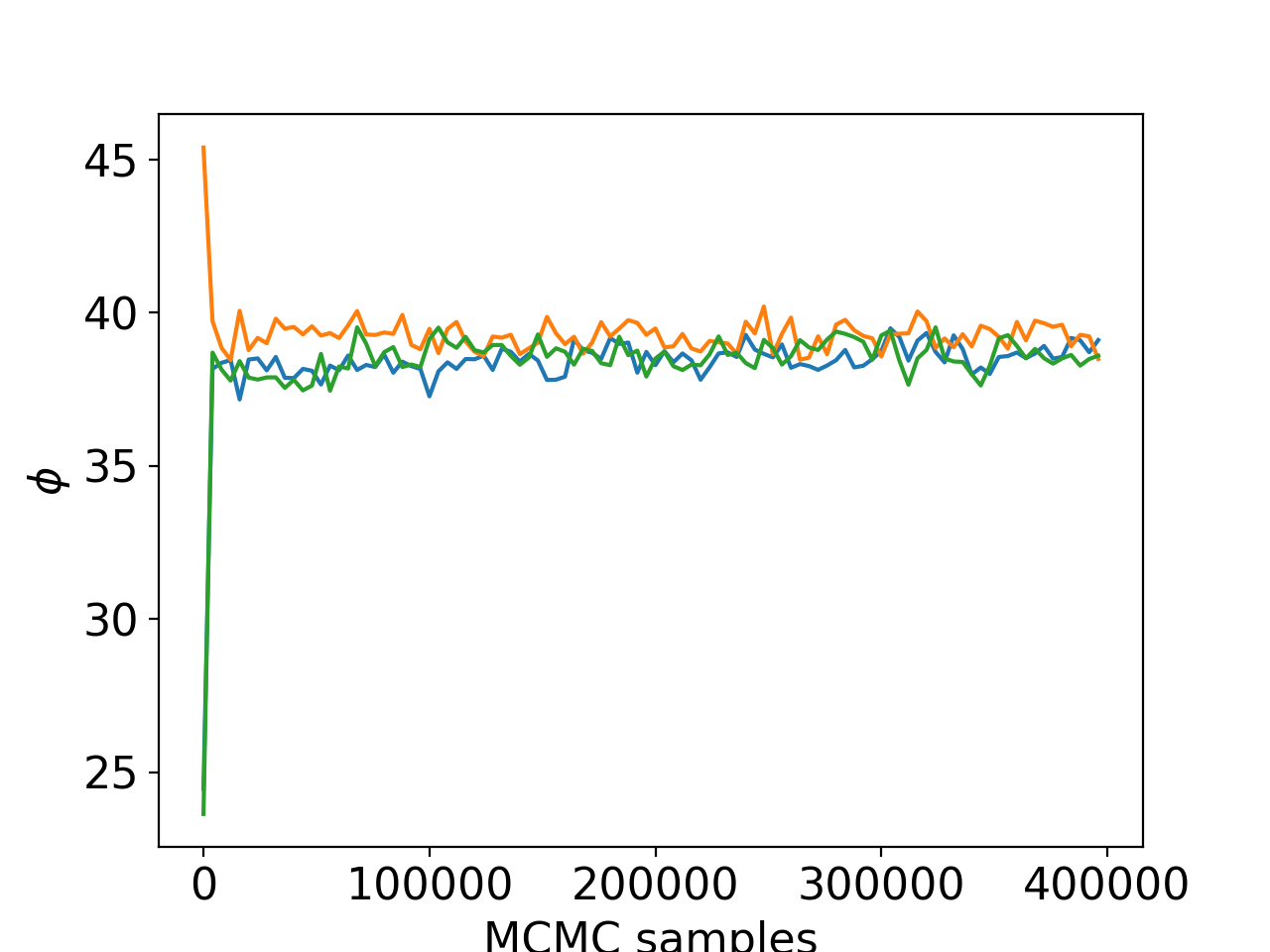}}
\subfloat[Logposterior trace]{\includegraphics[width=0.3\textwidth]{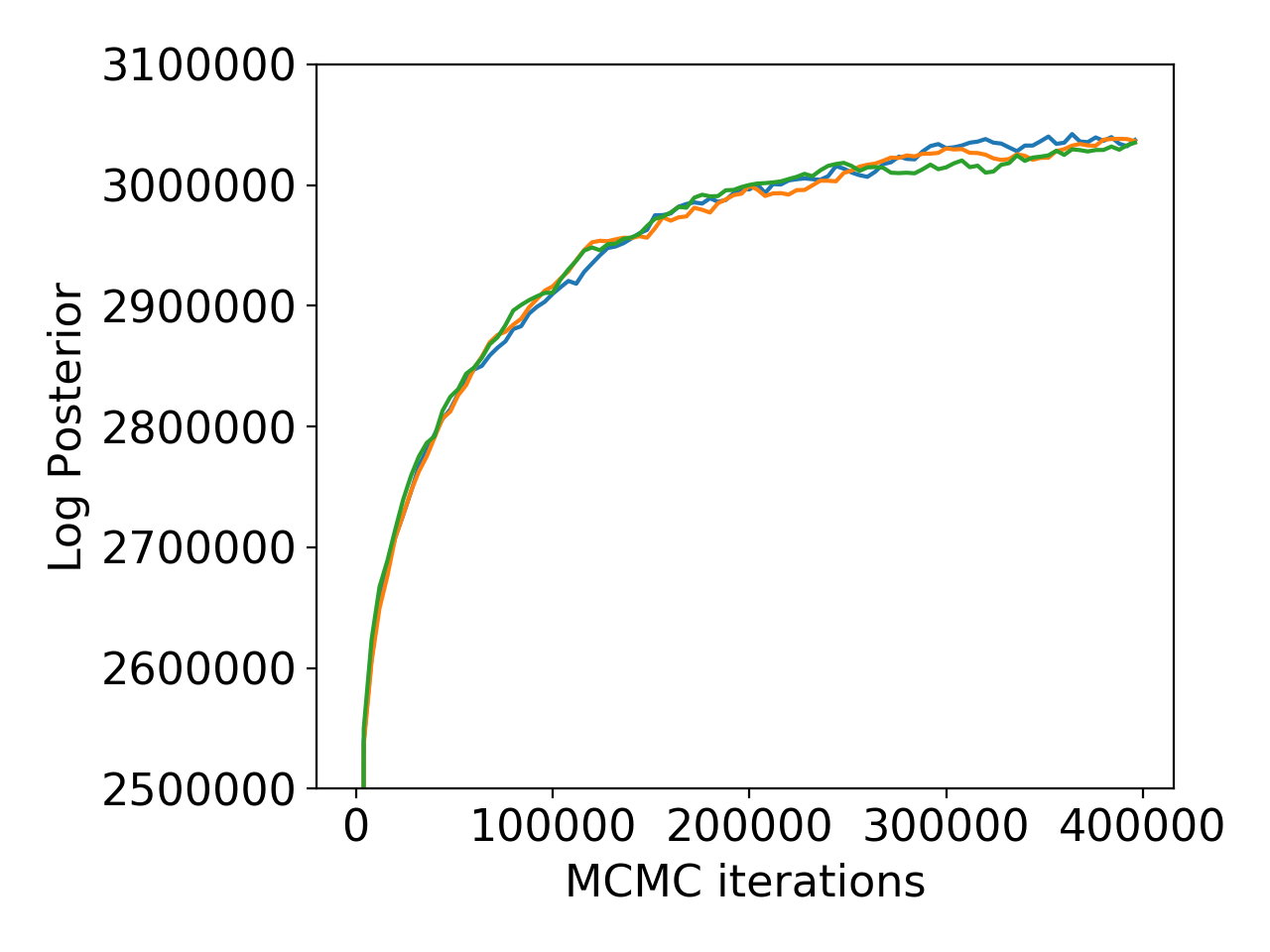}}

\caption{MCMC trace plots of hyperparameters and logposterior for the Reuters terror network.}
\label{fig:mcmc_trace_reuters_parameters}

\end{figure}

In Figure \ref{fig:mcmc_trace_reuters_parameters} we see the trace plots for the hyperparameters $\alpha, \sigma, \tau$ and $\phi$, as well at the logposterior, for the Reuters terror network. As before, we see that the hyperparameters have not yet converged. However, we found that running a longer chain provided similar credible intervals for the weights and degree distribution.

\end{document}